\documentclass[12pt]{article}
\usepackage[margin=0.9in]{geometry}
\usepackage{amsthm,amsmath,amssymb,amscd,enumerate, graphicx,caption,subcaption,bm,booktabs,longtable,colortbl,array,makecell}
\usepackage{setspace,array}
\usepackage{thmtools}
\usepackage{thm-restate}
\usepackage{color}
\usepackage{xcolor}
\usepackage{babel}
\usepackage{csquotes}
\usepackage{inputenc}
\usepackage[backend=biber, style=vancouver,sorting=none,natbib=true]{biblatex}
\usepackage[mathscr]{euscript}
\usepackage[shortlabels]{enumitem}
\usepackage{xr-hyper}
\usepackage{xr} 
\usepackage{hyperref}
\usepackage{algpseudocode}
\usepackage{algorithm}
\algrenewcommand\algorithmicindent{0.4em}

\usepackage{authblk}
\usepackage{siunitx}
\usepackage{multirow}
\usepackage{threeparttable}
\usepackage{threeparttablex}

\makeatletter
\newcommand*{\addFileDependency}[1]{
  \typeout{(#1)}
  \@addtofilelist{#1}
  \IfFileExists{#1}{}{\typeout{No file #1.}}
}
\makeatother

\newcommand*{\myexternaldocument}[1]{%
    \externaldocument{#1}%
    \addFileDependency{#1.tex}%
    \addFileDependency{#1.aux}%
}
\myexternaldocument{auc_supp}

\newif\ifarXiv
\arXivtrue 

\title{Behavior of prediction performance metrics with rare events}
\author[1]{Emily Minus}
\author[2,1]{R.~Yates Coley}
\author[2,1]{Susan~M. Shortreed}
\author[2,3,1,*]{Brian~D. Williamson}
\affil[1]{Department of Biostatistics, University of Washington}
\affil[2]{Biostatistics Division, Kaiser Permanente Washington Health Research Institute}
\affil[3]{Vaccine and Infectious Disease Division, Fred Hutchinson Cancer Center}
\affil[*]{Corresponding Author: Brian D. Williamson. Email: brian.d.williamson@kp.org}

\renewcommand\Authands{ and }

\begin{document}
\maketitle

\begin{abstract}
\textbf{Objective}
Area under the receiving operator characteristic curve (AUC) is commonly reported alongside prediction models for binary outcomes. Recent articles have raised concerns that AUC might be a misleading measure of prediction performance in the rare event setting. This setting is common since many events of clinical importance are rare. We aimed to determine whether the bias and variance of AUC are driven by the number of events or the event rate. We also investigated the behavior of other commonly used measures of prediction performance, including positive predictive value, accuracy, sensitivity, and specificity.

\textbf{Study Design and Setting}
We conducted a simulation study to determine when or whether AUC is unstable in the rare event setting by varying the size of datasets used to train and evaluate prediction models. This plasmode simulation study was based on data from the Mental Health Research Network; the data contained 149 predictors and the outcome of interest, suicide attempt, which had event rate 0.92\% in the original dataset.

\textbf{Results}
Our results indicate that poor AUC behavior---as measured by empirical bias, variability of cross-validated AUC estimates, and empirical coverage of confidence intervals---is driven by the number of events in a rare-event setting, not event rate. Performance of sensitivity is driven by the number of events, while that of specificity is driven by the number of non-events. Other measures, including positive predictive value and accuracy, depend on the event rate even in large samples.

\textbf{Conclusion}
AUC is reliable in the rare event setting provided that the total number of events is moderately large; in our simulations, we observed near zero bias with 1000 events.

\textbf{Plain Language Summary}
Predicting self-harm or suicidal behavior is medically important for guiding clinicians in providing care to patients. Several research teams have developed and evaluated suicide risk prediction models based on health records data. Part of evaluating these models is calculating area under the receiver operating characteristic curve (AUC) and other prediction performance metrics. Self-harm and suicide are rare events. Recent research has raised concerns with using AUC in rare-event settings. We aimed to determine whether having a sufficiently large dataset could remove these concerns. In our experiments, we found that AUC can be used without concern in settings with 1000 events or more. Thus, AUC is a valid measure of suicide risk prediction model performance in many large healthcare databases.
\end{abstract}

\begin{center}{\small \textbf{Keywords:} Prediction, classification, machine learning, area under the receiver operating characteristic curve, model evaluation, rare outcome.}\end{center}

\section*{What is new}
Key findings
\begin{itemize}
    \item AUC and other binary-outcome prediction performance metrics can be used reliably in rare-outcome settings so long as the minimum class size is large enough.
    \item Multiple prediction performance metrics should be reported in binary outcome settings, along with estimates of variability.
\end{itemize}
What this adds to what is known?
\begin{itemize}
    \item We provide evidence that the bias and variance of AUC, sensitivity, and specificity are driven by effective sample size (number of events for sensitivity; number of non-events for specificity; and the minimum of these two for AUC). In our simulations, we found that 1000 events was sufficient to have near zero bias for AUC and near nominal confidence interval coverage.
    \item We provide evidence that the bias and variance of other binary outcome prediction performance metrics also depends on the event rate.
\end{itemize}
What is the implication and what should change now?
\begin{itemize}
    \item Our results imply that in settings with rare events but large sample sizes---for example, predicting suicide attempt using millions of health care visits---AUC and other prediction performance metrics should reliably capture prediction performance. These measures should be used when scientifically relevant.
\end{itemize}
\doublespacing

\section{Introduction}

Clinical decision-making can be guided by prediction models that identify people at risk for negative health outcomes  \autocite{patton_predictive_2023,obermeyer_predicting_2016,chen_machine_2017,simon_predicting_2018}. Assessing model performance across many metrics is important. Selecting any one particular metric depends on how the model will  be used in care \autocite{cook2007use,pepe2007letter,janket2007letter}. 
Sensitivity and specificity \autocite{metz_basic_1978}, Brier score \autocite{steyerberg2001internal}, and the area under the receiver operating characteristic curve \autocite[AUC;][]{hanley_meaning_1982} are all useful performance metrics when considering binary outcomes. Estimating and reporting prediction model performance in the rare event setting presents several challenges. First, even with a large number of observations, the absolute number of events may be small. In training predictive models, the number of events, not just the total sample size, is of importance \autocite{vergouwe2005substantial}. Second, a small event rate impacts the behavior of some performance measures  \autocite{lever_classification_2016, saito_precision-recall_2015}. 

AUC is often reported in the scientific literature to describe the performance of risk prediction models, as we and others have done when estimating models to predict suicide attempt \autocite{kessler2017predicting,barak2017predicting,simon_predicting_2018,belsher_prediction_2019,chen2020predicting,gradus2020prediction, penfold2021predicting,coley_racialethnic_2021,walker2021evaluation,shortreed2023,simon2024stability}. In our prior work, the per-visit 90-day suicide attempt rate was 0.62\% for visits to mental health specialty clinics and 0.26\% for general medical visits with mental health diagnoses; suicide deaths were even more rare, approximately 0.02\% and 0.01\%, respectively \autocite{simon_predicting_2018}. Recently, several studies have argued that AUC is a misleading and inappropriate measure of prediction model performance in the rare event setting \autocite{saito_precision-recall_2015, lever_classification_2016, steyerberg2018poor, adhikari_revisiting_2021, bokhari_clinical_2023}. In practice, when using large sample sizes, even with very rare events, AUC estimates of suicide risk prediction model performance have aligned with other performance metrics; in other words, multiple metrics yield similar conclusions about model adequacy and strength of risk discrimination \autocite{kessler2017predicting,gradus2020prediction,simon2024stability}. This is contrary to concerns raised in recent literature \autocite{lever_classification_2016,adhikari_revisiting_2021}. It is essential for researchers and decision makers to have a clear understanding of the statistical properties of performance measures in rare-outcome settings, including bias (including the potential for optimism) and stability (i.e., estimator variability), as many prediction models are developed to predict rare outcomes. In this paper we examine bias and variance of AUC and other performance metrics in a variety of settings, varying both the sample size and the event rate.

\section{Materials and Methods}
\subsection{Prediction performance metrics: notation and definitions}\label{sec:performance}

We assume $n$ independent and identically distributed observations drawn from a distribution $P$, with data unit $(X, Y) \sim P$, covariate vector $X \in \mathbb{R}^p$ and binary outcome $Y \in \{0, 1\}$. We further assume a prediction model, $f$, trained on a sample from $P$, such that $f$ returns a predicted probability of $Y$ given $X$. The AUC of $f$ is
\begin{align*}
AUC(f, P) := P\{f(X_1) < f(X_2) \mid Y_1 = 0, Y_2 = 1\},
\end{align*}
where $(X_1, Y_1)$ and $(X_2, Y_2)$ are independent draws from $P$. In other words, AUC is the probability that a randomly sampled non-event has a lower predicted probability than a randomly sampled event. We define other common performance metrics in Table~\ref{tab:other_measures}.

\begin{table}[]
    \centering
    \begin{tabular}{l|c}
    \hline
       Metric & Definition \\ \hline
        Sensitivity & $\text{Sens}(f, c, P) := \ P\{f(X) > c \mid Y = 1\}$ \\
        Specificity & $\text{Spec}(f, c, P) := \ P\{f(X) < c \mid Y = 0\}$ \\
        Positive predictive value & $PPV(f, c, P) := \ P\{Y = 1 \mid f(X) > c\}$ \\
        Negative predictive value & $NPV(f, c, P) := \ P\{Y = 0 \mid f(X) < c\}$ \\
        F$_{0.5}$ score & $F_{0.5}(f, c, P) := \ (1 + 0.5^2)\frac{\text{Sens}(f, c, P) PPV(f, c, P)}{0.5^2\text{Sens}(f, c, P) + PPV(f, c, P)}$\\
        F$_1$ score & $F_1(f, c, P) := \ \frac{2}{\text{Sens}(f, c, P)^{-1} + PPV(f, c, P)^{-1}}$\\
        Classification accuracy & $\text{accuracy}(f, c, P) := P\{Y = I(f(X) > c)\}$ \\
        Brier score & $\text{Brier}(f, P) := E_P[\{Y - f(X)\}^2]$ \\
        \hline
    \end{tabular}
    \caption{Common binary-outcome prediction metrics for a prediction function $f$ at threshold $c \in [0,1]$ with respect to distribution $P$.}
    \label{tab:other_measures}
\end{table}

\subsection{Simulation methods}\label{sec:sims}

We examined the behavior of AUC and the performance metrics in Table~\ref{tab:other_measures} in a variety of rare-event settings, with both large and small sample size.

\subsubsection{Data}\label{sec:data}

To maintain both the complex outcome-predictor relationships and the complex correlation structure among predictors in real-world data \autocite{simon_predicting_2018,coley_racialethnic_2021}, we used a plasmode simulation approach \autocite{schreck_statistical_2024}. We obtained a dataset $D$ with $3,081,420$ independent observations by randomly sampling one mental health visit per patient from across seven Mental Health Research Network sites (\url{https://mhresearchnetwork.org}). These visits were made by patients aged 13 years or older and occurred between January 1, 2009 and September 30, 2017 \autocite{coley_racialethnic_2021}. 
For patients with a suicide attempt (fatal or non-fatal) during this time, we sampled a mental health visit within 90 days; for those without a suicide attempt, a randomly sampled mental health visit was selected. The event rate in $D$, $R_0$, was 0.92\%. We considered 149 predictors, including demographic characteristics, mental health and substance abuse diagnoses (including past self-inflicted injuries or poisonings), other medical diagnoses, past inpatient or emergency mental health care, dispensed mental health medications, the patient health questionnaire (PHQ) 8-item score (PHQ-8) representing depressive symptoms, and the PHQ 9th item response indicating thoughts of death or self-harm \autocite{simon_predicting_2018}. All predictors except patient age in years and PHQ-8 score (0--24) were coded as binary indicators. The PHQ item 9, which has 4 possible responses (0,1,2,3), was coded using 3 binary variables indicating each possible non-zero response.

\subsubsection{Simulation scenarios}

We used three types of outcome-generating functions for our simulations: logistic regression, ridge regression \autocite{hoerl1970ridge}, and random forest models \autocite{breiman2001random}. Outcome-generating models for event rates of 0.92\%, 0.46\%, and 1.84\% were created by sampling 1 million visits from $D$ with and without events to match the desired event rate. The tuning parameters used for each outcome-generating model are provided in  Table~\ref{tab:generate_outcome_tuning_parameters}; 10-fold cross-validated estimates of performance of these models is presented in Table~\ref{tab:original_values}.  

\begin{table}
\centering
\caption{Cross-validated values of all estimands on the original dataset for each  algorithm and event rate. Thresholds for all measures apart from AUC are percentiles of predicted risk (``Percentile'' column). \label{tab:original_values}}
\centering
\resizebox{\ifdim\width>\linewidth\linewidth\else\width\fi}{!}{
\fontsize{12}{14}\selectfont
\begin{threeparttable}
\begin{tabular}[t]{llllllllllll}
\toprule
Algorithm & Event rate & AUC & Brier & Percentile & Sens. & Spec. & Acc. & PPV & NPV & F1 & F0.5\\
\midrule
 &  &  &  & 90 & 0.613 & 0.876 & 0.875 & 0.022 & 0.998 & 0.043 & 0.028\\
\cmidrule{5-12}
 &  &  &  & 95 & 0.503 & 0.93 & 0.928 & 0.032 & 0.998 & 0.06 & 0.039\\
\cmidrule{5-12}
 & \multirow{-3}{*}{\raggedright\arraybackslash 0.005} & \multirow{-3}{*}{\raggedright\arraybackslash 0.806} & \multirow{-3}{*}{\raggedright\arraybackslash 0.004} & 99 & 0.263 & 0.989 & 0.985 & 0.097 & 0.997 & 0.141 & 0.111\\
\cmidrule{2-12}
 &  &  &  & 90 & 0.65 & 0.872 & 0.87 & 0.045 & 0.996 & 0.084 & 0.055\\
\cmidrule{5-12}
 &  &  &  & 95 & 0.534 & 0.936 & 0.932 & 0.072 & 0.995 & 0.127 & 0.087\\
\cmidrule{5-12}
 & \multirow{-3}{*}{\raggedright\arraybackslash 0.009} & \multirow{-3}{*}{\raggedright\arraybackslash 0.832} & \multirow{-3}{*}{\raggedright\arraybackslash 0.008} & 99 & 0.194 & 0.996 & 0.988 & 0.296 & 0.993 & 0.234 & 0.268\\
\cmidrule{2-12}
 &  &  &  & 90 & 0.671 & 0.881 & 0.877 & 0.095 & 0.993 & 0.167 & 0.115\\
\cmidrule{5-12}
 &  &  &  & 95 & 0.536 & 0.945 & 0.937 & 0.154 & 0.991 & 0.239 & 0.18\\
\cmidrule{5-12}
\multirow{-9}{*}[1\dimexpr\aboverulesep+\belowrulesep+\cmidrulewidth]{\raggedright\arraybackslash RF} & \multirow{-3}{*}{\raggedright\arraybackslash 0.018} & \multirow{-3}{*}{\raggedright\arraybackslash 0.854} & \multirow{-3}{*}{\raggedright\arraybackslash 0.015} & 99 & 0.144 & 0.998 & 0.982 & 0.568 & 0.984 & 0.229 & 0.357\\
\cmidrule{1-12}
 &  &  &  & 90 & 0.642 & 0.902 & 0.901 & 0.029 & 0.998 & 0.056 & 0.036\\
\cmidrule{5-12}
 &  &  &  & 95 & 0.517 & 0.952 & 0.95 & 0.047 & 0.998 & 0.087 & 0.058\\
\cmidrule{5-12}
 & \multirow{-3}{*}{\raggedright\arraybackslash 0.005} & \multirow{-3}{*}{\raggedright\arraybackslash 0.865} & \multirow{-3}{*}{\raggedright\arraybackslash 0.004} & 99 & 0.257 & 0.991 & 0.988 & 0.118 & 0.997 & 0.162 & 0.132\\
\cmidrule{2-12}
 &  &  &  & 90 & 0.65 & 0.872 & 0.87 & 0.045 & 0.996 & 0.084 & 0.055\\
\cmidrule{5-12}
 &  &  &  & 95 & 0.534 & 0.936 & 0.932 & 0.072 & 0.995 & 0.127 & 0.087\\
\cmidrule{5-12}
 & \multirow{-3}{*}{\raggedright\arraybackslash 0.009} &  & \multirow{-3}{*}{\raggedright\arraybackslash 0.008} & 99 & 0.194 & 0.996 & 0.988 & 0.296 & 0.993 & 0.234 & 0.268\\
\cmidrule{2-2}
\cmidrule{4-12}
 &  &  &  & 90 & 0.632 & 0.91 & 0.905 & 0.116 & 0.993 & 0.196 & 0.139\\
\cmidrule{5-12}
 &  &  &  & 95 & 0.494 & 0.958 & 0.95 & 0.181 & 0.99 & 0.265 & 0.208\\
\cmidrule{5-12}
\multirow{-9}{*}[1\dimexpr\aboverulesep+\belowrulesep+\cmidrulewidth]{\raggedright\arraybackslash Ridge} & \multirow{-3}{*}{\raggedright\arraybackslash 0.018} & \multirow{-6}{*}[0.5\dimexpr\aboverulesep+\belowrulesep+\cmidrulewidth]{\raggedright\arraybackslash 0.866} & \multirow{-3}{*}{\raggedright\arraybackslash 0.016} & 99 & 0.219 & 0.994 & 0.98 & 0.404 & 0.986 & 0.284 & 0.346\\
\cmidrule{1-12}
 &  &  &  & 90 & 0.647 & 0.902 & 0.901 & 0.03 & 0.998 & 0.057 & 0.037\\
\cmidrule{5-12}
 &  &  &  & 95 & 0.517 & 0.952 & 0.95 & 0.047 & 0.998 & 0.087 & 0.058\\
\cmidrule{5-12}
 & \multirow{-3}{*}{\raggedright\arraybackslash 0.005} & \multirow{-3}{*}{\raggedright\arraybackslash 0.867} & \multirow{-3}{*}{\raggedright\arraybackslash 0.004} & 99 & 0.257 & 0.991 & 0.988 & 0.117 & 0.997 & 0.161 & 0.132\\
\cmidrule{2-12}
 &  &  &  & 90 & 0.643 & 0.905 & 0.903 & 0.059 & 0.996 & 0.108 & 0.072\\
\cmidrule{5-12}
 &  &  &  & 95 & 0.509 & 0.954 & 0.95 & 0.093 & 0.995 & 0.158 & 0.112\\
\cmidrule{5-12}
 & \multirow{-3}{*}{\raggedright\arraybackslash 0.009} &  & \multirow{-3}{*}{\raggedright\arraybackslash 0.008} & 99 & 0.247 & 0.992 & 0.985 & 0.228 & 0.993 & 0.237 & 0.232\\
\cmidrule{2-2}
\cmidrule{4-12}
 &  &  &  & 90 & 0.635 & 0.91 & 0.905 & 0.117 & 0.993 & 0.197 & 0.139\\
\cmidrule{5-12}
 &  &  &  & 95 & 0.494 & 0.958 & 0.95 & 0.181 & 0.99 & 0.265 & 0.207\\
\cmidrule{5-12}
\multirow{-9}{*}[1\dimexpr\aboverulesep+\belowrulesep+\cmidrulewidth]{\raggedright\arraybackslash GLM} & \multirow{-3}{*}{\raggedright\arraybackslash 0.018} & \multirow{-6}{*}[0.5\dimexpr\aboverulesep+\belowrulesep+\cmidrulewidth]{\raggedright\arraybackslash 0.868} & \multirow{-3}{*}{\raggedright\arraybackslash 0.016} & 99 & 0.22 & 0.994 & 0.98 & 0.405 & 0.986 & 0.286 & 0.347\\
\bottomrule
\end{tabular}
\begin{tablenotes}
\item Abbreviations: Sens: sensitivity; Spec: specificity; Acc: accuracy; GLM: logistic regression; RF: random forests; Ridge: ridge regression.
\end{tablenotes}
\end{threeparttable}}
\end{table}

 To investigate the behavior of prediction performance metrics, we varied both (i) the event rate, keeping sample size fixed, and (ii) the absolute number of events, increasing sample size appropriately as the event rate decreases. We varied $n \in \{5,000; 10,000; 50,000; 100,000; 1,000,000\}$ and considered the three event rates defined above. For each simulation iteration we generated a training data set by sampling covariates for $n$ observations using bootstrap sampling with replacement from $D$, and generated outcomes according to each prediction model \autocite[see, e.g.,][]{franklin2017comparing}. We then repeated this procedure once for each sample size, event rate, and outcome-generating model to generate an evaluation data set of 1.6 million observations. Performance in this evaluation dataset, a large independent sample, is intended to capture true model performance. For each scenario, defined by event rate, sample size, and algorithm, we conducted 2500 simulation iterations. 

\subsubsection{Developing clinical prediction models}

For each simulation iteration we estimated logistic regression, ridge logistic regression, and random forests prediction models. While logistic regression is a common prediction approach in binary outcome settings, in some rare-event cases, it may be unstable, particularly with a large number of predictors. Ridge regression was used as a penalized logistic regression method  \autocite{hoerl1970ridge}. Random forests with probability trees \autocite{malley2012probability} was selected as an example of a more complex prediction modeling approach that estimated interactions between predictors automatically from the data. We did not include any interactions between predictors in models estimated with logistic regression or ridge regression. We used global tuning parameters in model estimation for each simulation iteration because our goal was to understand how the behavior of prediction model performance measures depended on event rate and absolute number of events, not estimate the optimal model for each data set. For information on tuning parameters used for the simulations, see Supplementary Materials Section~\ref{sec:fixing_overview}.

\subsubsection{Assessing clinical prediction model performance}

We used 10-fold outcome-stratified cross-validation to estimate performance metrics for each model on the training dataset. At smaller training set sizes and event rates, some folds had no events despite using outcome-stratified sampling (i.e., there were fewer than 10 events in the sampled training data). Folds without events were excluded from calculation of prediction performance. We used the $90^\text{th}$, $95^\text{th}$, and $99^\text{th}$ percentiles of predicted probabilities from the training fold for metrics requiring thresholds (all except AUC and Brier score). 

Code implementing asymptotic standard errors (ASEs) was not available for many metrics; thus, we used empirical standard errors (ESEs) to describe variability. While the ESE is not available in a data analysis, in simulations it provides a best-case benchmark for variance estimation. The ESE was defined as the standard deviation of estimates across the 2500 simulation replications. For AUC, in addition to the ESE, we calculated ASEs, as code was readily available \autocite{cvAUC_ledell_2015,cvAUC}.  

We used the evaluation dataset to describe the true prediction performance in our population, computing all performance metrics using the fitted model and thresholds from each replication. The final true population value of each performance metric was defined as the average of the evaluation-set values across the 2500 simulation replications.

We measured reliability for the performance metrics by calculating the empirical bias, variance, and mean squared error (MSE) of the estimated performance metric across the training-set estimated metrics. The evaluation-estimated mean performance metric was treated as the gold standard, i.e., the true performance (see Table~\ref{tab:measures}). We calculated empirical coverage of 95\% confidence intervals (CIs) using the ASE for AUC and the ESE for all metrics, comparing the CI from each replication with the evaluation-estimated mean performance metric. 

\begin{table}[h]
\centering
\caption[Calculation of performance measures from the simulation replicates]{Calculation of reliability from the simulation replicates. Total number of simulations $r = 2500$, cross-validated performance metric $m_\text{CV}$ estimated on the training dataset, true performance metric $m_0$ obtained on the evaluation dataset, and $I_{95\%\text{CI}}(x)$ is an indicator function that equals one if $x$ falls within the bounds of the 95\% CI (inclusive on both sides), and zero otherwise.}
\begin{tabular}{l c} 
 \hline
 Performance measure & Formula or measure(s) \\ 
 \hline
 Empirical bias & $\displaystyle{\frac{1}{r}\sum^r_{i=1} (m_{i,\text{CV}} - m_{0})}$ \\
 Empirical standard error (ESE) & $\displaystyle{\frac{1}{r}\sum_{i = 1}^r \left(m_{i,CV} - \frac{1}{r}\sum_{i = 1}^r m_{i,CV}\right)^2}$ \\
 Empirical coverage of 95\% CI &  $\displaystyle{\frac{1}{r}\sum^r_{i=1} I_{95\%\text{CI},i}(m_0)}$ \\
 Empirical mean squared error & $\displaystyle{\frac{1}{r}\sum^r_{i=1} \Big(m_{i,\text{CV}} - m_0\Big)^2}$ \\
 \hline
\end{tabular}
\label{tab:measures}
\end{table}

\subsubsection{Computing}

All simulations were conducted in R version 4.1.1 \autocite{Rpackage}. Logistic regression was implemented using the base \texttt{R} \texttt{stats} package, ridge regression using  \texttt{glmnet} (version 4.1-4) \autocite{glmnet}, and random forests using \texttt{ranger} (version 0.14.1) \autocite{ranger}. ASE-based CIs for cross-validated AUC were obtained using \texttt{cvAUC} \autocite{cvAUC}. The \texttt{foreach} package was used to implement parallel computing \autocite{daniel_foreach_2022}. 

\section{Results}\label{sec:results} 

We define the \textit{effective sample size} (ESS) as the amount of information in the entire training sample used to estimate a metric. Therefore, for sensitivity, the ESS is the number of events in the training sample; for specificity, it is the number of non-events. For AUC, the ESS is the minimum of the number of events and non-events; because we consider event rates less than 50\%, in our case the ESS is the number of events. 
The true value of each performance measure depends on the sample size used to estimate the model, event rate, and model-estimation algorithm. The true values for AUC are shown in Table~\ref{tab:true_aucs}; for other estimands see  Table~\ref{tab:all_true_values_test}.

\begin{table}
\centering
\caption{True (evaluation-set-estimated mean) values of AUC, averaged over 2500 Monte-Carlo replications, for each sample size, algorithm, and event rate. \label{tab:true_aucs}}
\centering
\resizebox{\ifdim\width>\linewidth\linewidth\else\width\fi}{!}{
\fontsize{12}{14}\selectfont
\begin{threeparttable}
\begin{tabular}[t]{llllllllll}
\toprule
\multicolumn{1}{c}{ } & \multicolumn{3}{c}{Event rate = 0.009} & \multicolumn{3}{c}{Event rate = 0.018} & \multicolumn{3}{c}{Event rate = 0.005} \\
\cmidrule(l{3pt}r{3pt}){2-4} \cmidrule(l{3pt}r{3pt}){5-7} \cmidrule(l{3pt}r{3pt}){8-10}
n & GLM & RF & Ridge & GLM & RF & Ridge & GLM & RF & Ridge\\
\midrule
\num{5.00e+03} & 0.571 & 0.8 & 0.813 & 0.656 & 0.799 & 0.807 & 0.533 & 0.768 & 0.817\\
\cmidrule{1-10}
\num{1.00e+04} & 0.669 & 0.806 & 0.819 & 0.745 & 0.8 & 0.814 & 0.565 & 0.796 & 0.823\\
\cmidrule{1-10}
\num{5.00e+04} & 0.83 & 0.821 & 0.831 & 0.835 & 0.816 & 0.824 & 0.809 & 0.813 & 0.836\\
\cmidrule{1-10}
\num{1.00e+05} & 0.846 & 0.826 & 0.834 & 0.842 & 0.82 & 0.827 & 0.843 & 0.817 & 0.841\\
\cmidrule{1-10}
\num{1.00e+06} & 0.855 & 0.833 & 0.839 & 0.847 & 0.824 & 0.83 & 0.863 & 0.832 & 0.847\\
\bottomrule
\end{tabular}
\begin{tablenotes}
\item Abbreviations: GLM: logistic regression; RF: random forests; Ridge: ridge regression.
\end{tablenotes}
\end{threeparttable}}
\end{table}

\begin{figure}
\centering
\includegraphics[width=1\textwidth]{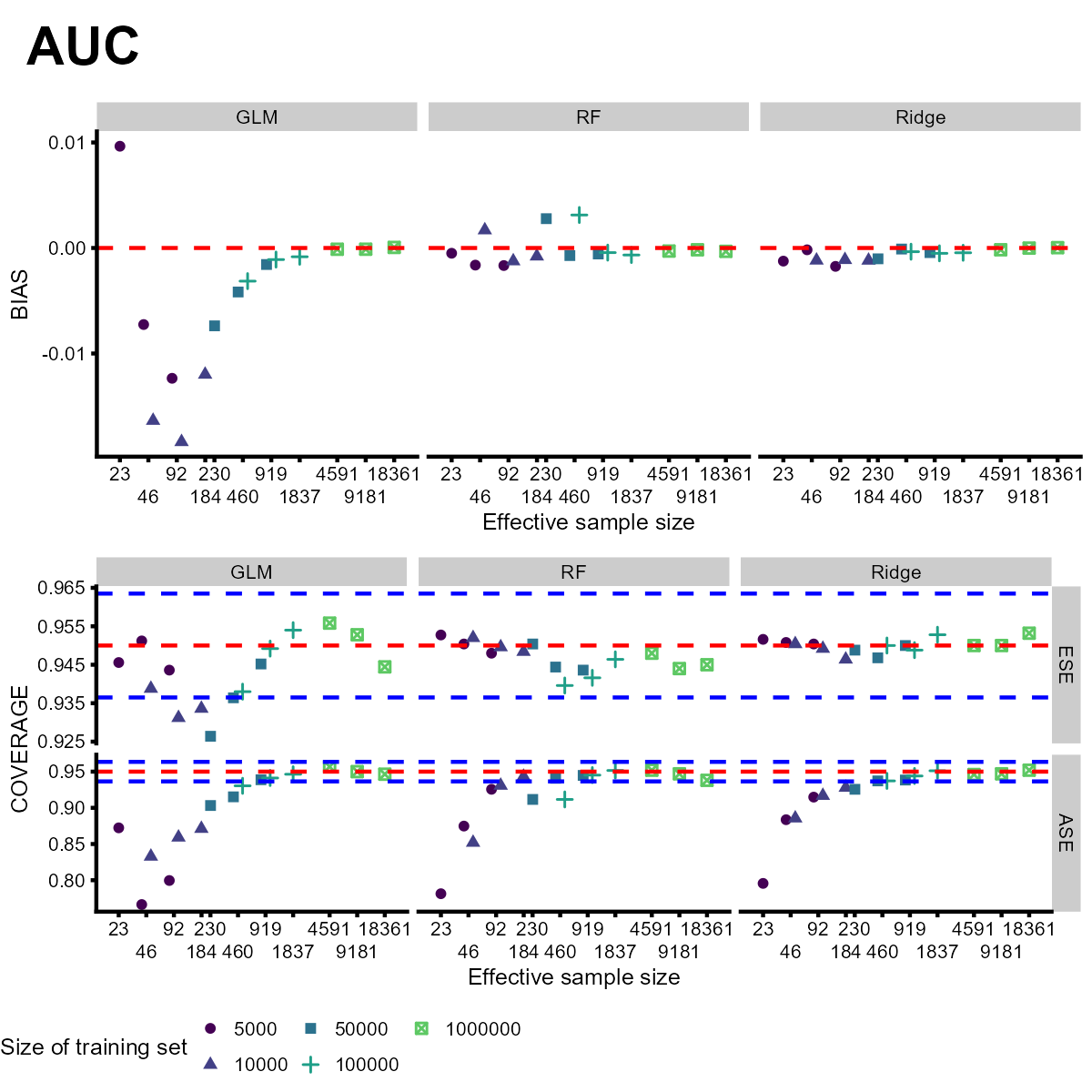}
\caption{Empirical bias and coverage of 95\% confidence intervals for estimating the evaluation-set AUC (values provided in Table~\ref{tab:true_aucs}) versus effective sample size (number of events in the training set) in the rows; columns show logistic regression (GLM) including all predictors, random forests (RF), and ridge logistic regression (Ridge) including all predictors. Estimates from the training dataset were compared to true values computed on the evaluation dataset. Colors and shapes show the training set size; for each training set size, an increasing event rate leads to a larger effective sample size. ESE = empirical standard error, ASE = asymptotic standard error. The blue dashed lines around 95\% coverage indicate one Monte-Carlo standard error.}
\label{fig:auc_main}
\end{figure}

Results describing the behavior of AUC estimates are in Figure~\ref{fig:auc_main}. Bias (top panel) for all algorithms decreases to zero as the ESS increases. 
For each sample size, bias decreases as the event rate increases, but once a minimum number of events is reached (approximately 1000 events), the event rate no longer impacts bias.
The three right farthest dots in each panel show AUC bias of zero at each event rate for a training sample size of 1,000,000.
In Figure~\ref{fig:auc_vs_n_supp}, we show that for each training set size, 
bias goes to zero for each event rate. These two plots show the ESS, not the event rate, drives AUC bias. This is supported by the AUC influence function form \autocite[][Theorem 4.1]{cvAUC_ledell_2015}: the estimated AUC (and its asymptotic variance) depends on the number of events and number of non-events, not the event rate itself.  

ESE-based AUC 95\% CI coverage (bottom panel of Figure~\ref{fig:auc_main}) is near the nominal level at all sample sizes, while ASE-based CI coverage followed the same pattern as bias, increasing to the nominal level by an ESS of 1000. In Figure~\ref{fig:auc_supp}, we show MSE and CI width for AUC, demonstrating the variance decreases with increasing ESS. We expect poor coverage of ASE-based intervals when the ESS is small, since these are asymptotic intervals \autocite{cvAUC_ledell_2015}. We observed poor AUC behavior (bias and low coverage) for the logistic regression model when the number of predictors exceeded the ESS. This is consistent with work that has shown increased bias and variability in parameter estimates when the number of events per variable is low. A commonly suggested guideline, and one supported by these simulations, is there should be ten or more events per variable to estimate a logistic regression model \autocite{peduzzi_simulation_1996}. 

\begin{figure}
\centering
\includegraphics[width=1\textwidth]{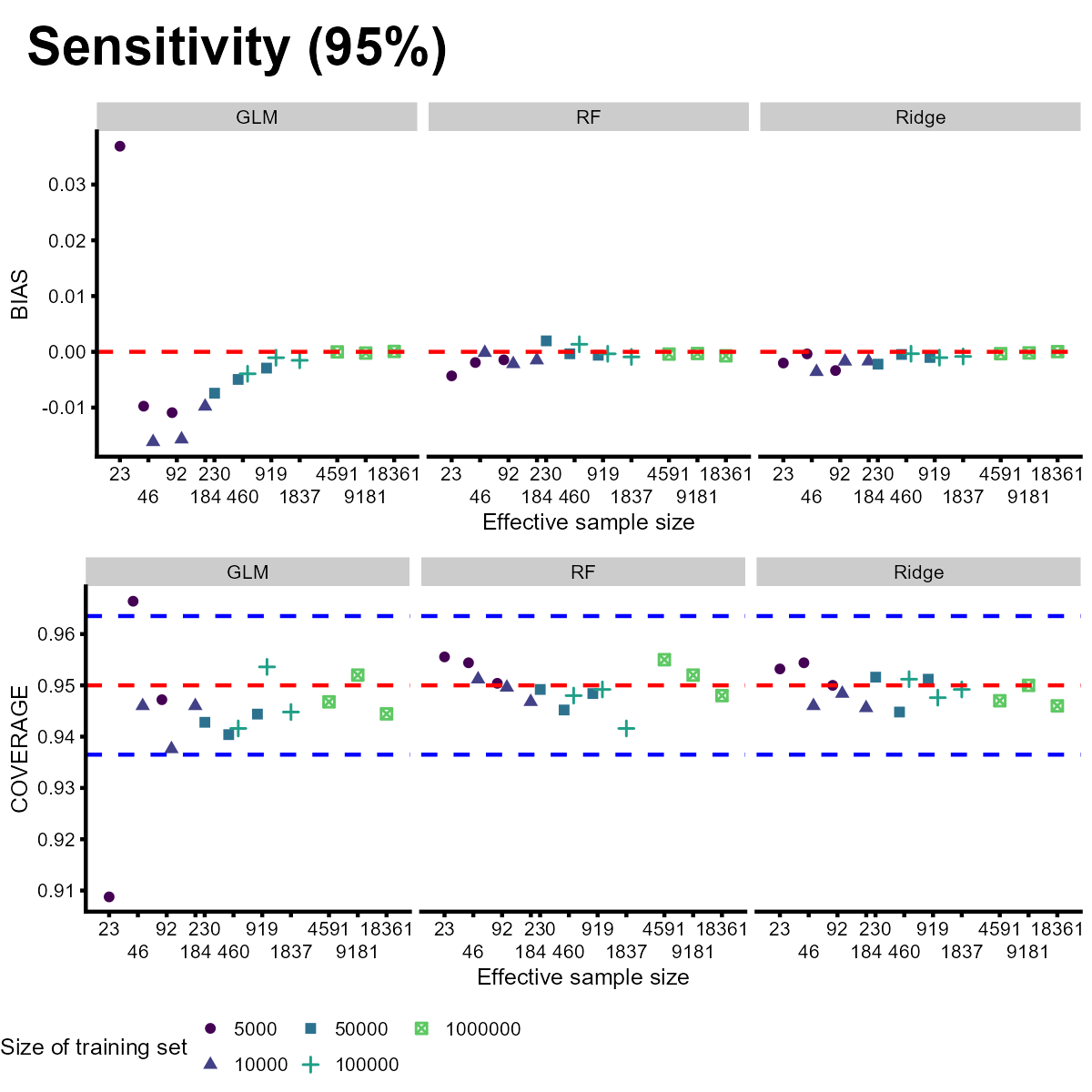}
\caption{Empirical bias and coverage of 95\% confidence intervals for estimating the evaluation-set sensitivity at the $95^\text{th}$ percentile of predicted risk versus effective sample size, using logistic regression (GLM), random forests (RF), and ridge regression (Ridge). Effective sample size for sensitivity is the number of events in the training set; for specificity, it is the number of non-events in the training set. Estimates from the training dataset were compared to true values computed on the evaluation dataset. Colors and shapes show the training set size; at each fixed number of events, the larger training set size implies a smaller event rate. The blue dashed lines around 95\% coverage indicate one Monte-Carlo standard error.}
\label{fig:sens_main}
\end{figure}

\begin{figure}
\centering
\includegraphics[width=1\textwidth]{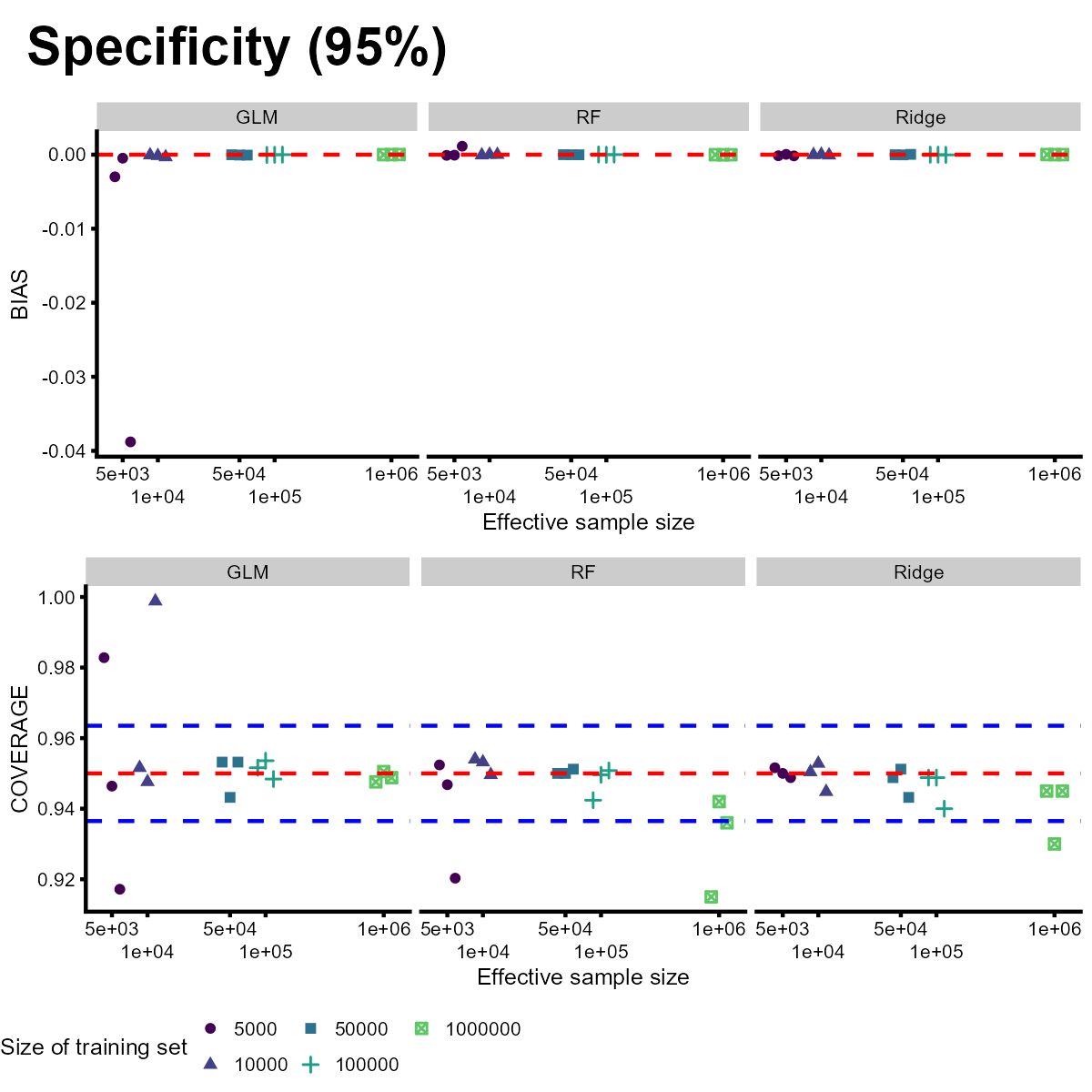}
\caption{Empirical bias and coverage of 95\% confidence intervals for estimating the evaluation-set specificity at the 95$^\text{th}$ percentile of predicted risk versus effective sample size, using logistic regression (GLM), random forests (RF), and ridge regression (Ridge). Effective sample size for sensitivity is the number of events in the training set; for specificity, it is the number of non-events in the training set. Estimates from the training dataset were compared to true values computed on the evaluation dataset. Colors and shapes show the training set size; at each fixed number of events, the larger training set size implies a smaller event rate. The blue dashed lines around 95\% coverage indicate one Monte-Carlo standard error.}
\label{fig:spec_main}
\end{figure}

Figures~\ref{fig:sens_main} and \ref{fig:spec_main} show that the performance of sensitivity and specificity at the 95$^\text{th}$ percentile of predicted risk are also driven by the ESS (see Supplementary Tables~\ref{tab:sens_90_test}--\ref{tab:spec_99_test} for 90$^\text{th}$ and 99$^\text{th}$ percentile, which have a similar pattern). Bias decreases with increasing ESS. ESE-based CI coverage is at the nominal level for sensitivity. For specificity, CI coverage follows a slightly different pattern; while mostly near the nominal level, when the true specificity is close to 1, there is instability in CI coverage estimates. 

\begin{figure}
\includegraphics[width=1\textwidth]{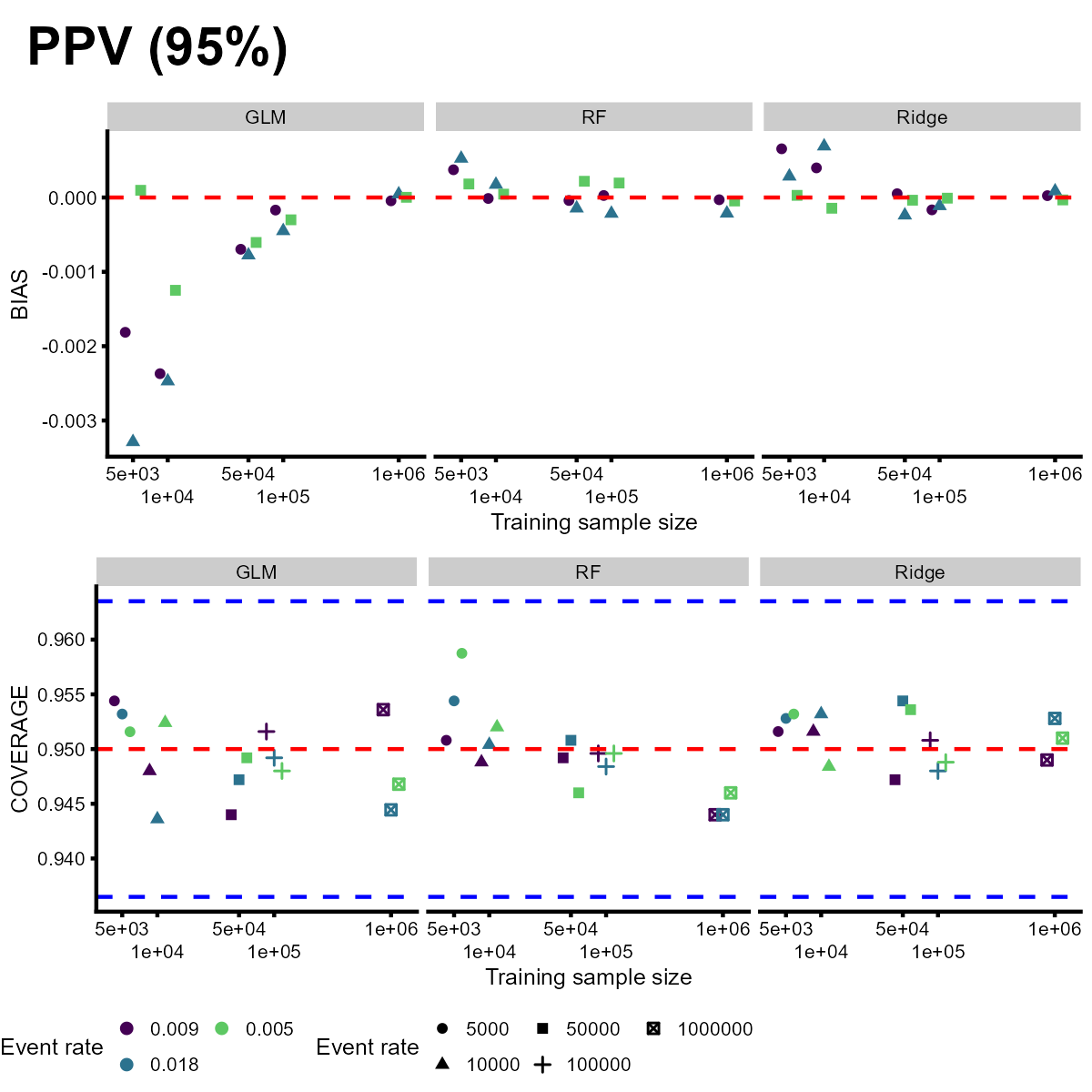}
\caption{Empirical bias and coverage of 95\% confidence intervals for estimating the evaluation-set PPV at the 95th percentile of predicted risk versus effective sample size (number of events in the training set) using logistic regression (GLM), random forests (RF), and ridge regression (Ridge). Estimates from the training dataset were compared to true values computed on the evaluation dataset. Colors and shapes show the training set size; at each fixed number of events, the larger training set size implies a smaller event rate. The blue dashed lines around 95\% coverage indicate one Monte-Carlo standard error.}
\label{fig:ppv_main}
\end{figure}

In contrast to AUC, sensitivity, and specificity, the performance of accuracy, PPV, NPV, $F_1$, $F_{0.5}$, and Brier score relied in part on the event rate itself, rather than just on the ESS. We display the results for PPV at the 95$^\text{th}$ percentile of predicted risk in Figure~\ref{fig:ppv_main}; performance for accuracy, NPV, $F_1$, $F_{0.5}$, and Brier score and the other algorithms is similar to the results for PPV (Supplementary Tables~\ref{tab:accuracy_90_test}--\ref{tab:brier_test}, which also include all metrics at the 90$^\text{th}$ and 99$^\text{th}$ percentiles). For these estimands, at a given training set size, bias and coverage tended to be worse at the lower event rates.

\section{Discussion}\label{sec:discussion}

Our numerical experiments demonstrate that bias and CI coverage of AUC, sensitivity, and specificity are driven by ESS, not  event rate. Stability was reached early for specificity, as we only considered rare events and the ESS for specificity is the number of non-events. For AUC and sensitivity, once a minimum number of events (approximately 1000) was reached, through {\it either} a large sample size or higher event rate, bias was near zero and CI coverage was near nominal. Even in the smallest event rate we considered, $R=0.0046$ ( approximately one event per 217 non-events) our simulations show AUC estimates had low bias as long as the training set was sufficiently large (i.e., $n$ = 400,000). 

 In contrast to AUC, the behavior of accuracy, Brier score, PPV, NPV, $F_1$ score, and $F_{0.5}$ score all showed some direct reliance on both event rate and training set size. This is not surprising, since the marginal outcome probability is included in the definition of these metrics. However, bias converged to zero and coverage approached the nominal level as the ESS grew, regardless of event rate. While we focused on binary classification in this article, ESS, rather than simply class proportions or number of events, likely plays an important role in bias and CI coverage of prediction performance metrics in settings with multi-class or time-to-event outcomes. Our work also emphasized the need to choose an appropriate prediction modeling approach for the data, which may include dimension reduction. For example, logistic regression did not perform well without a minimum number of events per predictor. Finally, our results highlight the importance of reporting multiple performance metrics with CIs, an approach which has been advocated by others \autocite{pepe2007letter,hand2012assessing, lever_classification_2016, adhikari_revisiting_2021}.  

There are some limitations to our work. First, we did not vary the strength of association between predictors and outcomes. It is possible bias could improve at smaller ESS if there are strong predictors of the outcome. Second, we do not have ASE formulas for metrics besides AUC. While the bootstrap might be tempting to construct standard errors, poor performance of ESE-based intervals, due to the performance metrics being near the boundary values of 0 or 1, suggests the standard bootstrap may not perform well. Bootstrap procedures designed for constructing CIs when estimates are near the boundary are available \autocite{andrews2000inconsistency}.  

Our work shows that AUC estimates are stable and reliable if the ESS is large enough. That is, unbiased AUC estimates are produced if the sample size is large enough even for low event rates. Additionally, AUC ASE-based CIs had nominal coverage at large enough sample sizes, indicating that ASE-based CIs can be reliably used.  

\section*{Acknowledgments}
Research reported in this publication was supported by the National Institute of Mental Health of the National Institutes of Health under award numbers R01 MH125821, U19-MH099201, and U19-MH121738. The content is solely the responsibility of the authors and does not necessarily represent the official views of the National Institutes of Health.

\newpage

\ifarXiv
\newif\ifnotarXiv
\notarXivfalse
\newif\ifarXiv
\newif\ifblinded

\ifnotarXiv
    \arXivfalse
    \blindedfalse
\else 
    \arXivtrue 
    \blindedfalse
\fi

\ifarXiv
\section*{SUPPLEMENTARY MATERIALS}
\fi

\ifnotarXiv
\documentclass[12pt]{article}
\usepackage[margin=0.9in]{geometry}
\usepackage{amsthm,amsmath,amssymb,amscd,enumerate, graphicx,caption,subcaption,bm,booktabs,longtable,colortbl,array,makecell}
\usepackage{setspace,array}
\usepackage{thmtools}
\usepackage{thm-restate}
\usepackage{color}
\usepackage{xcolor}
\usepackage{natbib}
\usepackage[mathscr]{euscript}
\usepackage[shortlabels]{enumitem}
\usepackage[colorlinks=true, urlcolor=blue, linkcolor=blue, citecolor=blue]{hyperref}
\usepackage{algpseudocode}
\usepackage{algorithm}
\algrenewcommand\algorithmicindent{0.4em}
\usepackage{natbib}
\usepackage[T1]{fontenc}
\usepackage[utf8]{inputenc}
\usepackage{authblk}
\usepackage{siunitx}
\usepackage{multirow}
\usepackage{threeparttable}
\usepackage{threeparttablex}
\title{Supplementary Materials for ``Behavior of prediction performance metrics with rare events''}

\ifblinded
\author{}
\else 
\author[1]{Emily Minus}
\author[2,1]{R.~Yates Coley}
\author[2,1]{Susan~M. Shortreed}
\author[2,3,1,*]{Brian~D. Williamson}
\affil[1]{Department of Biostatistics, University of Washington}
\affil[2]{Biostatistics Division, Kaiser Permanente Washington Health Research Institute}
\affil[3]{Vaccine and Infectious Disease Division, Fred Hutchinson Cancer Center}
\affil[*]{Corresponding Author: Brian D. Williamson. Email: brian.d.williamson@kp.org}

\renewcommand\Authands{ and }
\fi

\begin{document}

\maketitle
\fi

\makeatletter
\renewcommand \thesection{S\@arabic\c@section}
\renewcommand\thetable{S\@arabic\c@table}
\renewcommand \thefigure{S\@arabic\c@figure}
\makeatother
\setcounter{figure}{0}  
\setcounter{table}{0}
\setcounter{section}{0}

\section{Additional simulation results}

\subsection{Tuning parameters}

In Table~\ref{tab:generate_outcome_tuning_parameters}, we provide the tuning parameters used in the original outcome regression fits used to generate outcomes in the plasmode simulations presented in the main manuscript.
\begin{table}[h]
    \centering
    \begin{tabular}{ll|l}
       Event rate  & Algorithm & Tuning parameter(s) \\ \hline
        $R_0/2$ & GLM & --- \\
        $R_0$ & GLM & --- \\
        $2R_0$ & GLM & --- \\
        $R_0/2$ & Ridge & $\lambda = 0.000887$ \\
        $R_0$ & Ridge & $\lambda = 0.00165$ \\
        $2R_0$ & Ridge & $\lambda = 0.002987$ \\
        $R_0/2$ & RF & \texttt{mtry} = 12, \texttt{ntree} = 50, \texttt{min.node.size} = 10,000 \\
        $R_0$ & RF & \texttt{mtry} = 12, \texttt{ntree} = 50, \texttt{min.node.size} = 10,000 \\
        $2R_0$ & RF & \texttt{mtry} = 12, \texttt{ntree} = 50, \texttt{min.node.size} = 10,000 \\
    \end{tabular}
    \caption{Tuning parameters used in the original outcome regression functions for generating outcomes in the main plasmode simulations. GLM = logistic regression, Ridge = ridge regression, RF = random forests. \texttt{mtry} = the number of variables to consider for splitting (12 = $\sqrt{149}$, the total number of predictors); \texttt{ntree} = the number of trees; \texttt{min.node.size} = minimum node size.}
    \label{tab:generate_outcome_tuning_parameters}
\end{table}

In Table~\ref{tab:sim_tuning_parameters_generate_outcome}, we present the tuning parameters used to fit each algorithm at each sample size and event rate in the simulations. 
\begin{table}[h]
\centering
\caption{Tuning parameters used for ridge regression and random forests at each training-set sample size. For ridge regression, $\lambda$ tuning parameter values are provided. For random forests, minimum node size values are provided; the number of predictors sampled as candidates at each split was fixed at 12 (the floor of $\sqrt{149}$, where 149 is the number of predictors) and the number of trees was fixed at 50.}
\label{tab:sim_tuning_parameters_generate_outcome}
\begin{tabular}{ p{0.65in} l | c c c c c} 
 \hline
 & & \multicolumn{5}{c}{Training set size} \\
 Algorithm & Event rate &  $5\times 10^3$ & $1\times 10^4$ & $5\times 10^4$ & $1\times 10^5$ &  $1\times 10^6$\\ 
 \hline
 Ridge & $R_0/2$ & 0.368710 & 0.112562 & 0.007737 & 0.004101	&0.000887 \\
 Ridge & $R_0$ & 0.271674 & 0.056800 & 0.007044 & 0.001855 & 0.001650 \\
 Ridge & $2R_0$ & 0.137977 & 0.049776 & 0.003318 & 0.002995 & 0.002987 \\
 Random forests & $R_0/2$ & 1000 & 5000 & 5000 & 5000 & 15000 \\
 Random forests & $R_0$ & 1000 & 5000 & 5000 & 5000 & 15000 \\
 Random forests & $2R_0$ & 1000 & 5000 & 5000 & 5000 & 50000 \\
 \hline
\end{tabular}
\end{table}

\subsection{True values of estimands}

In Table~\ref{tab:all_true_values_test}, we present the true values for each estimand at each combination of sample size, event rate, and algorithm.

\begingroup\fontsize{12}{14}\selectfont

\begin{ThreePartTable}
\begin{TableNotes}
\item Abbreviations: GLM: logistic regression; RF: random forests; Ridge: ridge regression.
\end{TableNotes}
\begin{longtable}[t]{lllllllllll}
\caption{True (evaluation-set-estimated mean) values of all estimands, averaged over 2500 Monte-Carlo replications, for each sample size, algorithm, and event rate. \label{tab:all_true_values_test}}\\
\toprule
\multicolumn{2}{c}{ } & \multicolumn{3}{c}{Event rate = 0.009} & \multicolumn{3}{c}{Event rate = 0.018} & \multicolumn{3}{c}{Event rate = 0.005} \\
\cmidrule(l{3pt}r{3pt}){3-5} \cmidrule(l{3pt}r{3pt}){6-8} \cmidrule(l{3pt}r{3pt}){9-11}
n & Estimand & GLM & RF & Ridge & GLM & RF & Ridge & GLM & RF & Ridge\\
\midrule
\endfirsthead
\caption[]{True evaluation-set values of all estimands, averaged over 2500 Monte-Carlo replications, for each sample size, algorithm, and event rate. \textit{(continued)}}\\
\toprule
n & Estimand & GLM & RF & Ridge & GLM & RF & Ridge & GLM & RF & Ridge\\
\midrule
\endhead

\endfoot
\bottomrule
\insertTableNotes
\endlastfoot
\num{5.00e+03} & AUC & 0.571 & 0.8 & 0.813 & 0.656 & 0.799 & 0.807 & 0.533 & 0.768 & 0.817\\
\cmidrule{1-11}\pagebreak[0]
\num{1.00e+04} & AUC & 0.669 & 0.806 & 0.819 & 0.745 & 0.8 & 0.814 & 0.565 & 0.796 & 0.823\\
\cmidrule{1-11}\pagebreak[0]
\num{5.00e+04} & AUC & 0.83 & 0.821 & 0.831 & 0.835 & 0.816 & 0.824 & 0.809 & 0.813 & 0.836\\
\cmidrule{1-11}\pagebreak[0]
\num{1.00e+05} & AUC & 0.846 & 0.826 & 0.834 & 0.842 & 0.82 & 0.827 & 0.843 & 0.817 & 0.841\\
\cmidrule{1-11}\pagebreak[0]
\num{1.00e+06} & AUC & 0.855 & 0.833 & 0.839 & 0.847 & 0.824 & 0.83 & 0.863 & 0.832 & 0.847\\
\cmidrule{1-11}\pagebreak[0]
\num{5.00e+03} & Sensitivity 95\% & 0.215 & 0.325 & 0.429 & 0.272 & 0.314 & 0.401 & 0.329 & 0.301 & 0.442\\
\cmidrule{1-11}\pagebreak[0]
\num{1.00e+04} & Sensitivity 95\% & 0.309 & 0.336 & 0.438 & 0.35 & 0.319 & 0.411 & 0.229 & 0.325 & 0.453\\
\cmidrule{1-11}\pagebreak[0]
\num{5.00e+04} & Sensitivity 95\% & 0.458 & 0.354 & 0.46 & 0.44 & 0.332 & 0.429 & 0.454 & 0.353 & 0.479\\
\cmidrule{1-11}\pagebreak[0]
\num{1.00e+05} & Sensitivity 95\% & 0.479 & 0.359 & 0.466 & 0.451 & 0.336 & 0.434 & 0.492 & 0.358 & 0.487\\
\cmidrule{1-11}\pagebreak[0]
\num{1.00e+06} & Sensitivity 95\% & 0.495 & 0.369 & 0.475 & 0.46 & 0.34 & 0.44 & 0.522 & 0.372 & 0.5\\
\cmidrule{1-11}\pagebreak[0]
\num{5.00e+03} & Specificity 95\% & 0.951 & 0.952 & 0.953 & 0.954 & 0.954 & 0.956 & 0.758 & 0.947 & 0.951\\
\cmidrule{1-11}\pagebreak[0]
\num{1.00e+04} & Specificity 95\% & 0.953 & 0.953 & 0.953 & 0.956 & 0.955 & 0.956 & 0.952 & 0.951 & 0.952\\
\cmidrule{1-11}\pagebreak[0]
\num{5.00e+04} & Specificity 95\% & 0.954 & 0.953 & 0.954 & 0.957 & 0.955 & 0.957 & 0.952 & 0.951 & 0.952\\
\cmidrule{1-11}\pagebreak[0]
\num{1.00e+05} & Specificity 95\% & 0.954 & 0.953 & 0.954 & 0.957 & 0.955 & 0.957 & 0.952 & 0.951 & 0.952\\
\cmidrule{1-11}\pagebreak[0]
\num{1.00e+06} & Specificity 95\% & 0.954 & 0.953 & 0.954 & 0.957 & 0.955 & 0.957 & 0.952 & 0.952 & 0.952\\
\cmidrule{1-11}\pagebreak[0]
\num{5.00e+03} & PPV 95\% & 0.039 & 0.059 & 0.078 & 0.093 & 0.11 & 0.136 & 0.013 & 0.027 & 0.043\\
\cmidrule{1-11}\pagebreak[0]
\num{1.00e+04} & PPV 95\% & 0.057 & 0.061 & 0.08 & 0.12 & 0.112 & 0.14 & 0.023 & 0.03 & 0.044\\
\cmidrule{1-11}\pagebreak[0]
\num{5.00e+04} & PPV 95\% & 0.084 & 0.065 & 0.084 & 0.15 & 0.117 & 0.146 & 0.044 & 0.033 & 0.046\\
\cmidrule{1-11}\pagebreak[0]
\num{1.00e+05} & PPV 95\% & 0.088 & 0.066 & 0.085 & 0.153 & 0.118 & 0.148 & 0.048 & 0.034 & 0.047\\
\cmidrule{1-11}\pagebreak[0]
\num{1.00e+06} & PPV 95\% & 0.091 & 0.068 & 0.087 & 0.156 & 0.119 & 0.15 & 0.051 & 0.035 & 0.048\\
\cmidrule{1-11}\pagebreak[0]
\num{5.00e+03} & Sensitivity 90\% & 0.29 & 0.47 & 0.539 & 0.368 & 0.458 & 0.516 & 0.409 & 0.438 & 0.552\\
\cmidrule{1-11}\pagebreak[0]
\num{1.00e+04} & Sensitivity 90\% & 0.405 & 0.481 & 0.548 & 0.463 & 0.461 & 0.527 & 0.305 & 0.469 & 0.561\\
\cmidrule{1-11}\pagebreak[0]
\num{5.00e+04} & Sensitivity 90\% & 0.575 & 0.502 & 0.572 & 0.564 & 0.478 & 0.547 & 0.565 & 0.5 & 0.587\\
\cmidrule{1-11}\pagebreak[0]
\num{1.00e+05} & Sensitivity 90\% & 0.596 & 0.511 & 0.578 & 0.575 & 0.489 & 0.553 & 0.605 & 0.507 & 0.596\\
\cmidrule{1-11}\pagebreak[0]
\num{1.00e+06} & Sensitivity 90\% & 0.613 & 0.524 & 0.587 & 0.584 & 0.496 & 0.558 & 0.636 & 0.525 & 0.608\\
\cmidrule{1-11}\pagebreak[0]
\num{5.00e+03} & Sensitivity 99\% & 0.113 & 0.128 & 0.221 & 0.132 & 0.118 & 0.189 & 0.103 & 0.107 & 0.234\\
\cmidrule{1-11}\pagebreak[0]
\num{1.00e+04} & Sensitivity 99\% & 0.156 & 0.14 & 0.231 & 0.165 & 0.126 & 0.198 & 0.117 & 0.129 & 0.249\\
\cmidrule{1-11}\pagebreak[0]
\num{5.00e+04} & Sensitivity 99\% & 0.242 & 0.147 & 0.249 & 0.213 & 0.13 & 0.211 & 0.248 & 0.148 & 0.271\\
\cmidrule{1-11}\pagebreak[0]
\num{1.00e+05} & Sensitivity 99\% & 0.257 & 0.151 & 0.254 & 0.221 & 0.133 & 0.215 & 0.276 & 0.15 & 0.279\\
\cmidrule{1-11}\pagebreak[0]
\num{1.00e+06} & Sensitivity 99\% & 0.271 & 0.152 & 0.261 & 0.228 & 0.133 & 0.22 & 0.302 & 0.155 & 0.291\\
\cmidrule{1-11}\pagebreak[0]
\num{5.00e+03} & Specificity 90\% & 0.902 & 0.903 & 0.903 & 0.906 & 0.906 & 0.907 & 0.679 & 0.896 & 0.902\\
\cmidrule{1-11}\pagebreak[0]
\num{1.00e+04} & Specificity 90\% & 0.904 & 0.903 & 0.904 & 0.907 & 0.906 & 0.907 & 0.902 & 0.901 & 0.902\\
\cmidrule{1-11}\pagebreak[0]
\num{5.00e+04} & Specificity 90\% & 0.904 & 0.904 & 0.904 & 0.908 & 0.907 & 0.908 & 0.902 & 0.901 & 0.902\\
\cmidrule{1-11}\pagebreak[0]
\num{1.00e+05} & Specificity 90\% & 0.905 & 0.904 & 0.904 & 0.908 & 0.907 & 0.908 & 0.902 & 0.901 & 0.902\\
\cmidrule{1-11}\pagebreak[0]
\num{1.00e+06} & Specificity 90\% & 0.905 & 0.904 & 0.904 & 0.908 & 0.907 & 0.908 & 0.903 & 0.902 & 0.902\\
\cmidrule{1-11}\pagebreak[0]
\num{5.00e+03} & Specificity 99\% & 0.988 & 0.991 & 0.992 & 0.99 & 0.992 & 0.993 & 0.982 & 0.99 & 0.991\\
\cmidrule{1-11}\pagebreak[0]
\num{1.00e+04} & Specificity 99\% & 0.991 & 0.991 & 0.992 & 0.992 & 0.992 & 0.993 & 0.99 & 0.99 & 0.991\\
\cmidrule{1-11}\pagebreak[0]
\num{5.00e+04} & Specificity 99\% & 0.992 & 0.991 & 0.992 & 0.993 & 0.992 & 0.993 & 0.991 & 0.991 & 0.991\\
\cmidrule{1-11}\pagebreak[0]
\num{1.00e+05} & Specificity 99\% & 0.992 & 0.991 & 0.992 & 0.994 & 0.992 & 0.994 & 0.991 & 0.991 & 0.991\\
\cmidrule{1-11}\pagebreak[0]
\num{1.00e+06} & Specificity 99\% & 0.992 & 0.991 & 0.992 & 0.994 & 0.992 & 0.994 & 0.991 & 0.991 & 0.991\\
\cmidrule{1-11}\pagebreak[0]
\num{5.00e+03} & PPV 90\% & 0.027 & 0.043 & 0.049 & 0.064 & 0.08 & 0.088 & 0.009 & 0.02 & 0.027\\
\cmidrule{1-11}\pagebreak[0]
\num{1.00e+04} & PPV 90\% & 0.038 & 0.044 & 0.05 & 0.08 & 0.081 & 0.09 & 0.015 & 0.022 & 0.027\\
\cmidrule{1-11}\pagebreak[0]
\num{5.00e+04} & PPV 90\% & 0.053 & 0.046 & 0.052 & 0.096 & 0.084 & 0.093 & 0.028 & 0.023 & 0.028\\
\cmidrule{1-11}\pagebreak[0]
\num{1.00e+05} & PPV 90\% & 0.055 & 0.047 & 0.053 & 0.098 & 0.086 & 0.094 & 0.029 & 0.024 & 0.029\\
\cmidrule{1-11}\pagebreak[0]
\num{1.00e+06} & PPV 90\% & 0.056 & 0.048 & 0.054 & 0.099 & 0.087 & 0.095 & 0.031 & 0.025 & 0.03\\
\cmidrule{1-11}\pagebreak[0]
\num{5.00e+03} & PPV 99\% & 0.081 & 0.118 & 0.204 & 0.192 & 0.21 & 0.327 & 0.028 & 0.054 & 0.114\\
\cmidrule{1-11}\pagebreak[0]
\num{1.00e+04} & PPV 99\% & 0.135 & 0.128 & 0.213 & 0.269 & 0.221 & 0.34 & 0.053 & 0.061 & 0.121\\
\cmidrule{1-11}\pagebreak[0]
\num{5.00e+04} & PPV 99\% & 0.221 & 0.136 & 0.228 & 0.362 & 0.229 & 0.36 & 0.121 & 0.07 & 0.132\\
\cmidrule{1-11}\pagebreak[0]
\num{1.00e+05} & PPV 99\% & 0.235 & 0.138 & 0.232 & 0.375 & 0.233 & 0.367 & 0.134 & 0.071 & 0.136\\
\cmidrule{1-11}\pagebreak[0]
\num{1.00e+06} & PPV 99\% & 0.248 & 0.14 & 0.239 & 0.387 & 0.234 & 0.375 & 0.147 & 0.073 & 0.141\\
\cmidrule{1-11}\pagebreak[0]
\num{5.00e+03} & F0.5 90\% & 0.033 & 0.052 & 0.06 & 0.076 & 0.096 & 0.105 & 0.011 & 0.024 & 0.033\\
\cmidrule{1-11}\pagebreak[0]
\num{1.00e+04} & F0.5 90\% & 0.046 & 0.054 & 0.061 & 0.095 & 0.097 & 0.108 & 0.019 & 0.027 & 0.034\\
\cmidrule{1-11}\pagebreak[0]
\num{5.00e+04} & F0.5 90\% & 0.064 & 0.056 & 0.064 & 0.115 & 0.1 & 0.112 & 0.034 & 0.029 & 0.035\\
\cmidrule{1-11}\pagebreak[0]
\num{1.00e+05} & F0.5 90\% & 0.067 & 0.057 & 0.065 & 0.117 & 0.103 & 0.113 & 0.036 & 0.029 & 0.036\\
\cmidrule{1-11}\pagebreak[0]
\num{1.00e+06} & F0.5 90\% & 0.069 & 0.059 & 0.066 & 0.119 & 0.104 & 0.114 & 0.038 & 0.031 & 0.036\\
\cmidrule{1-11}\pagebreak[0]
\num{5.00e+03} & F0.5 95\% & 0.047 & 0.071 & 0.093 & 0.107 & 0.126 & 0.156 & 0.016 & 0.033 & 0.052\\
\cmidrule{1-11}\pagebreak[0]
\num{1.00e+04} & F0.5 95\% & 0.069 & 0.073 & 0.096 & 0.139 & 0.129 & 0.161 & 0.028 & 0.037 & 0.054\\
\cmidrule{1-11}\pagebreak[0]
\num{5.00e+04} & F0.5 95\% & 0.1 & 0.078 & 0.101 & 0.172 & 0.134 & 0.168 & 0.054 & 0.04 & 0.057\\
\cmidrule{1-11}\pagebreak[0]
\num{1.00e+05} & F0.5 95\% & 0.105 & 0.079 & 0.102 & 0.177 & 0.136 & 0.171 & 0.058 & 0.041 & 0.058\\
\cmidrule{1-11}\pagebreak[0]
\num{1.00e+06} & F0.5 95\% & 0.108 & 0.081 & 0.104 & 0.18 & 0.137 & 0.173 & 0.062 & 0.043 & 0.059\\
\cmidrule{1-11}\pagebreak[0]
\num{5.00e+03} & F0.5 99\% & 0.085 & 0.12 & 0.207 & 0.175 & 0.181 & 0.284 & 0.033 & 0.06 & 0.127\\
\cmidrule{1-11}\pagebreak[0]
\num{1.00e+04} & F0.5 99\% & 0.139 & 0.13 & 0.216 & 0.238 & 0.192 & 0.296 & 0.06 & 0.068 & 0.135\\
\cmidrule{1-11}\pagebreak[0]
\num{5.00e+04} & F0.5 99\% & 0.225 & 0.138 & 0.232 & 0.318 & 0.199 & 0.315 & 0.135 & 0.078 & 0.147\\
\cmidrule{1-11}\pagebreak[0]
\num{1.00e+05} & F0.5 99\% & 0.239 & 0.141 & 0.236 & 0.329 & 0.202 & 0.322 & 0.149 & 0.079 & 0.151\\
\cmidrule{1-11}\pagebreak[0]
\num{1.00e+06} & F0.5 99\% & 0.252 & 0.142 & 0.243 & 0.34 & 0.203 & 0.328 & 0.164 & 0.082 & 0.157\\
\cmidrule{1-11}\pagebreak[0]
\num{5.00e+03} & F1 90\% & 0.049 & 0.078 & 0.09 & 0.109 & 0.136 & 0.15 & 0.018 & 0.038 & 0.051\\
\cmidrule{1-11}\pagebreak[0]
\num{1.00e+04} & F1 90\% & 0.069 & 0.081 & 0.092 & 0.136 & 0.137 & 0.153 & 0.029 & 0.042 & 0.052\\
\cmidrule{1-11}\pagebreak[0]
\num{5.00e+04} & F1 90\% & 0.097 & 0.084 & 0.096 & 0.164 & 0.143 & 0.159 & 0.052 & 0.045 & 0.054\\
\cmidrule{1-11}\pagebreak[0]
\num{1.00e+05} & F1 90\% & 0.1 & 0.086 & 0.097 & 0.167 & 0.146 & 0.161 & 0.056 & 0.045 & 0.055\\
\cmidrule{1-11}\pagebreak[0]
\num{1.00e+06} & F1 90\% & 0.103 & 0.088 & 0.098 & 0.17 & 0.148 & 0.163 & 0.059 & 0.047 & 0.056\\
\cmidrule{1-11}\pagebreak[0]
\num{5.00e+03} & F1 95\% & 0.066 & 0.1 & 0.132 & 0.139 & 0.162 & 0.203 & 0.024 & 0.05 & 0.078\\
\cmidrule{1-11}\pagebreak[0]
\num{1.00e+04} & F1 95\% & 0.097 & 0.104 & 0.135 & 0.179 & 0.166 & 0.209 & 0.041 & 0.056 & 0.08\\
\cmidrule{1-11}\pagebreak[0]
\num{5.00e+04} & F1 95\% & 0.142 & 0.11 & 0.142 & 0.223 & 0.173 & 0.218 & 0.081 & 0.061 & 0.085\\
\cmidrule{1-11}\pagebreak[0]
\num{1.00e+05} & F1 95\% & 0.148 & 0.111 & 0.144 & 0.229 & 0.175 & 0.221 & 0.087 & 0.061 & 0.086\\
\cmidrule{1-11}\pagebreak[0]
\num{1.00e+06} & F1 95\% & 0.153 & 0.114 & 0.147 & 0.233 & 0.177 & 0.224 & 0.092 & 0.064 & 0.088\\
\cmidrule{1-11}\pagebreak[0]
\num{5.00e+03} & F1 99\% & 0.094 & 0.123 & 0.211 & 0.156 & 0.15 & 0.239 & 0.044 & 0.071 & 0.153\\
\cmidrule{1-11}\pagebreak[0]
\num{1.00e+04} & F1 99\% & 0.144 & 0.133 & 0.221 & 0.204 & 0.16 & 0.249 & 0.073 & 0.082 & 0.163\\
\cmidrule{1-11}\pagebreak[0]
\num{5.00e+04} & F1 99\% & 0.231 & 0.141 & 0.238 & 0.268 & 0.166 & 0.266 & 0.162 & 0.095 & 0.177\\
\cmidrule{1-11}\pagebreak[0]
\num{1.00e+05} & F1 99\% & 0.245 & 0.144 & 0.242 & 0.278 & 0.169 & 0.272 & 0.18 & 0.096 & 0.182\\
\cmidrule{1-11}\pagebreak[0]
\num{1.00e+06} & F1 99\% & 0.259 & 0.146 & 0.25 & 0.287 & 0.17 & 0.277 & 0.198 & 0.099 & 0.19\\
\cmidrule{1-11}\pagebreak[0]
\num{5.00e+03} & NPV 90\% & 0.993 & 0.995 & 0.995 & 0.988 & 0.989 & 0.991 & 0.996 & 0.997 & 0.998\\
\cmidrule{1-11}\pagebreak[0]
\num{1.00e+04} & NPV 90\% & 0.994 & 0.995 & 0.995 & 0.99 & 0.989 & 0.991 & 0.996 & 0.997 & 0.998\\
\cmidrule{1-11}\pagebreak[0]
\num{5.00e+04} & NPV 90\% & 0.996 & 0.995 & 0.996 & 0.992 & 0.99 & 0.991 & 0.998 & 0.997 & 0.998\\
\cmidrule{1-11}\pagebreak[0]
\num{1.00e+05} & NPV 90\% & 0.996 & 0.995 & 0.996 & 0.992 & 0.99 & 0.992 & 0.998 & 0.997 & 0.998\\
\cmidrule{1-11}\pagebreak[0]
\num{1.00e+06} & NPV 90\% & 0.996 & 0.995 & 0.996 & 0.992 & 0.99 & 0.992 & 0.998 & 0.998 & 0.998\\
\cmidrule{1-11}\pagebreak[0]
\num{5.00e+03} & NPV 95\% & 0.992 & 0.993 & 0.994 & 0.987 & 0.987 & 0.989 & 0.996 & 0.997 & 0.997\\
\cmidrule{1-11}\pagebreak[0]
\num{1.00e+04} & NPV 95\% & 0.993 & 0.994 & 0.995 & 0.988 & 0.987 & 0.989 & 0.996 & 0.997 & 0.997\\
\cmidrule{1-11}\pagebreak[0]
\num{5.00e+04} & NPV 95\% & 0.995 & 0.994 & 0.995 & 0.99 & 0.988 & 0.99 & 0.997 & 0.997 & 0.997\\
\cmidrule{1-11}\pagebreak[0]
\num{1.00e+05} & NPV 95\% & 0.995 & 0.994 & 0.995 & 0.99 & 0.988 & 0.99 & 0.997 & 0.997 & 0.997\\
\cmidrule{1-11}\pagebreak[0]
\num{1.00e+06} & NPV 95\% & 0.995 & 0.994 & 0.995 & 0.99 & 0.988 & 0.99 & 0.998 & 0.997 & 0.997\\
\cmidrule{1-11}\pagebreak[0]
\num{5.00e+03} & NPV 99\% & 0.992 & 0.992 & 0.993 & 0.985 & 0.984 & 0.986 & 0.996 & 0.996 & 0.996\\
\cmidrule{1-11}\pagebreak[0]
\num{1.00e+04} & NPV 99\% & 0.992 & 0.992 & 0.993 & 0.986 & 0.984 & 0.986 & 0.996 & 0.996 & 0.996\\
\cmidrule{1-11}\pagebreak[0]
\num{5.00e+04} & NPV 99\% & 0.993 & 0.992 & 0.993 & 0.986 & 0.985 & 0.986 & 0.996 & 0.996 & 0.996\\
\cmidrule{1-11}\pagebreak[0]
\num{1.00e+05} & NPV 99\% & 0.993 & 0.992 & 0.993 & 0.987 & 0.985 & 0.986 & 0.996 & 0.996 & 0.996\\
\cmidrule{1-11}\pagebreak[0]
\num{1.00e+06} & NPV 99\% & 0.993 & 0.992 & 0.993 & 0.987 & 0.985 & 0.987 & 0.997 & 0.996 & 0.997\\
\cmidrule{1-11}\pagebreak[0]
\num{5.00e+03} & Accuracy 90\% & 0.897 & 0.899 & 0.9 & 0.897 & 0.898 & 0.9 & 0.677 & 0.894 & 0.9\\
\cmidrule{1-11}\pagebreak[0]
\num{1.00e+04} & Accuracy 90\% & 0.899 & 0.899 & 0.901 & 0.9 & 0.898 & 0.901 & 0.9 & 0.899 & 0.9\\
\cmidrule{1-11}\pagebreak[0]
\num{5.00e+04} & Accuracy 90\% & 0.901 & 0.9 & 0.901 & 0.902 & 0.899 & 0.902 & 0.901 & 0.9 & 0.901\\
\cmidrule{1-11}\pagebreak[0]
\num{1.00e+05} & Accuracy 90\% & 0.902 & 0.9 & 0.901 & 0.903 & 0.9 & 0.902 & 0.901 & 0.9 & 0.901\\
\cmidrule{1-11}\pagebreak[0]
\num{1.00e+06} & Accuracy 90\% & 0.902 & 0.9 & 0.902 & 0.903 & 0.9 & 0.902 & 0.901 & 0.9 & 0.901\\
\cmidrule{1-11}\pagebreak[0]
\num{5.00e+03} & Accuracy 95\% & 0.944 & 0.946 & 0.948 & 0.943 & 0.943 & 0.946 & 0.756 & 0.944 & 0.949\\
\cmidrule{1-11}\pagebreak[0]
\num{1.00e+04} & Accuracy 95\% & 0.947 & 0.947 & 0.949 & 0.945 & 0.944 & 0.947 & 0.948 & 0.948 & 0.949\\
\cmidrule{1-11}\pagebreak[0]
\num{5.00e+04} & Accuracy 95\% & 0.949 & 0.947 & 0.949 & 0.948 & 0.944 & 0.948 & 0.95 & 0.948 & 0.95\\
\cmidrule{1-11}\pagebreak[0]
\num{1.00e+05} & Accuracy 95\% & 0.95 & 0.947 & 0.949 & 0.948 & 0.944 & 0.948 & 0.95 & 0.948 & 0.95\\
\cmidrule{1-11}\pagebreak[0]
\num{1.00e+06} & Accuracy 95\% & 0.95 & 0.948 & 0.95 & 0.949 & 0.944 & 0.948 & 0.95 & 0.949 & 0.95\\
\cmidrule{1-11}\pagebreak[0]
\num{5.00e+03} & Accuracy 99\% & 0.98 & 0.983 & 0.985 & 0.976 & 0.977 & 0.979 & 0.978 & 0.986 & 0.987\\
\cmidrule{1-11}\pagebreak[0]
\num{1.00e+04} & Accuracy 99\% & 0.983 & 0.983 & 0.985 & 0.978 & 0.977 & 0.98 & 0.986 & 0.986 & 0.988\\
\cmidrule{1-11}\pagebreak[0]
\num{5.00e+04} & Accuracy 99\% & 0.985 & 0.984 & 0.985 & 0.98 & 0.977 & 0.98 & 0.988 & 0.987 & 0.988\\
\cmidrule{1-11}\pagebreak[0]
\num{1.00e+05} & Accuracy 99\% & 0.986 & 0.984 & 0.985 & 0.98 & 0.977 & 0.98 & 0.988 & 0.987 & 0.988\\
\cmidrule{1-11}\pagebreak[0]
\num{1.00e+06} & Accuracy 99\% & 0.986 & 0.984 & 0.986 & 0.981 & 0.977 & 0.98 & 0.988 & 0.987 & 0.988\\
\cmidrule{1-11}\pagebreak[0]
\num{5.00e+03} & Brier score & 0.013 & 0.009 & 0.009 & 0.019 & 0.017 & 0.015 & 0.015 & 0.009 & 0.005\\
\cmidrule{1-11}\pagebreak[0]
\num{1.00e+04} & Brier score & 0.01 & 0.009 & 0.008 & 0.017 & 0.017 & 0.015 & 0.006 & 0.011 & 0.005\\
\cmidrule{1-11}\pagebreak[0]
\num{5.00e+04} & Brier score & 0.008 & 0.009 & 0.008 & 0.015 & 0.016 & 0.015 & 0.005 & 0.01 & 0.005\\
\cmidrule{1-11}\pagebreak[0]
\num{1.00e+05} & Brier score & 0.008 & 0.009 & 0.008 & 0.015 & 0.016 & 0.015 & 0.005 & 0.01 & 0.005\\
\cmidrule{1-11}\pagebreak[0]
\num{1.00e+06} & Brier score & 0.008 & 0.009 & 0.008 & 0.015 & 0.016 & 0.015 & 0.004 & 0.005 & 0.004\\*
\end{longtable}
\end{ThreePartTable}
\endgroup{}

\subsection{Behavior of AUC}

\begin{figure}
\centering
\includegraphics[width=1\textwidth]{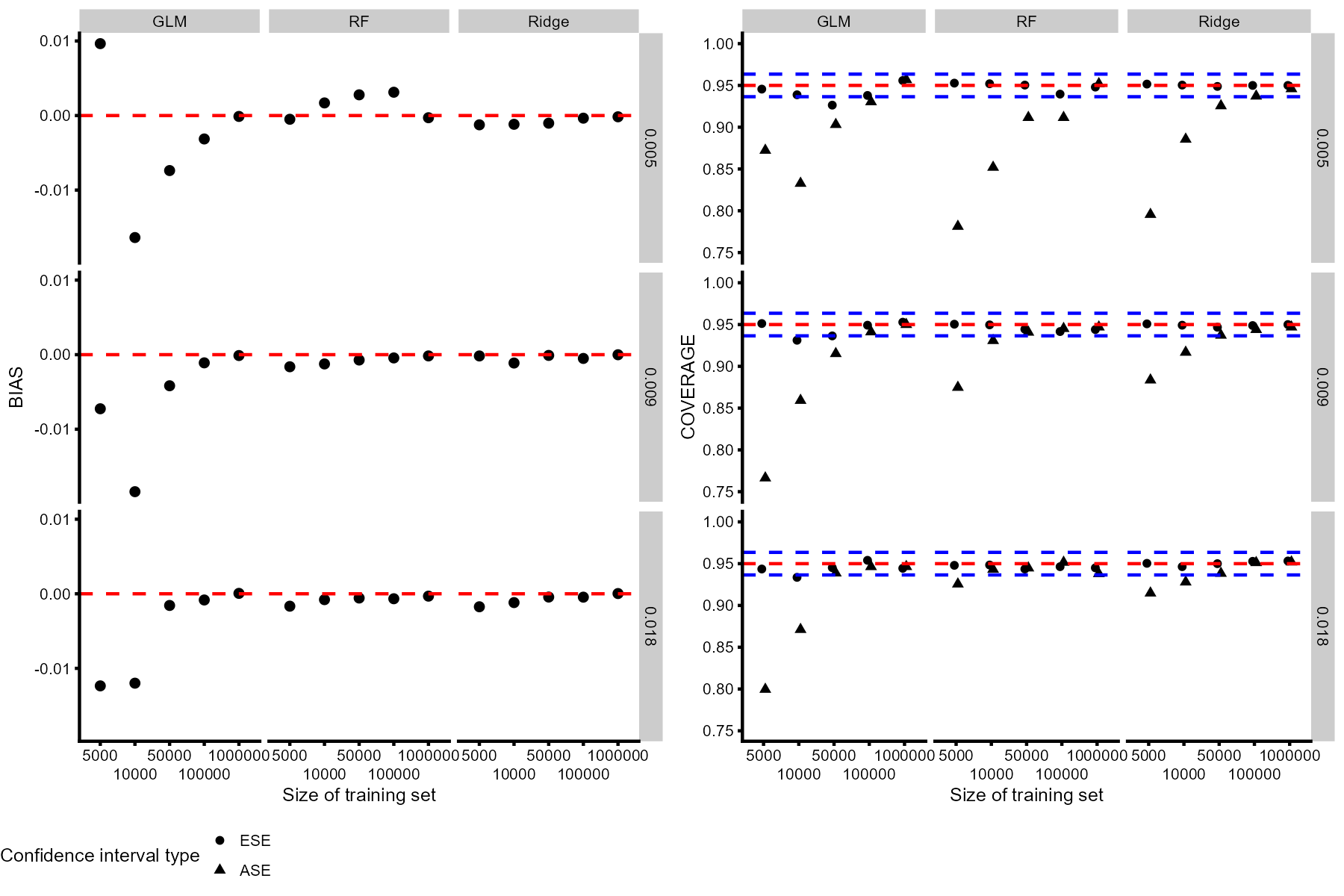}
\caption{Empirical bias and coverage of 95\% confidence intervals for estimating the evaluation-set AUC (values provided in Table~3) versus training set size at the three event rates (rows); columns show logistic regression (GLM) including all predictors, random forests (RF), and ridge logistic regression (Ridge) including all predictors. ESE = empirical standard error, ASE = asymptotic standard error. The blue dashed lines around 95\% coverage indicate one Monte-Carlo standard error.}
\label{fig:auc_vs_n_supp}
\end{figure}

In Figure~\ref{fig:auc_vs_n_supp}, we show bias and coverage for estimating AUC versus training set size. The rows are the event rates, while the columns are the algorithms. We see that the trends in bias and coverage do not depend on event rate, but instead depend on the number of events.

\begin{figure}
\centering
\includegraphics[width=1\textwidth]{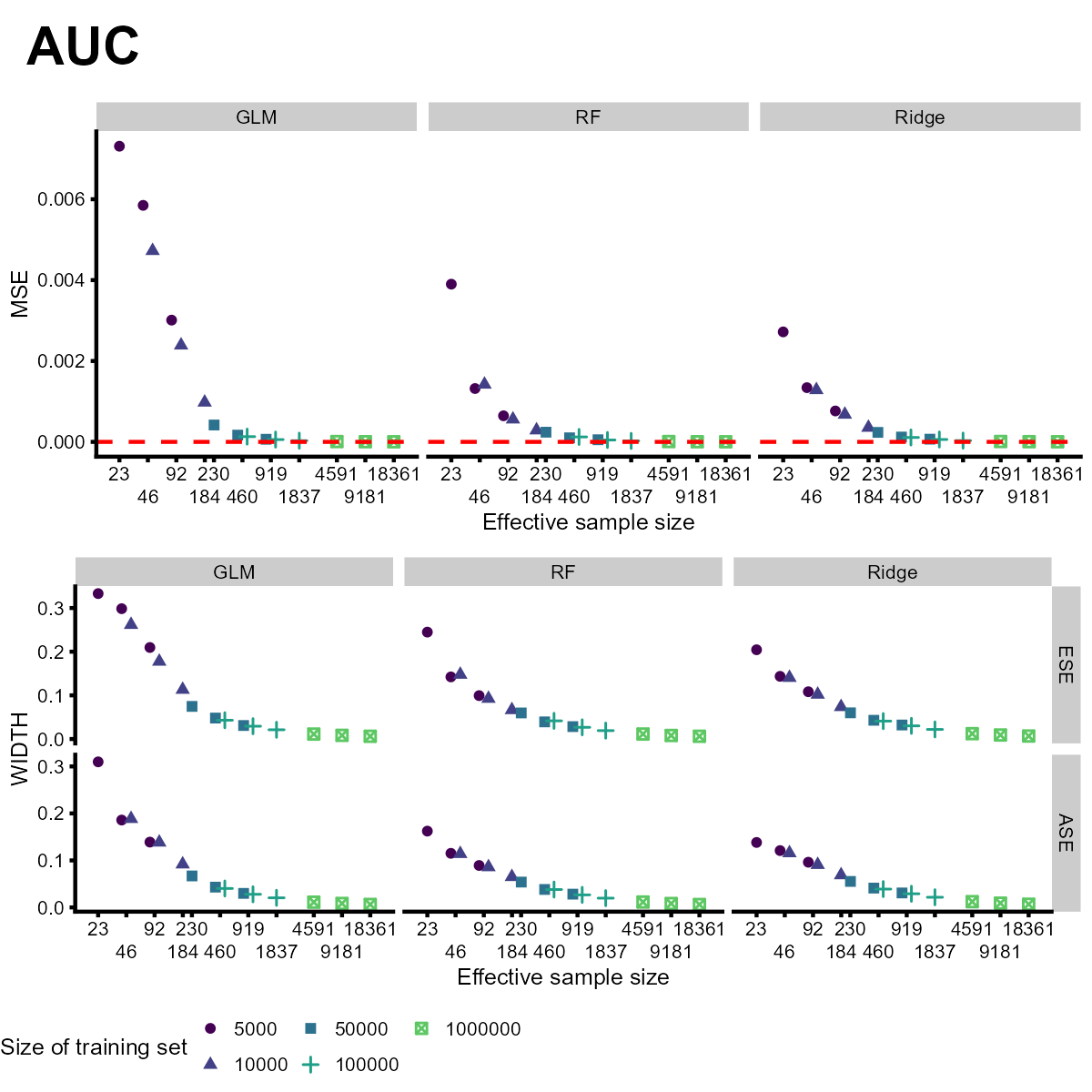}
\caption{Empirical mean squared error and 95\% confidence interval width for estimating the evaluation-set AUC (values provided in Table~3) versus effective sample size; columns show logistic regression (GLM) including all predictors, random forests (RF), and ridge logistic regression (Ridge) including all predictors. ESE = empirical standard error, ASE = asymptotic standard error. The blue dashed lines around 95\% coverage indicate one Monte-Carlo standard error.}
\label{fig:auc_supp}
\end{figure}

In Figure~\ref{fig:auc_supp}, we show the empirical mean squared error and confidence interval width for estimating AUC.

\subsection{Behavior of other prediction metrics}
We present behavior for sensitivity, specificity, and PPV at the 90th and 99th percentiles of predicted risk; accuracy, NPV, F$_{1}$, and F$_{0.5}$ at the 90th, 95th, and 99th percentiles of predicted risk;  and the Brier score in Tables~\ref{tab:sens_90_test}--\ref{tab:brier_test}.

\begin{table}
\centering
\caption{Results for Sensitivity 90\% (percentile of predicted risk), measured with respect to mean evaluation-set performance. Bias and mean squared error (MSE) are truncated at 0.001; coverage and width are computed for intervals based on the empirical standard error (ESE). All algorithms (`GLM' = logistic regression, `RF' = random forests, `Ridge' = ridge regression), event rates, and training sample sizes are shown.\label{tab:sens_90_test}}
\centering
\resizebox{\ifdim\width>\linewidth\linewidth\else\width\fi}{!}{
\fontsize{9}{11}\selectfont
\begin{tabular}[t]{l>{\raggedright\arraybackslash}p{5em}lrll>{\raggedleft\arraybackslash}p{5em}>{\raggedleft\arraybackslash}p{5em}}
\toprule
Algorithm & Event rate & N & N event & Bias & MSE & Coverage (ESE) & Width (ESE)\\
\midrule
GLM & \num{4.59e-03} & \num{5e+03} & 23 & 0.045 & 0.065 & 0.947 & 1.131\\
GLM & \num{4.59e-03} & \num{1e+04} & 46 & -0.02 & 0.007 & 0.940 & 0.307\\
GLM & \num{4.59e-03} & \num{5e+04} & 230 & -0.008 & 0.001 & 0.944 & 0.136\\
GLM & \num{4.59e-03} & \num{1e+05} & 460 & -0.004 & $<$ 0.001 & 0.948 & 0.090\\
GLM & \num{4.59e-03} & \num{1e+06} & 4591 & $<$ 0.001 & $<$ 0.001 & 0.944 & 0.028\\
GLM & \num{9.18e-03} & \num{5e+03} & 46 & -0.013 & 0.008 & 0.959 & 0.351\\
GLM & \num{9.18e-03} & \num{1e+04} & 92 & -0.019 & 0.004 & 0.938 & 0.235\\
GLM & \num{9.18e-03} & \num{5e+04} & 460 & -0.005 & $<$ 0.001 & 0.945 & 0.099\\
GLM & \num{9.18e-03} & \num{1e+05} & 919 & -0.001 & $<$ 0.001 & 0.952 & 0.065\\
GLM & \num{9.18e-03} & \num{1e+06} & 9181 & $<$ 0.001 & $<$ 0.001 & 0.954 & 0.020\\
GLM & \num{1.84e-02} & \num{5e+03} & 92 & -0.014 & 0.004 & 0.947 & 0.243\\
GLM & \num{1.84e-02} & \num{1e+04} & 184 & -0.011 & 0.002 & 0.946 & 0.172\\
GLM & \num{1.84e-02} & \num{5e+04} & 919 & -0.002 & $<$ 0.001 & 0.941 & 0.067\\
GLM & \num{1.84e-02} & \num{1e+05} & 1837 & -0.001 & $<$ 0.001 & 0.952 & 0.047\\
GLM & \num{1.84e-02} & \num{1e+06} & 18361 & $<$ 0.001 & $<$ 0.001 & 0.945 & 0.015\\
RF & \num{4.59e-03} & \num{5e+03} & 23 & -0.006 & 0.014 & 0.951 & 0.487\\
RF & \num{4.59e-03} & \num{1e+04} & 46 & $<$ 0.001 & 0.008 & 0.947 & 0.319\\
RF & \num{4.59e-03} & \num{5e+04} & 230 & 0.002 & 0.002 & 0.950 & 0.134\\
RF & \num{4.59e-03} & \num{1e+05} & 460 & 0.002 & 0.002 & 0.948 & 0.097\\
RF & \num{4.59e-03} & \num{1e+06} & 4591 & $<$ 0.001 & $<$ 0.001 & 0.946 & 0.028\\
RF & \num{9.18e-03} & \num{5e+03} & 46 & -0.002 & 0.006 & 0.952 & 0.317\\
RF & \num{9.18e-03} & \num{1e+04} & 92 & -0.002 & 0.003 & 0.950 & 0.210\\
RF & \num{9.18e-03} & \num{5e+04} & 460 & $<$ 0.001 & $<$ 0.001 & 0.948 & 0.089\\
RF & \num{9.18e-03} & \num{1e+05} & 919 & -0.001 & $<$ 0.001 & 0.952 & 0.064\\
RF & \num{9.18e-03} & \num{1e+06} & 9181 & $<$ 0.001 & $<$ 0.001 & 0.944 & 0.019\\
RF & \num{1.84e-02} & \num{5e+03} & 92 & -0.002 & 0.003 & 0.953 & 0.216\\
RF & \num{1.84e-02} & \num{1e+04} & 184 & $<$ 0.001 & 0.002 & 0.949 & 0.149\\
RF & \num{1.84e-02} & \num{5e+04} & 919 & $<$ 0.001 & $<$ 0.001 & 0.950 & 0.063\\
RF & \num{1.84e-02} & \num{1e+05} & 1837 & -0.002 & $<$ 0.001 & 0.939 & 0.045\\
RF & \num{1.84e-02} & \num{1e+06} & 18361 & $<$ 0.001 & $<$ 0.001 & 0.952 & 0.015\\
Ridge & \num{4.59e-03} & \num{5e+03} & 23 & $<$ 0.001 & 0.012 & 0.950 & 0.426\\
Ridge & \num{4.59e-03} & \num{1e+04} & 46 & -0.002 & 0.005 & 0.946 & 0.287\\
Ridge & \num{4.59e-03} & \num{5e+04} & 230 & -0.003 & 0.001 & 0.952 & 0.128\\
Ridge & \num{4.59e-03} & \num{1e+05} & 460 & $<$ 0.001 & $<$ 0.001 & 0.949 & 0.088\\
Ridge & \num{4.59e-03} & \num{1e+06} & 4591 & $<$ 0.001 & $<$ 0.001 & 0.952 & 0.028\\
Ridge & \num{9.18e-03} & \num{5e+03} & 46 & 0.001 & 0.006 & 0.947 & 0.301\\
Ridge & \num{9.18e-03} & \num{1e+04} & 92 & -0.001 & 0.003 & 0.952 & 0.206\\
Ridge & \num{9.18e-03} & \num{5e+04} & 460 & $<$ 0.001 & $<$ 0.001 & 0.950 & 0.091\\
Ridge & \num{9.18e-03} & \num{1e+05} & 919 & $<$ 0.001 & $<$ 0.001 & 0.950 & 0.063\\
Ridge & \num{9.18e-03} & \num{1e+06} & 9181 & $<$ 0.001 & $<$ 0.001 & 0.952 & 0.020\\
Ridge & \num{1.84e-02} & \num{5e+03} & 92 & -0.003 & 0.003 & 0.947 & 0.214\\
Ridge & \num{1.84e-02} & \num{1e+04} & 184 & -0.002 & 0.002 & 0.946 & 0.152\\
Ridge & \num{1.84e-02} & \num{5e+04} & 919 & $<$ 0.001 & $<$ 0.001 & 0.950 & 0.067\\
Ridge & \num{1.84e-02} & \num{1e+05} & 1837 & -0.001 & $<$ 0.001 & 0.950 & 0.046\\
Ridge & \num{1.84e-02} & \num{1e+06} & 18361 & $<$ 0.001 & $<$ 0.001 & 0.953 & 0.015\\
\bottomrule
\end{tabular}}
\end{table}

\begin{table}
\centering
\caption{Results for Sensitivity 99\% (percentile of predicted risk), measured with respect to mean evaluation-set performance. Bias and mean squared error (MSE) are truncated at 0.001; coverage and width are computed for intervals based on the empirical standard error (ESE). All algorithms (`GLM' = logistic regression, `RF' = random forests, `Ridge' = ridge regression), event rates, and training sample sizes are shown.\label{tab:sens_99_test}}
\centering
\resizebox{\ifdim\width>\linewidth\linewidth\else\width\fi}{!}{
\fontsize{9}{11}\selectfont
\begin{tabular}[t]{l>{\raggedright\arraybackslash}p{5em}lrll>{\raggedleft\arraybackslash}p{5em}>{\raggedleft\arraybackslash}p{5em}}
\toprule
Algorithm & Event rate & N & N event & Bias & MSE & Coverage (ESE) & Width (ESE)\\
\midrule
GLM & \num{4.59e-03} & \num{5e+03} & 23 & 0.009 & 0.005 & 0.958 & 0.301\\
GLM & \num{4.59e-03} & \num{1e+04} & 46 & -0.008 & 0.003 & 0.952 & 0.193\\
GLM & \num{4.59e-03} & \num{5e+04} & 230 & -0.006 & $<$ 0.001 & 0.946 & 0.116\\
GLM & \num{4.59e-03} & \num{1e+05} & 460 & -0.003 & $<$ 0.001 & 0.949 & 0.080\\
GLM & \num{4.59e-03} & \num{1e+06} & 4591 & $<$ 0.001 & $<$ 0.001 & 0.954 & 0.024\\
GLM & \num{9.18e-03} & \num{5e+03} & 46 & -0.003 & 0.003 & 0.954 & 0.205\\
GLM & \num{9.18e-03} & \num{1e+04} & 92 & -0.007 & 0.002 & 0.949 & 0.161\\
GLM & \num{9.18e-03} & \num{5e+04} & 460 & -0.003 & $<$ 0.001 & 0.945 & 0.080\\
GLM & \num{9.18e-03} & \num{1e+05} & 919 & -0.001 & $<$ 0.001 & 0.945 & 0.054\\
GLM & \num{9.18e-03} & \num{1e+06} & 9181 & $<$ 0.001 & $<$ 0.001 & 0.951 & 0.017\\
GLM & \num{1.84e-02} & \num{5e+03} & 92 & -0.004 & 0.002 & 0.954 & 0.154\\
GLM & \num{1.84e-02} & \num{1e+04} & 184 & -0.004 & 0.001 & 0.950 & 0.117\\
GLM & \num{1.84e-02} & \num{5e+04} & 919 & -0.002 & $<$ 0.001 & 0.946 & 0.050\\
GLM & \num{1.84e-02} & \num{1e+05} & 1837 & -0.001 & $<$ 0.001 & 0.946 & 0.034\\
GLM & \num{1.84e-02} & \num{1e+06} & 18361 & $<$ 0.001 & $<$ 0.001 & 0.951 & 0.010\\
RF & \num{4.59e-03} & \num{5e+03} & 23 & -0.001 & 0.005 & 0.958 & 0.317\\
RF & \num{4.59e-03} & \num{1e+04} & 46 & $<$ 0.001 & 0.003 & 0.966 & 0.234\\
RF & \num{4.59e-03} & \num{5e+04} & 230 & $<$ 0.001 & $<$ 0.001 & 0.957 & 0.097\\
RF & \num{4.59e-03} & \num{1e+05} & 460 & $<$ 0.001 & $<$ 0.001 & 0.950 & 0.068\\
RF & \num{4.59e-03} & \num{1e+06} & 4591 & $<$ 0.001 & $<$ 0.001 & 0.949 & 0.020\\
RF & \num{9.18e-03} & \num{5e+03} & 46 & -0.002 & 0.003 & 0.960 & 0.232\\
RF & \num{9.18e-03} & \num{1e+04} & 92 & -0.002 & 0.002 & 0.948 & 0.160\\
RF & \num{9.18e-03} & \num{5e+04} & 460 & $<$ 0.001 & $<$ 0.001 & 0.946 & 0.063\\
RF & \num{9.18e-03} & \num{1e+05} & 919 & $<$ 0.001 & $<$ 0.001 & 0.949 & 0.044\\
RF & \num{9.18e-03} & \num{1e+06} & 9181 & $<$ 0.001 & $<$ 0.001 & 0.945 & 0.014\\
RF & \num{1.84e-02} & \num{5e+03} & 92 & $<$ 0.001 & 0.001 & 0.956 & 0.148\\
RF & \num{1.84e-02} & \num{1e+04} & 184 & -0.001 & $<$ 0.001 & 0.950 & 0.102\\
RF & \num{1.84e-02} & \num{5e+04} & 919 & $<$ 0.001 & $<$ 0.001 & 0.949 & 0.041\\
RF & \num{1.84e-02} & \num{1e+05} & 1837 & $<$ 0.001 & $<$ 0.001 & 0.953 & 0.028\\
RF & \num{1.84e-02} & \num{1e+06} & 18361 & $<$ 0.001 & $<$ 0.001 & 0.952 & 0.009\\
Ridge & \num{4.59e-03} & \num{5e+03} & 23 & -0.004 & 0.008 & 0.962 & 0.369\\
Ridge & \num{4.59e-03} & \num{1e+04} & 46 & -0.003 & 0.004 & 0.952 & 0.257\\
Ridge & \num{4.59e-03} & \num{5e+04} & 230 & -0.002 & $<$ 0.001 & 0.951 & 0.112\\
Ridge & \num{4.59e-03} & \num{1e+05} & 460 & $<$ 0.001 & $<$ 0.001 & 0.948 & 0.080\\
Ridge & \num{4.59e-03} & \num{1e+06} & 4591 & $<$ 0.001 & $<$ 0.001 & 0.950 & 0.025\\
Ridge & \num{9.18e-03} & \num{5e+03} & 46 & -0.001 & 0.004 & 0.953 & 0.244\\
Ridge & \num{9.18e-03} & \num{1e+04} & 92 & -0.001 & 0.002 & 0.952 & 0.173\\
Ridge & \num{9.18e-03} & \num{5e+04} & 460 & $<$ 0.001 & $<$ 0.001 & 0.947 & 0.076\\
Ridge & \num{9.18e-03} & \num{1e+05} & 919 & -0.001 & $<$ 0.001 & 0.951 & 0.054\\
Ridge & \num{9.18e-03} & \num{1e+06} & 9181 & $<$ 0.001 & $<$ 0.001 & 0.952 & 0.016\\
Ridge & \num{1.84e-02} & \num{5e+03} & 92 & -0.003 & 0.002 & 0.952 & 0.159\\
Ridge & \num{1.84e-02} & \num{1e+04} & 184 & -0.001 & 0.001 & 0.955 & 0.113\\
Ridge & \num{1.84e-02} & \num{5e+04} & 919 & $<$ 0.001 & $<$ 0.001 & 0.953 & 0.049\\
Ridge & \num{1.84e-02} & \num{1e+05} & 1837 & $<$ 0.001 & $<$ 0.001 & 0.947 & 0.034\\
Ridge & \num{1.84e-02} & \num{1e+06} & 18361 & $<$ 0.001 & $<$ 0.001 & 0.952 & 0.011\\
\bottomrule
\end{tabular}}
\end{table}

\begin{table}
\centering
\caption{Results for Specificity 90\% (percentile of predicted risk), measured with respect to mean evaluation-set performance. Bias and mean squared error (MSE) are truncated at 0.001; coverage and width are computed for intervals based on the empirical standard error (ESE). All algorithms (`GLM' = logistic regression, `RF' = random forests, `Ridge' = ridge regression), event rates, and training sample sizes are shown.\label{tab:spec_90_test}}
\centering
\resizebox{\ifdim\width>\linewidth\linewidth\else\width\fi}{!}{
\fontsize{9}{11}\selectfont
\begin{tabular}[t]{l>{\raggedright\arraybackslash}p{5em}lrll>{\raggedleft\arraybackslash}p{5em}>{\raggedleft\arraybackslash}p{5em}}
\toprule
Algorithm & Event rate & N & N event & Bias & MSE & Coverage (ESE) & Width (ESE)\\
\midrule
GLM & \num{4.59e-03} & \num{5e+03} & 23 & -0.048 & 0.075 & 0.952 & 1.215\\
GLM & \num{4.59e-03} & \num{1e+04} & 46 & $<$ 0.001 & $<$ 0.001 & 0.998 & 0.034\\
GLM & \num{4.59e-03} & \num{5e+04} & 230 & $<$ 0.001 & $<$ 0.001 & 0.948 & 0.002\\
GLM & \num{4.59e-03} & \num{1e+05} & 460 & $<$ 0.001 & $<$ 0.001 & 0.951 & 0.001\\
GLM & \num{4.59e-03} & \num{1e+06} & 4591 & $<$ 0.001 & $<$ 0.001 & 0.950 & 0.000\\
GLM & \num{9.18e-03} & \num{5e+03} & 46 & -0.004 & 0.001 & 0.980 & 0.166\\
GLM & \num{9.18e-03} & \num{1e+04} & 92 & $<$ 0.001 & $<$ 0.001 & 0.949 & 0.007\\
GLM & \num{9.18e-03} & \num{5e+04} & 460 & $<$ 0.001 & $<$ 0.001 & 0.948 & 0.002\\
GLM & \num{9.18e-03} & \num{1e+05} & 919 & $<$ 0.001 & $<$ 0.001 & 0.952 & 0.001\\
GLM & \num{9.18e-03} & \num{1e+06} & 9181 & $<$ 0.001 & $<$ 0.001 & 0.928 & 0.000\\
GLM & \num{1.84e-02} & \num{5e+03} & 92 & $<$ 0.001 & $<$ 0.001 & 0.947 & 0.011\\
GLM & \num{1.84e-02} & \num{1e+04} & 184 & $<$ 0.001 & $<$ 0.001 & 0.949 & 0.007\\
GLM & \num{1.84e-02} & \num{5e+04} & 919 & $<$ 0.001 & $<$ 0.001 & 0.951 & 0.002\\
GLM & \num{1.84e-02} & \num{1e+05} & 1837 & $<$ 0.001 & $<$ 0.001 & 0.954 & 0.002\\
GLM & \num{1.84e-02} & \num{1e+06} & 18361 & $<$ 0.001 & $<$ 0.001 & 0.935 & 0.000\\
RF & \num{4.59e-03} & \num{5e+03} & 23 & $<$ 0.001 & $<$ 0.001 & 0.937 & 0.011\\
RF & \num{4.59e-03} & \num{1e+04} & 46 & $<$ 0.001 & $<$ 0.001 & 0.949 & 0.006\\
RF & \num{4.59e-03} & \num{5e+04} & 230 & $<$ 0.001 & $<$ 0.001 & 0.953 & 0.002\\
RF & \num{4.59e-03} & \num{1e+05} & 460 & $<$ 0.001 & $<$ 0.001 & 0.948 & 0.002\\
RF & \num{4.59e-03} & \num{1e+06} & 4591 & $<$ 0.001 & $<$ 0.001 & 0.946 & 0.000\\
RF & \num{9.18e-03} & \num{5e+03} & 46 & $<$ 0.001 & $<$ 0.001 & 0.946 & 0.008\\
RF & \num{9.18e-03} & \num{1e+04} & 92 & $<$ 0.001 & $<$ 0.001 & 0.948 & 0.005\\
RF & \num{9.18e-03} & \num{5e+04} & 460 & $<$ 0.001 & $<$ 0.001 & 0.951 & 0.002\\
RF & \num{9.18e-03} & \num{1e+05} & 919 & $<$ 0.001 & $<$ 0.001 & 0.955 & 0.001\\
RF & \num{9.18e-03} & \num{1e+06} & 9181 & $<$ 0.001 & $<$ 0.001 & 0.952 & 0.000\\
RF & \num{1.84e-02} & \num{5e+03} & 92 & $<$ 0.001 & $<$ 0.001 & 0.953 & 0.009\\
RF & \num{1.84e-02} & \num{1e+04} & 184 & $<$ 0.001 & $<$ 0.001 & 0.948 & 0.006\\
RF & \num{1.84e-02} & \num{5e+04} & 919 & $<$ 0.001 & $<$ 0.001 & 0.949 & 0.002\\
RF & \num{1.84e-02} & \num{1e+05} & 1837 & $<$ 0.001 & $<$ 0.001 & 0.945 & 0.002\\
RF & \num{1.84e-02} & \num{1e+06} & 18361 & $<$ 0.001 & $<$ 0.001 & 0.945 & 0.000\\
Ridge & \num{4.59e-03} & \num{5e+03} & 23 & $<$ 0.001 & $<$ 0.001 & 0.946 & 0.006\\
Ridge & \num{4.59e-03} & \num{1e+04} & 46 & $<$ 0.001 & $<$ 0.001 & 0.948 & 0.004\\
Ridge & \num{4.59e-03} & \num{5e+04} & 230 & $<$ 0.001 & $<$ 0.001 & 0.946 & 0.002\\
Ridge & \num{4.59e-03} & \num{1e+05} & 460 & $<$ 0.001 & $<$ 0.001 & 0.952 & 0.001\\
Ridge & \num{4.59e-03} & \num{1e+06} & 4591 & $<$ 0.001 & $<$ 0.001 & 0.939 & 0.000\\
Ridge & \num{9.18e-03} & \num{5e+03} & 46 & $<$ 0.001 & $<$ 0.001 & 0.948 & 0.006\\
Ridge & \num{9.18e-03} & \num{1e+04} & 92 & $<$ 0.001 & $<$ 0.001 & 0.944 & 0.005\\
Ridge & \num{9.18e-03} & \num{5e+04} & 460 & $<$ 0.001 & $<$ 0.001 & 0.949 & 0.002\\
Ridge & \num{9.18e-03} & \num{1e+05} & 919 & $<$ 0.001 & $<$ 0.001 & 0.954 & 0.001\\
Ridge & \num{9.18e-03} & \num{1e+06} & 9181 & $<$ 0.001 & $<$ 0.001 & 0.944 & 0.000\\
Ridge & \num{1.84e-02} & \num{5e+03} & 92 & $<$ 0.001 & $<$ 0.001 & 0.952 & 0.007\\
Ridge & \num{1.84e-02} & \num{1e+04} & 184 & $<$ 0.001 & $<$ 0.001 & 0.952 & 0.005\\
Ridge & \num{1.84e-02} & \num{5e+04} & 919 & $<$ 0.001 & $<$ 0.001 & 0.944 & 0.002\\
Ridge & \num{1.84e-02} & \num{1e+05} & 1837 & $<$ 0.001 & $<$ 0.001 & 0.951 & 0.001\\
Ridge & \num{1.84e-02} & \num{1e+06} & 18361 & $<$ 0.001 & $<$ 0.001 & 0.927 & 0.000\\
\bottomrule
\end{tabular}}
\end{table}

\begin{table}
\centering
\caption{Results for Specificity 99\% (percentile of predicted risk), measured with respect to mean evaluation-set performance. Bias and mean squared error (MSE) are truncated at 0.001; coverage and width are computed for intervals based on the empirical standard error (ESE). All algorithms (`GLM' = logistic regression, `RF' = random forests, `Ridge' = ridge regression), event rates, and training sample sizes are shown.\label{tab:spec_99_test}}
\centering
\resizebox{\ifdim\width>\linewidth\linewidth\else\width\fi}{!}{
\fontsize{9}{11}\selectfont
\begin{tabular}[t]{l>{\raggedright\arraybackslash}p{5em}lrll>{\raggedleft\arraybackslash}p{5em}>{\raggedleft\arraybackslash}p{5em}}
\toprule
Algorithm & Event rate & N & N event & Bias & MSE & Coverage (ESE) & Width (ESE)\\
\midrule
GLM & \num{4.59e-03} & \num{5e+03} & 23 & -0.003 & $<$ 0.001 & 0.900 & 0.019\\
GLM & \num{4.59e-03} & \num{1e+04} & 46 & $<$ 0.001 & $<$ 0.001 & 0.932 & 0.003\\
GLM & \num{4.59e-03} & \num{5e+04} & 230 & $<$ 0.001 & $<$ 0.001 & 0.940 & 0.001\\
GLM & \num{4.59e-03} & \num{1e+05} & 460 & $<$ 0.001 & $<$ 0.001 & 0.953 & 0.001\\
GLM & \num{4.59e-03} & \num{1e+06} & 4591 & $<$ 0.001 & $<$ 0.001 & 0.917 & 0.000\\
GLM & \num{9.18e-03} & \num{5e+03} & 46 & -0.001 & $<$ 0.001 & 0.936 & 0.007\\
GLM & \num{9.18e-03} & \num{1e+04} & 92 & $<$ 0.001 & $<$ 0.001 & 0.944 & 0.003\\
GLM & \num{9.18e-03} & \num{5e+04} & 460 & $<$ 0.001 & $<$ 0.001 & 0.950 & 0.001\\
GLM & \num{9.18e-03} & \num{1e+05} & 919 & $<$ 0.001 & $<$ 0.001 & 0.948 & 0.001\\
GLM & \num{9.18e-03} & \num{1e+06} & 9181 & $<$ 0.001 & $<$ 0.001 & 0.951 & 0.000\\
GLM & \num{1.84e-02} & \num{5e+03} & 92 & $<$ 0.001 & $<$ 0.001 & 0.938 & 0.006\\
GLM & \num{1.84e-02} & \num{1e+04} & 184 & $<$ 0.001 & $<$ 0.001 & 0.943 & 0.003\\
GLM & \num{1.84e-02} & \num{5e+04} & 919 & $<$ 0.001 & $<$ 0.001 & 0.948 & 0.001\\
GLM & \num{1.84e-02} & \num{1e+05} & 1837 & $<$ 0.001 & $<$ 0.001 & 0.952 & 0.001\\
GLM & \num{1.84e-02} & \num{1e+06} & 18361 & $<$ 0.001 & $<$ 0.001 & 0.897 & 0.000\\
RF & \num{4.59e-03} & \num{5e+03} & 23 & 0.001 & $<$ 0.001 & 0.786 & 0.004\\
RF & \num{4.59e-03} & \num{1e+04} & 46 & $<$ 0.001 & $<$ 0.001 & 0.942 & 0.002\\
RF & \num{4.59e-03} & \num{5e+04} & 230 & $<$ 0.001 & $<$ 0.001 & 0.948 & 0.001\\
RF & \num{4.59e-03} & \num{1e+05} & 460 & $<$ 0.001 & $<$ 0.001 & 0.944 & 0.001\\
RF & \num{4.59e-03} & \num{1e+06} & 4591 & $<$ 0.001 & $<$ 0.001 & 0.935 & 0.000\\
RF & \num{9.18e-03} & \num{5e+03} & 46 & $<$ 0.001 & $<$ 0.001 & 0.950 & 0.004\\
RF & \num{9.18e-03} & \num{1e+04} & 92 & $<$ 0.001 & $<$ 0.001 & 0.953 & 0.002\\
RF & \num{9.18e-03} & \num{5e+04} & 460 & $<$ 0.001 & $<$ 0.001 & 0.954 & 0.001\\
RF & \num{9.18e-03} & \num{1e+05} & 919 & $<$ 0.001 & $<$ 0.001 & 0.952 & 0.001\\
RF & \num{9.18e-03} & \num{1e+06} & 9181 & $<$ 0.001 & $<$ 0.001 & 0.925 & 0.000\\
RF & \num{1.84e-02} & \num{5e+03} & 92 & $<$ 0.001 & $<$ 0.001 & 0.948 & 0.004\\
RF & \num{1.84e-02} & \num{1e+04} & 184 & $<$ 0.001 & $<$ 0.001 & 0.950 & 0.003\\
RF & \num{1.84e-02} & \num{5e+04} & 919 & $<$ 0.001 & $<$ 0.001 & 0.948 & 0.001\\
RF & \num{1.84e-02} & \num{1e+05} & 1837 & $<$ 0.001 & $<$ 0.001 & 0.953 & 0.001\\
RF & \num{1.84e-02} & \num{1e+06} & 18361 & $<$ 0.001 & $<$ 0.001 & 0.961 & 0.000\\
Ridge & \num{4.59e-03} & \num{5e+03} & 23 & $<$ 0.001 & $<$ 0.001 & 0.942 & 0.003\\
Ridge & \num{4.59e-03} & \num{1e+04} & 46 & $<$ 0.001 & $<$ 0.001 & 0.955 & 0.002\\
Ridge & \num{4.59e-03} & \num{5e+04} & 230 & $<$ 0.001 & $<$ 0.001 & 0.948 & 0.001\\
Ridge & \num{4.59e-03} & \num{1e+05} & 460 & $<$ 0.001 & $<$ 0.001 & 0.947 & 0.001\\
Ridge & \num{4.59e-03} & \num{1e+06} & 4591 & $<$ 0.001 & $<$ 0.001 & 0.861 & 0.000\\
Ridge & \num{9.18e-03} & \num{5e+03} & 46 & $<$ 0.001 & $<$ 0.001 & 0.944 & 0.003\\
Ridge & \num{9.18e-03} & \num{1e+04} & 92 & $<$ 0.001 & $<$ 0.001 & 0.951 & 0.002\\
Ridge & \num{9.18e-03} & \num{5e+04} & 460 & $<$ 0.001 & $<$ 0.001 & 0.958 & 0.001\\
Ridge & \num{9.18e-03} & \num{1e+05} & 919 & $<$ 0.001 & $<$ 0.001 & 0.941 & 0.001\\
Ridge & \num{9.18e-03} & \num{1e+06} & 9181 & $<$ 0.001 & $<$ 0.001 & 0.922 & 0.000\\
Ridge & \num{1.84e-02} & \num{5e+03} & 92 & $<$ 0.001 & $<$ 0.001 & 0.942 & 0.003\\
Ridge & \num{1.84e-02} & \num{1e+04} & 184 & $<$ 0.001 & $<$ 0.001 & 0.947 & 0.002\\
Ridge & \num{1.84e-02} & \num{5e+04} & 919 & $<$ 0.001 & $<$ 0.001 & 0.949 & 0.001\\
Ridge & \num{1.84e-02} & \num{1e+05} & 1837 & $<$ 0.001 & $<$ 0.001 & 0.941 & 0.001\\
Ridge & \num{1.84e-02} & \num{1e+06} & 18361 & $<$ 0.001 & $<$ 0.001 & 0.888 & 0.000\\
\bottomrule
\end{tabular}}
\end{table}

\begin{table}
\centering
\caption{Results for Accuracy 90\% (percentile of predicted risk), measured with respect to mean evaluation-set performance. Bias and mean squared error (MSE) are truncated at 0.001; coverage and width are computed for intervals based on the empirical standard error (ESE). All algorithms (`GLM' = logistic regression, `RF' = random forests, `Ridge' = ridge regression), event rates, and training sample sizes are shown.\label{tab:accuracy_90_test}}
\centering
\resizebox{\ifdim\width>\linewidth\linewidth\else\width\fi}{!}{
\fontsize{9}{11}\selectfont
\begin{tabular}[t]{l>{\raggedright\arraybackslash}p{5em}lrll>{\raggedleft\arraybackslash}p{5em}>{\raggedleft\arraybackslash}p{5em}}
\toprule
Algorithm & Event rate & N & N event & Bias & MSE & Coverage (ESE) & Width (ESE)\\
\midrule
GLM & \num{4.59e-03} & \num{5e+03} & 23 & -0.048 & 0.074 & 0.952 & 1.204\\
GLM & \num{4.59e-03} & \num{1e+04} & 46 & $<$ 0.001 & $<$ 0.001 & 0.998 & 0.033\\
GLM & \num{4.59e-03} & \num{5e+04} & 230 & $<$ 0.001 & $<$ 0.001 & 0.950 & 0.003\\
GLM & \num{4.59e-03} & \num{1e+05} & 460 & $<$ 0.001 & $<$ 0.001 & 0.950 & 0.002\\
GLM & \num{4.59e-03} & \num{1e+06} & 4591 & $<$ 0.001 & $<$ 0.001 & 0.943 & 0.000\\
GLM & \num{9.18e-03} & \num{5e+03} & 46 & -0.004 & 0.001 & 0.981 & 0.163\\
GLM & \num{9.18e-03} & \num{1e+04} & 92 & $<$ 0.001 & $<$ 0.001 & 0.950 & 0.008\\
GLM & \num{9.18e-03} & \num{5e+04} & 460 & $<$ 0.001 & $<$ 0.001 & 0.946 & 0.003\\
GLM & \num{9.18e-03} & \num{1e+05} & 919 & $<$ 0.001 & $<$ 0.001 & 0.950 & 0.002\\
GLM & \num{9.18e-03} & \num{1e+06} & 9181 & $<$ 0.001 & $<$ 0.001 & 0.930 & 0.000\\
GLM & \num{1.84e-02} & \num{5e+03} & 92 & $<$ 0.001 & $<$ 0.001 & 0.953 & 0.012\\
GLM & \num{1.84e-02} & \num{1e+04} & 184 & $<$ 0.001 & $<$ 0.001 & 0.950 & 0.008\\
GLM & \num{1.84e-02} & \num{5e+04} & 919 & $<$ 0.001 & $<$ 0.001 & 0.951 & 0.003\\
GLM & \num{1.84e-02} & \num{1e+05} & 1837 & $<$ 0.001 & $<$ 0.001 & 0.950 & 0.002\\
GLM & \num{1.84e-02} & \num{1e+06} & 18361 & $<$ 0.001 & $<$ 0.001 & 0.943 & 0.001\\
RF & \num{4.59e-03} & \num{5e+03} & 23 & $<$ 0.001 & $<$ 0.001 & 0.937 & 0.011\\
RF & \num{4.59e-03} & \num{1e+04} & 46 & $<$ 0.001 & $<$ 0.001 & 0.954 & 0.006\\
RF & \num{4.59e-03} & \num{5e+04} & 230 & $<$ 0.001 & $<$ 0.001 & 0.948 & 0.003\\
RF & \num{4.59e-03} & \num{1e+05} & 460 & $<$ 0.001 & $<$ 0.001 & 0.944 & 0.002\\
RF & \num{4.59e-03} & \num{1e+06} & 4591 & $<$ 0.001 & $<$ 0.001 & 0.955 & 0.000\\
RF & \num{9.18e-03} & \num{5e+03} & 46 & $<$ 0.001 & $<$ 0.001 & 0.951 & 0.009\\
RF & \num{9.18e-03} & \num{1e+04} & 92 & $<$ 0.001 & $<$ 0.001 & 0.956 & 0.006\\
RF & \num{9.18e-03} & \num{5e+04} & 460 & $<$ 0.001 & $<$ 0.001 & 0.951 & 0.002\\
RF & \num{9.18e-03} & \num{1e+05} & 919 & $<$ 0.001 & $<$ 0.001 & 0.952 & 0.002\\
RF & \num{9.18e-03} & \num{1e+06} & 9181 & $<$ 0.001 & $<$ 0.001 & 0.906 & 0.000\\
RF & \num{1.84e-02} & \num{5e+03} & 92 & $<$ 0.001 & $<$ 0.001 & 0.952 & 0.010\\
RF & \num{1.84e-02} & \num{1e+04} & 184 & $<$ 0.001 & $<$ 0.001 & 0.951 & 0.007\\
RF & \num{1.84e-02} & \num{5e+04} & 919 & $<$ 0.001 & $<$ 0.001 & 0.949 & 0.003\\
RF & \num{1.84e-02} & \num{1e+05} & 1837 & $<$ 0.001 & $<$ 0.001 & 0.941 & 0.002\\
RF & \num{1.84e-02} & \num{1e+06} & 18361 & $<$ 0.001 & $<$ 0.001 & 0.945 & 0.001\\
Ridge & \num{4.59e-03} & \num{5e+03} & 23 & $<$ 0.001 & $<$ 0.001 & 0.952 & 0.007\\
Ridge & \num{4.59e-03} & \num{1e+04} & 46 & $<$ 0.001 & $<$ 0.001 & 0.948 & 0.005\\
Ridge & \num{4.59e-03} & \num{5e+04} & 230 & $<$ 0.001 & $<$ 0.001 & 0.947 & 0.002\\
Ridge & \num{4.59e-03} & \num{1e+05} & 460 & $<$ 0.001 & $<$ 0.001 & 0.949 & 0.001\\
Ridge & \num{4.59e-03} & \num{1e+06} & 4591 & $<$ 0.001 & $<$ 0.001 & 0.917 & 0.000\\
Ridge & \num{9.18e-03} & \num{5e+03} & 46 & $<$ 0.001 & $<$ 0.001 & 0.945 & 0.007\\
Ridge & \num{9.18e-03} & \num{1e+04} & 92 & $<$ 0.001 & $<$ 0.001 & 0.947 & 0.005\\
Ridge & \num{9.18e-03} & \num{5e+04} & 460 & $<$ 0.001 & $<$ 0.001 & 0.952 & 0.002\\
Ridge & \num{9.18e-03} & \num{1e+05} & 919 & $<$ 0.001 & $<$ 0.001 & 0.953 & 0.002\\
Ridge & \num{9.18e-03} & \num{1e+06} & 9181 & $<$ 0.001 & $<$ 0.001 & 0.948 & 0.000\\
Ridge & \num{1.84e-02} & \num{5e+03} & 92 & $<$ 0.001 & $<$ 0.001 & 0.957 & 0.009\\
Ridge & \num{1.84e-02} & \num{1e+04} & 184 & $<$ 0.001 & $<$ 0.001 & 0.950 & 0.006\\
Ridge & \num{1.84e-02} & \num{5e+04} & 919 & $<$ 0.001 & $<$ 0.001 & 0.946 & 0.003\\
Ridge & \num{1.84e-02} & \num{1e+05} & 1837 & $<$ 0.001 & $<$ 0.001 & 0.946 & 0.002\\
Ridge & \num{1.84e-02} & \num{1e+06} & 18361 & $<$ 0.001 & $<$ 0.001 & 0.947 & 0.001\\
\bottomrule
\end{tabular}}
\end{table}

\begin{table}
\centering
\caption{Results for Accuracy 95\% (percentile of predicted risk), measured with respect to mean evaluation-set performance. Bias and mean squared error (MSE) are truncated at 0.001; coverage and width are computed for intervals based on the empirical standard error (ESE). All algorithms (`GLM' = logistic regression, `RF' = random forests, `Ridge' = ridge regression), event rates, and training sample sizes are shown.\label{tab:accuracy_95_test}}
\centering
\resizebox{\ifdim\width>\linewidth\linewidth\else\width\fi}{!}{
\fontsize{9}{11}\selectfont
\begin{tabular}[t]{l>{\raggedright\arraybackslash}p{5em}lrll>{\raggedleft\arraybackslash}p{5em}>{\raggedleft\arraybackslash}p{5em}}
\toprule
Algorithm & Event rate & N & N event & Bias & MSE & Coverage (ESE) & Width (ESE)\\
\midrule
GLM & \num{4.59e-03} & \num{5e+03} & 23 & -0.039 & 0.064 & 0.917 & 1.188\\
GLM & \num{4.59e-03} & \num{1e+04} & 46 & $<$ 0.001 & $<$ 0.001 & 0.999 & 0.028\\
GLM & \num{4.59e-03} & \num{5e+04} & 230 & $<$ 0.001 & $<$ 0.001 & 0.946 & 0.002\\
GLM & \num{4.59e-03} & \num{1e+05} & 460 & $<$ 0.001 & $<$ 0.001 & 0.950 & 0.001\\
GLM & \num{4.59e-03} & \num{1e+06} & 4591 & $<$ 0.001 & $<$ 0.001 & 0.945 & 0.000\\
GLM & \num{9.18e-03} & \num{5e+03} & 46 & -0.003 & $<$ 0.001 & 0.983 & 0.142\\
GLM & \num{9.18e-03} & \num{1e+04} & 92 & $<$ 0.001 & $<$ 0.001 & 0.951 & 0.006\\
GLM & \num{9.18e-03} & \num{5e+04} & 460 & $<$ 0.001 & $<$ 0.001 & 0.952 & 0.002\\
GLM & \num{9.18e-03} & \num{1e+05} & 919 & $<$ 0.001 & $<$ 0.001 & 0.942 & 0.001\\
GLM & \num{9.18e-03} & \num{1e+06} & 9181 & $<$ 0.001 & $<$ 0.001 & 0.956 & 0.000\\
GLM & \num{1.84e-02} & \num{5e+03} & 92 & $<$ 0.001 & $<$ 0.001 & 0.941 & 0.009\\
GLM & \num{1.84e-02} & \num{1e+04} & 184 & $<$ 0.001 & $<$ 0.001 & 0.944 & 0.006\\
GLM & \num{1.84e-02} & \num{5e+04} & 919 & $<$ 0.001 & $<$ 0.001 & 0.950 & 0.002\\
GLM & \num{1.84e-02} & \num{1e+05} & 1837 & $<$ 0.001 & $<$ 0.001 & 0.948 & 0.002\\
GLM & \num{1.84e-02} & \num{1e+06} & 18361 & $<$ 0.001 & $<$ 0.001 & 0.941 & 0.001\\
RF & \num{4.59e-03} & \num{5e+03} & 23 & 0.001 & $<$ 0.001 & 0.923 & 0.008\\
RF & \num{4.59e-03} & \num{1e+04} & 46 & $<$ 0.001 & $<$ 0.001 & 0.954 & 0.005\\
RF & \num{4.59e-03} & \num{5e+04} & 230 & $<$ 0.001 & $<$ 0.001 & 0.952 & 0.002\\
RF & \num{4.59e-03} & \num{1e+05} & 460 & $<$ 0.001 & $<$ 0.001 & 0.947 & 0.001\\
RF & \num{4.59e-03} & \num{1e+06} & 4591 & $<$ 0.001 & $<$ 0.001 & 0.948 & 0.000\\
RF & \num{9.18e-03} & \num{5e+03} & 46 & $<$ 0.001 & $<$ 0.001 & 0.944 & 0.008\\
RF & \num{9.18e-03} & \num{1e+04} & 92 & $<$ 0.001 & $<$ 0.001 & 0.951 & 0.005\\
RF & \num{9.18e-03} & \num{5e+04} & 460 & $<$ 0.001 & $<$ 0.001 & 0.948 & 0.002\\
RF & \num{9.18e-03} & \num{1e+05} & 919 & $<$ 0.001 & $<$ 0.001 & 0.946 & 0.001\\
RF & \num{9.18e-03} & \num{1e+06} & 9181 & $<$ 0.001 & $<$ 0.001 & 0.947 & 0.000\\
RF & \num{1.84e-02} & \num{5e+03} & 92 & $<$ 0.001 & $<$ 0.001 & 0.946 & 0.009\\
RF & \num{1.84e-02} & \num{1e+04} & 184 & $<$ 0.001 & $<$ 0.001 & 0.946 & 0.006\\
RF & \num{1.84e-02} & \num{5e+04} & 919 & $<$ 0.001 & $<$ 0.001 & 0.948 & 0.003\\
RF & \num{1.84e-02} & \num{1e+05} & 1837 & $<$ 0.001 & $<$ 0.001 & 0.951 & 0.002\\
RF & \num{1.84e-02} & \num{1e+06} & 18361 & $<$ 0.001 & $<$ 0.001 & 0.946 & 0.001\\
Ridge & \num{4.59e-03} & \num{5e+03} & 23 & $<$ 0.001 & $<$ 0.001 & 0.955 & 0.005\\
Ridge & \num{4.59e-03} & \num{1e+04} & 46 & $<$ 0.001 & $<$ 0.001 & 0.946 & 0.004\\
Ridge & \num{4.59e-03} & \num{5e+04} & 230 & $<$ 0.001 & $<$ 0.001 & 0.951 & 0.002\\
Ridge & \num{4.59e-03} & \num{1e+05} & 460 & $<$ 0.001 & $<$ 0.001 & 0.946 & 0.001\\
Ridge & \num{4.59e-03} & \num{1e+06} & 4591 & $<$ 0.001 & $<$ 0.001 & 0.948 & 0.000\\
Ridge & \num{9.18e-03} & \num{5e+03} & 46 & $<$ 0.001 & $<$ 0.001 & 0.947 & 0.006\\
Ridge & \num{9.18e-03} & \num{1e+04} & 92 & $<$ 0.001 & $<$ 0.001 & 0.948 & 0.005\\
Ridge & \num{9.18e-03} & \num{5e+04} & 460 & $<$ 0.001 & $<$ 0.001 & 0.954 & 0.002\\
Ridge & \num{9.18e-03} & \num{1e+05} & 919 & $<$ 0.001 & $<$ 0.001 & 0.948 & 0.001\\
Ridge & \num{9.18e-03} & \num{1e+06} & 9181 & $<$ 0.001 & $<$ 0.001 & 0.943 & 0.000\\
Ridge & \num{1.84e-02} & \num{5e+03} & 92 & $<$ 0.001 & $<$ 0.001 & 0.950 & 0.008\\
Ridge & \num{1.84e-02} & \num{1e+04} & 184 & $<$ 0.001 & $<$ 0.001 & 0.945 & 0.006\\
Ridge & \num{1.84e-02} & \num{5e+04} & 919 & $<$ 0.001 & $<$ 0.001 & 0.946 & 0.002\\
Ridge & \num{1.84e-02} & \num{1e+05} & 1837 & $<$ 0.001 & $<$ 0.001 & 0.945 & 0.002\\
Ridge & \num{1.84e-02} & \num{1e+06} & 18361 & $<$ 0.001 & $<$ 0.001 & 0.936 & 0.001\\
\bottomrule
\end{tabular}}
\end{table}

\begin{table}
\centering
\caption{Results for Accuracy 99\% (percentile of predicted risk), measured with respect to mean evaluation-set performance. Bias and mean squared error (MSE) are truncated at 0.001; coverage and width are computed for intervals based on the empirical standard error (ESE). All algorithms (`GLM' = logistic regression, `RF' = random forests, `Ridge' = ridge regression), event rates, and training sample sizes are shown.\label{tab:accuracy_99_test}}
\centering
\resizebox{\ifdim\width>\linewidth\linewidth\else\width\fi}{!}{
\fontsize{9}{11}\selectfont
\begin{tabular}[t]{l>{\raggedright\arraybackslash}p{5em}lrll>{\raggedleft\arraybackslash}p{5em}>{\raggedleft\arraybackslash}p{5em}}
\toprule
Algorithm & Event rate & N & N event & Bias & MSE & Coverage (ESE) & Width (ESE)\\
\midrule
GLM & \num{4.59e-03} & \num{5e+03} & 23 & -0.003 & $<$ 0.001 & 0.890 & 0.017\\
GLM & \num{4.59e-03} & \num{1e+04} & 46 & $<$ 0.001 & $<$ 0.001 & 0.930 & 0.004\\
GLM & \num{4.59e-03} & \num{5e+04} & 230 & $<$ 0.001 & $<$ 0.001 & 0.938 & 0.001\\
GLM & \num{4.59e-03} & \num{1e+05} & 460 & $<$ 0.001 & $<$ 0.001 & 0.954 & 0.001\\
GLM & \num{4.59e-03} & \num{1e+06} & 4591 & $<$ 0.001 & $<$ 0.001 & 0.941 & 0.000\\
GLM & \num{9.18e-03} & \num{5e+03} & 46 & -0.001 & $<$ 0.001 & 0.921 & 0.006\\
GLM & \num{9.18e-03} & \num{1e+04} & 92 & $<$ 0.001 & $<$ 0.001 & 0.942 & 0.004\\
GLM & \num{9.18e-03} & \num{5e+04} & 460 & $<$ 0.001 & $<$ 0.001 & 0.946 & 0.002\\
GLM & \num{9.18e-03} & \num{1e+05} & 919 & $<$ 0.001 & $<$ 0.001 & 0.947 & 0.001\\
GLM & \num{9.18e-03} & \num{1e+06} & 9181 & $<$ 0.001 & $<$ 0.001 & 0.951 & 0.000\\
GLM & \num{1.84e-02} & \num{5e+03} & 92 & $<$ 0.001 & $<$ 0.001 & 0.938 & 0.008\\
GLM & \num{1.84e-02} & \num{1e+04} & 184 & $<$ 0.001 & $<$ 0.001 & 0.952 & 0.005\\
GLM & \num{1.84e-02} & \num{5e+04} & 919 & $<$ 0.001 & $<$ 0.001 & 0.951 & 0.002\\
GLM & \num{1.84e-02} & \num{1e+05} & 1837 & $<$ 0.001 & $<$ 0.001 & 0.950 & 0.002\\
GLM & \num{1.84e-02} & \num{1e+06} & 18361 & $<$ 0.001 & $<$ 0.001 & 0.925 & 0.000\\
RF & \num{4.59e-03} & \num{5e+03} & 23 & 0.001 & $<$ 0.001 & 0.849 & 0.005\\
RF & \num{4.59e-03} & \num{1e+04} & 46 & $<$ 0.001 & $<$ 0.001 & 0.961 & 0.004\\
RF & \num{4.59e-03} & \num{5e+04} & 230 & $<$ 0.001 & $<$ 0.001 & 0.947 & 0.001\\
RF & \num{4.59e-03} & \num{1e+05} & 460 & $<$ 0.001 & $<$ 0.001 & 0.942 & 0.001\\
RF & \num{4.59e-03} & \num{1e+06} & 4591 & $<$ 0.001 & $<$ 0.001 & 0.887 & 0.000\\
RF & \num{9.18e-03} & \num{5e+03} & 46 & $<$ 0.001 & $<$ 0.001 & 0.953 & 0.006\\
RF & \num{9.18e-03} & \num{1e+04} & 92 & $<$ 0.001 & $<$ 0.001 & 0.954 & 0.004\\
RF & \num{9.18e-03} & \num{5e+04} & 460 & $<$ 0.001 & $<$ 0.001 & 0.949 & 0.002\\
RF & \num{9.18e-03} & \num{1e+05} & 919 & $<$ 0.001 & $<$ 0.001 & 0.943 & 0.001\\
RF & \num{9.18e-03} & \num{1e+06} & 9181 & $<$ 0.001 & $<$ 0.001 & 0.948 & 0.000\\
RF & \num{1.84e-02} & \num{5e+03} & 92 & $<$ 0.001 & $<$ 0.001 & 0.949 & 0.008\\
RF & \num{1.84e-02} & \num{1e+04} & 184 & $<$ 0.001 & $<$ 0.001 & 0.948 & 0.005\\
RF & \num{1.84e-02} & \num{5e+04} & 919 & $<$ 0.001 & $<$ 0.001 & 0.948 & 0.002\\
RF & \num{1.84e-02} & \num{1e+05} & 1837 & $<$ 0.001 & $<$ 0.001 & 0.948 & 0.002\\
RF & \num{1.84e-02} & \num{1e+06} & 18361 & $<$ 0.001 & $<$ 0.001 & 0.946 & 0.001\\
Ridge & \num{4.59e-03} & \num{5e+03} & 23 & $<$ 0.001 & $<$ 0.001 & 0.949 & 0.004\\
Ridge & \num{4.59e-03} & \num{1e+04} & 46 & $<$ 0.001 & $<$ 0.001 & 0.958 & 0.003\\
Ridge & \num{4.59e-03} & \num{5e+04} & 230 & $<$ 0.001 & $<$ 0.001 & 0.952 & 0.001\\
Ridge & \num{4.59e-03} & \num{1e+05} & 460 & $<$ 0.001 & $<$ 0.001 & 0.951 & 0.001\\
Ridge & \num{4.59e-03} & \num{1e+06} & 4591 & $<$ 0.001 & $<$ 0.001 & 0.922 & 0.000\\
Ridge & \num{9.18e-03} & \num{5e+03} & 46 & $<$ 0.001 & $<$ 0.001 & 0.956 & 0.005\\
Ridge & \num{9.18e-03} & \num{1e+04} & 92 & $<$ 0.001 & $<$ 0.001 & 0.947 & 0.004\\
Ridge & \num{9.18e-03} & \num{5e+04} & 460 & $<$ 0.001 & $<$ 0.001 & 0.945 & 0.002\\
Ridge & \num{9.18e-03} & \num{1e+05} & 919 & $<$ 0.001 & $<$ 0.001 & 0.948 & 0.001\\
Ridge & \num{9.18e-03} & \num{1e+06} & 9181 & $<$ 0.001 & $<$ 0.001 & 0.933 & 0.000\\
Ridge & \num{1.84e-02} & \num{5e+03} & 92 & $<$ 0.001 & $<$ 0.001 & 0.956 & 0.007\\
Ridge & \num{1.84e-02} & \num{1e+04} & 184 & $<$ 0.001 & $<$ 0.001 & 0.946 & 0.005\\
Ridge & \num{1.84e-02} & \num{5e+04} & 919 & $<$ 0.001 & $<$ 0.001 & 0.950 & 0.002\\
Ridge & \num{1.84e-02} & \num{1e+05} & 1837 & $<$ 0.001 & $<$ 0.001 & 0.956 & 0.002\\
Ridge & \num{1.84e-02} & \num{1e+06} & 18361 & $<$ 0.001 & $<$ 0.001 & 0.930 & 0.001\\
\bottomrule
\end{tabular}}
\end{table}

\begin{table}
\centering
\caption{Results for PPV 90\% (percentile of predicted risk), measured with respect to mean evaluation-set performance. Bias and mean squared error (MSE) are truncated at 0.001; coverage and width are computed for intervals based on the empirical standard error (ESE). All algorithms (`GLM' = logistic regression, `RF' = random forests, `Ridge' = ridge regression), event rates, and training sample sizes are shown.\label{tab:PPV_90_test}}
\centering
\resizebox{\ifdim\width>\linewidth\linewidth\else\width\fi}{!}{
\fontsize{9}{11}\selectfont
\begin{tabular}[t]{l>{\raggedright\arraybackslash}p{5em}lrll>{\raggedleft\arraybackslash}p{5em}>{\raggedleft\arraybackslash}p{5em}}
\toprule
Algorithm & Event rate & N & N event & Bias & MSE & Coverage (ESE) & Width (ESE)\\
\midrule
GLM & \num{4.59e-03} & \num{5e+03} & 23 & $<$ 0.001 & $<$ 0.001 & 0.950 & 0.021\\
GLM & \num{4.59e-03} & \num{1e+04} & 46 & $<$ 0.001 & $<$ 0.001 & 0.950 & 0.020\\
GLM & \num{4.59e-03} & \num{5e+04} & 230 & $<$ 0.001 & $<$ 0.001 & 0.951 & 0.010\\
GLM & \num{4.59e-03} & \num{1e+05} & 460 & $<$ 0.001 & $<$ 0.001 & 0.951 & 0.007\\
GLM & \num{4.59e-03} & \num{1e+06} & 4591 & $<$ 0.001 & $<$ 0.001 & 0.948 & 0.002\\
GLM & \num{9.18e-03} & \num{5e+03} & 46 & -0.001 & $<$ 0.001 & 0.952 & 0.037\\
GLM & \num{9.18e-03} & \num{1e+04} & 92 & -0.001 & $<$ 0.001 & 0.945 & 0.029\\
GLM & \num{9.18e-03} & \num{5e+04} & 460 & $<$ 0.001 & $<$ 0.001 & 0.948 & 0.014\\
GLM & \num{9.18e-03} & \num{1e+05} & 919 & $<$ 0.001 & $<$ 0.001 & 0.952 & 0.009\\
GLM & \num{9.18e-03} & \num{1e+06} & 9181 & $<$ 0.001 & $<$ 0.001 & 0.952 & 0.003\\
GLM & \num{1.84e-02} & \num{5e+03} & 92 & -0.002 & $<$ 0.001 & 0.948 & 0.055\\
GLM & \num{1.84e-02} & \num{1e+04} & 184 & -0.001 & $<$ 0.001 & 0.946 & 0.041\\
GLM & \num{1.84e-02} & \num{5e+04} & 919 & $<$ 0.001 & $<$ 0.001 & 0.948 & 0.017\\
GLM & \num{1.84e-02} & \num{1e+05} & 1837 & $<$ 0.001 & $<$ 0.001 & 0.950 & 0.012\\
GLM & \num{1.84e-02} & \num{1e+06} & 18361 & $<$ 0.001 & $<$ 0.001 & 0.953 & 0.004\\
RF & \num{4.59e-03} & \num{5e+03} & 23 & $<$ 0.001 & $<$ 0.001 & 0.954 & 0.028\\
RF & \num{4.59e-03} & \num{1e+04} & 46 & $<$ 0.001 & $<$ 0.001 & 0.942 & 0.020\\
RF & \num{4.59e-03} & \num{5e+04} & 230 & $<$ 0.001 & $<$ 0.001 & 0.952 & 0.009\\
RF & \num{4.59e-03} & \num{1e+05} & 460 & $<$ 0.001 & $<$ 0.001 & 0.950 & 0.006\\
RF & \num{4.59e-03} & \num{1e+06} & 4591 & $<$ 0.001 & $<$ 0.001 & 0.951 & 0.002\\
RF & \num{9.18e-03} & \num{5e+03} & 46 & $<$ 0.001 & $<$ 0.001 & 0.955 & 0.039\\
RF & \num{9.18e-03} & \num{1e+04} & 92 & $<$ 0.001 & $<$ 0.001 & 0.951 & 0.027\\
RF & \num{9.18e-03} & \num{5e+04} & 460 & $<$ 0.001 & $<$ 0.001 & 0.950 & 0.012\\
RF & \num{9.18e-03} & \num{1e+05} & 919 & $<$ 0.001 & $<$ 0.001 & 0.942 & 0.008\\
RF & \num{9.18e-03} & \num{1e+06} & 9181 & $<$ 0.001 & $<$ 0.001 & 0.943 & 0.003\\
RF & \num{1.84e-02} & \num{5e+03} & 92 & $<$ 0.001 & $<$ 0.001 & 0.954 & 0.051\\
RF & \num{1.84e-02} & \num{1e+04} & 184 & $<$ 0.001 & $<$ 0.001 & 0.948 & 0.035\\
RF & \num{1.84e-02} & \num{5e+04} & 919 & $<$ 0.001 & $<$ 0.001 & 0.948 & 0.015\\
RF & \num{1.84e-02} & \num{1e+05} & 1837 & $<$ 0.001 & $<$ 0.001 & 0.951 & 0.011\\
RF & \num{1.84e-02} & \num{1e+06} & 18361 & $<$ 0.001 & $<$ 0.001 & 0.951 & 0.004\\
Ridge & \num{4.59e-03} & \num{5e+03} & 23 & $<$ 0.001 & $<$ 0.001 & 0.949 & 0.029\\
Ridge & \num{4.59e-03} & \num{1e+04} & 46 & $<$ 0.001 & $<$ 0.001 & 0.952 & 0.021\\
Ridge & \num{4.59e-03} & \num{5e+04} & 230 & $<$ 0.001 & $<$ 0.001 & 0.944 & 0.010\\
Ridge & \num{4.59e-03} & \num{1e+05} & 460 & $<$ 0.001 & $<$ 0.001 & 0.947 & 0.007\\
Ridge & \num{4.59e-03} & \num{1e+06} & 4591 & $<$ 0.001 & $<$ 0.001 & 0.955 & 0.002\\
Ridge & \num{9.18e-03} & \num{5e+03} & 46 & $<$ 0.001 & $<$ 0.001 & 0.955 & 0.040\\
Ridge & \num{9.18e-03} & \num{1e+04} & 92 & $<$ 0.001 & $<$ 0.001 & 0.955 & 0.029\\
Ridge & \num{9.18e-03} & \num{5e+04} & 460 & $<$ 0.001 & $<$ 0.001 & 0.954 & 0.013\\
Ridge & \num{9.18e-03} & \num{1e+05} & 919 & $<$ 0.001 & $<$ 0.001 & 0.951 & 0.009\\
Ridge & \num{9.18e-03} & \num{1e+06} & 9181 & $<$ 0.001 & $<$ 0.001 & 0.948 & 0.003\\
Ridge & \num{1.84e-02} & \num{5e+03} & 92 & $<$ 0.001 & $<$ 0.001 & 0.948 & 0.052\\
Ridge & \num{1.84e-02} & \num{1e+04} & 184 & $<$ 0.001 & $<$ 0.001 & 0.948 & 0.037\\
Ridge & \num{1.84e-02} & \num{5e+04} & 919 & $<$ 0.001 & $<$ 0.001 & 0.950 & 0.017\\
Ridge & \num{1.84e-02} & \num{1e+05} & 1837 & $<$ 0.001 & $<$ 0.001 & 0.947 & 0.011\\
Ridge & \num{1.84e-02} & \num{1e+06} & 18361 & $<$ 0.001 & $<$ 0.001 & 0.952 & 0.004\\
\bottomrule
\end{tabular}}
\end{table}

\begin{table}
\centering
\caption{Results for PPV 99\% (percentile of predicted risk), measured with respect to mean evaluation-set performance. Bias and mean squared error (MSE) are truncated at 0.001; coverage and width are computed for intervals based on the empirical standard error (ESE). All algorithms (`GLM' = logistic regression, `RF' = random forests, `Ridge' = ridge regression), event rates, and training sample sizes are shown.\label{tab:PPV_99_test}}
\centering
\resizebox{\ifdim\width>\linewidth\linewidth\else\width\fi}{!}{
\fontsize{9}{11}\selectfont
\begin{tabular}[t]{l>{\raggedright\arraybackslash}p{5em}lrll>{\raggedleft\arraybackslash}p{5em}>{\raggedleft\arraybackslash}p{5em}}
\toprule
Algorithm & Event rate & N & N event & Bias & MSE & Coverage (ESE) & Width (ESE)\\
\midrule
GLM & \num{4.59e-03} & \num{5e+03} & 23 & -0.001 & $<$ 0.001 & 0.960 & 0.079\\
GLM & \num{4.59e-03} & \num{1e+04} & 46 & -0.004 & $<$ 0.001 & 0.959 & 0.102\\
GLM & \num{4.59e-03} & \num{5e+04} & 230 & -0.002 & $<$ 0.001 & 0.952 & 0.066\\
GLM & \num{4.59e-03} & \num{1e+05} & 460 & $<$ 0.001 & $<$ 0.001 & 0.952 & 0.046\\
GLM & \num{4.59e-03} & \num{1e+06} & 4591 & $<$ 0.001 & $<$ 0.001 & 0.948 & 0.014\\
GLM & \num{9.18e-03} & \num{5e+03} & 46 & -0.005 & 0.002 & 0.964 & 0.167\\
GLM & \num{9.18e-03} & \num{1e+04} & 92 & -0.006 & 0.001 & 0.951 & 0.164\\
GLM & \num{9.18e-03} & \num{5e+04} & 460 & -0.002 & $<$ 0.001 & 0.944 & 0.085\\
GLM & \num{9.18e-03} & \num{1e+05} & 919 & $<$ 0.001 & $<$ 0.001 & 0.950 & 0.057\\
GLM & \num{9.18e-03} & \num{1e+06} & 9181 & $<$ 0.001 & $<$ 0.001 & 0.951 & 0.018\\
GLM & \num{1.84e-02} & \num{5e+03} & 92 & -0.008 & 0.004 & 0.963 & 0.263\\
GLM & \num{1.84e-02} & \num{1e+04} & 184 & -0.006 & 0.003 & 0.948 & 0.216\\
GLM & \num{1.84e-02} & \num{5e+04} & 919 & -0.003 & $<$ 0.001 & 0.950 & 0.096\\
GLM & \num{1.84e-02} & \num{1e+05} & 1837 & -0.001 & $<$ 0.001 & 0.952 & 0.066\\
GLM & \num{1.84e-02} & \num{1e+06} & 18361 & $<$ 0.001 & $<$ 0.001 & 0.951 & 0.020\\
RF & \num{4.59e-03} & \num{5e+03} & 23 & 0.002 & 0.002 & 0.954 & 0.187\\
RF & \num{4.59e-03} & \num{1e+04} & 46 & $<$ 0.001 & $<$ 0.001 & 0.965 & 0.123\\
RF & \num{4.59e-03} & \num{5e+04} & 230 & $<$ 0.001 & $<$ 0.001 & 0.958 & 0.050\\
RF & \num{4.59e-03} & \num{1e+05} & 460 & $<$ 0.001 & $<$ 0.001 & 0.947 & 0.035\\
RF & \num{4.59e-03} & \num{1e+06} & 4591 & $<$ 0.001 & $<$ 0.001 & 0.940 & 0.010\\
RF & \num{9.18e-03} & \num{5e+03} & 46 & 0.002 & 0.003 & 0.958 & 0.252\\
RF & \num{9.18e-03} & \num{1e+04} & 92 & $<$ 0.001 & 0.001 & 0.950 & 0.165\\
RF & \num{9.18e-03} & \num{5e+04} & 460 & $<$ 0.001 & $<$ 0.001 & 0.948 & 0.063\\
RF & \num{9.18e-03} & \num{1e+05} & 919 & $<$ 0.001 & $<$ 0.001 & 0.951 & 0.044\\
RF & \num{9.18e-03} & \num{1e+06} & 9181 & $<$ 0.001 & $<$ 0.001 & 0.948 & 0.014\\
RF & \num{1.84e-02} & \num{5e+03} & 92 & 0.002 & 0.005 & 0.956 & 0.302\\
RF & \num{1.84e-02} & \num{1e+04} & 184 & $<$ 0.001 & 0.002 & 0.947 & 0.199\\
RF & \num{1.84e-02} & \num{5e+04} & 919 & $<$ 0.001 & $<$ 0.001 & 0.954 & 0.076\\
RF & \num{1.84e-02} & \num{1e+05} & 1837 & $<$ 0.001 & $<$ 0.001 & 0.956 & 0.053\\
RF & \num{1.84e-02} & \num{1e+06} & 18361 & $<$ 0.001 & $<$ 0.001 & 0.958 & 0.017\\
Ridge & \num{4.59e-03} & \num{5e+03} & 23 & 0.002 & 0.003 & 0.948 & 0.224\\
Ridge & \num{4.59e-03} & \num{1e+04} & 46 & $<$ 0.001 & 0.001 & 0.948 & 0.152\\
Ridge & \num{4.59e-03} & \num{5e+04} & 230 & $<$ 0.001 & $<$ 0.001 & 0.950 & 0.065\\
Ridge & \num{4.59e-03} & \num{1e+05} & 460 & $<$ 0.001 & $<$ 0.001 & 0.946 & 0.046\\
Ridge & \num{4.59e-03} & \num{1e+06} & 4591 & $<$ 0.001 & $<$ 0.001 & 0.944 & 0.014\\
Ridge & \num{9.18e-03} & \num{5e+03} & 46 & 0.005 & 0.005 & 0.948 & 0.285\\
Ridge & \num{9.18e-03} & \num{1e+04} & 92 & 0.004 & 0.002 & 0.946 & 0.193\\
Ridge & \num{9.18e-03} & \num{5e+04} & 460 & $<$ 0.001 & $<$ 0.001 & 0.950 & 0.081\\
Ridge & \num{9.18e-03} & \num{1e+05} & 919 & $<$ 0.001 & $<$ 0.001 & 0.951 & 0.058\\
Ridge & \num{9.18e-03} & \num{1e+06} & 9181 & $<$ 0.001 & $<$ 0.001 & 0.948 & 0.017\\
Ridge & \num{1.84e-02} & \num{5e+03} & 92 & 0.002 & 0.006 & 0.954 & 0.332\\
Ridge & \num{1.84e-02} & \num{1e+04} & 184 & 0.004 & 0.003 & 0.957 & 0.219\\
Ridge & \num{1.84e-02} & \num{5e+04} & 919 & $<$ 0.001 & $<$ 0.001 & 0.951 & 0.092\\
Ridge & \num{1.84e-02} & \num{1e+05} & 1837 & $<$ 0.001 & $<$ 0.001 & 0.946 & 0.064\\
Ridge & \num{1.84e-02} & \num{1e+06} & 18361 & $<$ 0.001 & $<$ 0.001 & 0.952 & 0.020\\
\bottomrule
\end{tabular}}
\end{table}

\begin{table}
\centering
\caption{Results for NPV 90\% (percentile of predicted risk), measured with respect to mean evaluation-set performance. Bias and mean squared error (MSE) are truncated at 0.001; coverage and width are computed for intervals based on the empirical standard error (ESE). All algorithms (`GLM' = logistic regression, `RF' = random forests, `Ridge' = ridge regression), event rates, and training sample sizes are shown.\label{tab:NPV_90_test}}
\centering
\resizebox{\ifdim\width>\linewidth\linewidth\else\width\fi}{!}{
\fontsize{9}{11}\selectfont
\begin{tabular}[t]{l>{\raggedright\arraybackslash}p{5em}lrll>{\raggedleft\arraybackslash}p{5em}>{\raggedleft\arraybackslash}p{5em}}
\toprule
Algorithm & Event rate & N & N event & Bias & MSE & Coverage (ESE) & Width (ESE)\\
\midrule
GLM & \num{4.59e-03} & \num{5e+03} & 23 & $<$ 0.001 & $<$ 0.001 & 0.958 & 0.004\\
GLM & \num{4.59e-03} & \num{1e+04} & 46 & $<$ 0.001 & $<$ 0.001 & 0.951 & 0.002\\
GLM & \num{4.59e-03} & \num{5e+04} & 230 & $<$ 0.001 & $<$ 0.001 & 0.928 & 0.001\\
GLM & \num{4.59e-03} & \num{1e+05} & 460 & $<$ 0.001 & $<$ 0.001 & 0.932 & 0.001\\
GLM & \num{4.59e-03} & \num{1e+06} & 4591 & $<$ 0.001 & $<$ 0.001 & 0.892 & 0.000\\
GLM & \num{9.18e-03} & \num{5e+03} & 46 & $<$ 0.001 & $<$ 0.001 & 0.939 & 0.005\\
GLM & \num{9.18e-03} & \num{1e+04} & 92 & $<$ 0.001 & $<$ 0.001 & 0.942 & 0.003\\
GLM & \num{9.18e-03} & \num{5e+04} & 460 & $<$ 0.001 & $<$ 0.001 & 0.944 & 0.001\\
GLM & \num{9.18e-03} & \num{1e+05} & 919 & $<$ 0.001 & $<$ 0.001 & 0.953 & 0.001\\
GLM & \num{9.18e-03} & \num{1e+06} & 9181 & $<$ 0.001 & $<$ 0.001 & 0.911 & 0.000\\
GLM & \num{1.84e-02} & \num{5e+03} & 92 & $<$ 0.001 & $<$ 0.001 & 0.949 & 0.006\\
GLM & \num{1.84e-02} & \num{1e+04} & 184 & $<$ 0.001 & $<$ 0.001 & 0.945 & 0.004\\
GLM & \num{1.84e-02} & \num{5e+04} & 919 & $<$ 0.001 & $<$ 0.001 & 0.942 & 0.002\\
GLM & \num{1.84e-02} & \num{1e+05} & 1837 & $<$ 0.001 & $<$ 0.001 & 0.949 & 0.001\\
GLM & \num{1.84e-02} & \num{1e+06} & 18361 & $<$ 0.001 & $<$ 0.001 & 0.917 & 0.000\\
RF & \num{4.59e-03} & \num{5e+03} & 23 & $<$ 0.001 & $<$ 0.001 & 0.939 & 0.003\\
RF & \num{4.59e-03} & \num{1e+04} & 46 & $<$ 0.001 & $<$ 0.001 & 0.954 & 0.002\\
RF & \num{4.59e-03} & \num{5e+04} & 230 & $<$ 0.001 & $<$ 0.001 & 0.951 & 0.001\\
RF & \num{4.59e-03} & \num{1e+05} & 460 & $<$ 0.001 & $<$ 0.001 & 0.950 & 0.001\\
RF & \num{4.59e-03} & \num{1e+06} & 4591 & $<$ 0.001 & $<$ 0.001 & 0.937 & 0.000\\
RF & \num{9.18e-03} & \num{5e+03} & 46 & $<$ 0.001 & $<$ 0.001 & 0.952 & 0.004\\
RF & \num{9.18e-03} & \num{1e+04} & 92 & $<$ 0.001 & $<$ 0.001 & 0.947 & 0.003\\
RF & \num{9.18e-03} & \num{5e+04} & 460 & $<$ 0.001 & $<$ 0.001 & 0.951 & 0.001\\
RF & \num{9.18e-03} & \num{1e+05} & 919 & $<$ 0.001 & $<$ 0.001 & 0.952 & 0.001\\
RF & \num{9.18e-03} & \num{1e+06} & 9181 & $<$ 0.001 & $<$ 0.001 & 0.885 & 0.000\\
RF & \num{1.84e-02} & \num{5e+03} & 92 & $<$ 0.001 & $<$ 0.001 & 0.953 & 0.006\\
RF & \num{1.84e-02} & \num{1e+04} & 184 & $<$ 0.001 & $<$ 0.001 & 0.950 & 0.004\\
RF & \num{1.84e-02} & \num{5e+04} & 919 & $<$ 0.001 & $<$ 0.001 & 0.954 & 0.002\\
RF & \num{1.84e-02} & \num{1e+05} & 1837 & $<$ 0.001 & $<$ 0.001 & 0.955 & 0.001\\
RF & \num{1.84e-02} & \num{1e+06} & 18361 & $<$ 0.001 & $<$ 0.001 & 0.936 & 0.000\\
Ridge & \num{4.59e-03} & \num{5e+03} & 23 & $<$ 0.001 & $<$ 0.001 & 0.956 & 0.003\\
Ridge & \num{4.59e-03} & \num{1e+04} & 46 & $<$ 0.001 & $<$ 0.001 & 0.947 & 0.002\\
Ridge & \num{4.59e-03} & \num{5e+04} & 230 & $<$ 0.001 & $<$ 0.001 & 0.947 & 0.001\\
Ridge & \num{4.59e-03} & \num{1e+05} & 460 & $<$ 0.001 & $<$ 0.001 & 0.943 & 0.001\\
Ridge & \num{4.59e-03} & \num{1e+06} & 4591 & $<$ 0.001 & $<$ 0.001 & 0.943 & 0.000\\
Ridge & \num{9.18e-03} & \num{5e+03} & 46 & $<$ 0.001 & $<$ 0.001 & 0.956 & 0.004\\
Ridge & \num{9.18e-03} & \num{1e+04} & 92 & $<$ 0.001 & $<$ 0.001 & 0.947 & 0.003\\
Ridge & \num{9.18e-03} & \num{5e+04} & 460 & $<$ 0.001 & $<$ 0.001 & 0.946 & 0.001\\
Ridge & \num{9.18e-03} & \num{1e+05} & 919 & $<$ 0.001 & $<$ 0.001 & 0.946 & 0.001\\
Ridge & \num{9.18e-03} & \num{1e+06} & 9181 & $<$ 0.001 & $<$ 0.001 & 0.947 & 0.000\\
Ridge & \num{1.84e-02} & \num{5e+03} & 92 & $<$ 0.001 & $<$ 0.001 & 0.956 & 0.006\\
Ridge & \num{1.84e-02} & \num{1e+04} & 184 & $<$ 0.001 & $<$ 0.001 & 0.949 & 0.004\\
Ridge & \num{1.84e-02} & \num{5e+04} & 919 & $<$ 0.001 & $<$ 0.001 & 0.952 & 0.002\\
Ridge & \num{1.84e-02} & \num{1e+05} & 1837 & $<$ 0.001 & $<$ 0.001 & 0.955 & 0.001\\
Ridge & \num{1.84e-02} & \num{1e+06} & 18361 & $<$ 0.001 & $<$ 0.001 & 0.939 & 0.000\\
\bottomrule
\end{tabular}}
\end{table}

\begin{table}
\centering
\caption{Results for NPV 95\% (percentile of predicted risk), measured with respect to mean evaluation-set performance. Bias and mean squared error (MSE) are truncated at 0.001; coverage and width are computed for intervals based on the empirical standard error (ESE). All algorithms (`GLM' = logistic regression, `RF' = random forests, `Ridge' = ridge regression), event rates, and training sample sizes are shown.\label{tab:NPV_95_test}}
\centering
\resizebox{\ifdim\width>\linewidth\linewidth\else\width\fi}{!}{
\fontsize{9}{11}\selectfont
\begin{tabular}[t]{l>{\raggedright\arraybackslash}p{5em}lrll>{\raggedleft\arraybackslash}p{5em}>{\raggedleft\arraybackslash}p{5em}}
\toprule
Algorithm & Event rate & N & N event & Bias & MSE & Coverage (ESE) & Width (ESE)\\
\midrule
GLM & \num{4.59e-03} & \num{5e+03} & 23 & $<$ 0.001 & $<$ 0.001 & 0.937 & 0.004\\
GLM & \num{4.59e-03} & \num{1e+04} & 46 & $<$ 0.001 & $<$ 0.001 & 0.946 & 0.002\\
GLM & \num{4.59e-03} & \num{5e+04} & 230 & $<$ 0.001 & $<$ 0.001 & 0.948 & 0.001\\
GLM & \num{4.59e-03} & \num{1e+05} & 460 & $<$ 0.001 & $<$ 0.001 & 0.943 & 0.001\\
GLM & \num{4.59e-03} & \num{1e+06} & 4591 & $<$ 0.001 & $<$ 0.001 & 0.858 & 0.000\\
GLM & \num{9.18e-03} & \num{5e+03} & 46 & $<$ 0.001 & $<$ 0.001 & 0.947 & 0.005\\
GLM & \num{9.18e-03} & \num{1e+04} & 92 & $<$ 0.001 & $<$ 0.001 & 0.944 & 0.003\\
GLM & \num{9.18e-03} & \num{5e+04} & 460 & $<$ 0.001 & $<$ 0.001 & 0.942 & 0.001\\
GLM & \num{9.18e-03} & \num{1e+05} & 919 & $<$ 0.001 & $<$ 0.001 & 0.950 & 0.001\\
GLM & \num{9.18e-03} & \num{1e+06} & 9181 & $<$ 0.001 & $<$ 0.001 & 0.920 & 0.000\\
GLM & \num{1.84e-02} & \num{5e+03} & 92 & $<$ 0.001 & $<$ 0.001 & 0.946 & 0.006\\
GLM & \num{1.84e-02} & \num{1e+04} & 184 & $<$ 0.001 & $<$ 0.001 & 0.945 & 0.004\\
GLM & \num{1.84e-02} & \num{5e+04} & 919 & $<$ 0.001 & $<$ 0.001 & 0.949 & 0.002\\
GLM & \num{1.84e-02} & \num{1e+05} & 1837 & $<$ 0.001 & $<$ 0.001 & 0.950 & 0.001\\
GLM & \num{1.84e-02} & \num{1e+06} & 18361 & $<$ 0.001 & $<$ 0.001 & 0.922 & 0.000\\
RF & \num{4.59e-03} & \num{5e+03} & 23 & $<$ 0.001 & $<$ 0.001 & 0.947 & 0.003\\
RF & \num{4.59e-03} & \num{1e+04} & 46 & $<$ 0.001 & $<$ 0.001 & 0.947 & 0.002\\
RF & \num{4.59e-03} & \num{5e+04} & 230 & $<$ 0.001 & $<$ 0.001 & 0.950 & 0.001\\
RF & \num{4.59e-03} & \num{1e+05} & 460 & $<$ 0.001 & $<$ 0.001 & 0.948 & 0.001\\
RF & \num{4.59e-03} & \num{1e+06} & 4591 & $<$ 0.001 & $<$ 0.001 & 0.939 & 0.000\\
RF & \num{9.18e-03} & \num{5e+03} & 46 & $<$ 0.001 & $<$ 0.001 & 0.944 & 0.005\\
RF & \num{9.18e-03} & \num{1e+04} & 92 & $<$ 0.001 & $<$ 0.001 & 0.949 & 0.003\\
RF & \num{9.18e-03} & \num{5e+04} & 460 & $<$ 0.001 & $<$ 0.001 & 0.944 & 0.001\\
RF & \num{9.18e-03} & \num{1e+05} & 919 & $<$ 0.001 & $<$ 0.001 & 0.950 & 0.001\\
RF & \num{9.18e-03} & \num{1e+06} & 9181 & $<$ 0.001 & $<$ 0.001 & 0.937 & 0.000\\
RF & \num{1.84e-02} & \num{5e+03} & 92 & $<$ 0.001 & $<$ 0.001 & 0.957 & 0.007\\
RF & \num{1.84e-02} & \num{1e+04} & 184 & $<$ 0.001 & $<$ 0.001 & 0.950 & 0.005\\
RF & \num{1.84e-02} & \num{5e+04} & 919 & $<$ 0.001 & $<$ 0.001 & 0.948 & 0.002\\
RF & \num{1.84e-02} & \num{1e+05} & 1837 & $<$ 0.001 & $<$ 0.001 & 0.950 & 0.001\\
RF & \num{1.84e-02} & \num{1e+06} & 18361 & $<$ 0.001 & $<$ 0.001 & 0.956 & 0.000\\
Ridge & \num{4.59e-03} & \num{5e+03} & 23 & $<$ 0.001 & $<$ 0.001 & 0.945 & 0.003\\
Ridge & \num{4.59e-03} & \num{1e+04} & 46 & $<$ 0.001 & $<$ 0.001 & 0.947 & 0.002\\
Ridge & \num{4.59e-03} & \num{5e+04} & 230 & $<$ 0.001 & $<$ 0.001 & 0.941 & 0.001\\
Ridge & \num{4.59e-03} & \num{1e+05} & 460 & $<$ 0.001 & $<$ 0.001 & 0.949 & 0.001\\
Ridge & \num{4.59e-03} & \num{1e+06} & 4591 & $<$ 0.001 & $<$ 0.001 & 0.856 & 0.000\\
Ridge & \num{9.18e-03} & \num{5e+03} & 46 & $<$ 0.001 & $<$ 0.001 & 0.949 & 0.004\\
Ridge & \num{9.18e-03} & \num{1e+04} & 92 & $<$ 0.001 & $<$ 0.001 & 0.947 & 0.003\\
Ridge & \num{9.18e-03} & \num{5e+04} & 460 & $<$ 0.001 & $<$ 0.001 & 0.946 & 0.001\\
Ridge & \num{9.18e-03} & \num{1e+05} & 919 & $<$ 0.001 & $<$ 0.001 & 0.942 & 0.001\\
Ridge & \num{9.18e-03} & \num{1e+06} & 9181 & $<$ 0.001 & $<$ 0.001 & 0.926 & 0.000\\
Ridge & \num{1.84e-02} & \num{5e+03} & 92 & $<$ 0.001 & $<$ 0.001 & 0.950 & 0.006\\
Ridge & \num{1.84e-02} & \num{1e+04} & 184 & $<$ 0.001 & $<$ 0.001 & 0.946 & 0.004\\
Ridge & \num{1.84e-02} & \num{5e+04} & 919 & $<$ 0.001 & $<$ 0.001 & 0.949 & 0.002\\
Ridge & \num{1.84e-02} & \num{1e+05} & 1837 & $<$ 0.001 & $<$ 0.001 & 0.950 & 0.001\\
Ridge & \num{1.84e-02} & \num{1e+06} & 18361 & $<$ 0.001 & $<$ 0.001 & 0.917 & 0.000\\
\bottomrule
\end{tabular}}
\end{table}

\begin{table}
\centering
\caption{Results for NPV 99\% (percentile of predicted risk), measured with respect to mean evaluation-set performance. Bias and mean squared error (MSE) are truncated at 0.001; coverage and width are computed for intervals based on the empirical standard error (ESE). All algorithms (`GLM' = logistic regression, `RF' = random forests, `Ridge' = ridge regression), event rates, and training sample sizes are shown.\label{tab:NPV_99_test}}
\centering
\resizebox{\ifdim\width>\linewidth\linewidth\else\width\fi}{!}{
\fontsize{9}{11}\selectfont
\begin{tabular}[t]{l>{\raggedright\arraybackslash}p{5em}lrll>{\raggedleft\arraybackslash}p{5em}>{\raggedleft\arraybackslash}p{5em}}
\toprule
Algorithm & Event rate & N & N event & Bias & MSE & Coverage (ESE) & Width (ESE)\\
\midrule
GLM & \num{4.59e-03} & \num{5e+03} & 23 & $<$ 0.001 & $<$ 0.001 & 0.950 & 0.004\\
GLM & \num{4.59e-03} & \num{1e+04} & 46 & $<$ 0.001 & $<$ 0.001 & 0.952 & 0.002\\
GLM & \num{4.59e-03} & \num{5e+04} & 230 & $<$ 0.001 & $<$ 0.001 & 0.951 & 0.001\\
GLM & \num{4.59e-03} & \num{1e+05} & 460 & $<$ 0.001 & $<$ 0.001 & 0.949 & 0.001\\
GLM & \num{4.59e-03} & \num{1e+06} & 4591 & $<$ 0.001 & $<$ 0.001 & 0.928 & 0.000\\
GLM & \num{9.18e-03} & \num{5e+03} & 46 & $<$ 0.001 & $<$ 0.001 & 0.954 & 0.005\\
GLM & \num{9.18e-03} & \num{1e+04} & 92 & $<$ 0.001 & $<$ 0.001 & 0.956 & 0.003\\
GLM & \num{9.18e-03} & \num{5e+04} & 460 & $<$ 0.001 & $<$ 0.001 & 0.951 & 0.001\\
GLM & \num{9.18e-03} & \num{1e+05} & 919 & $<$ 0.001 & $<$ 0.001 & 0.954 & 0.001\\
GLM & \num{9.18e-03} & \num{1e+06} & 9181 & $<$ 0.001 & $<$ 0.001 & 0.926 & 0.000\\
GLM & \num{1.84e-02} & \num{5e+03} & 92 & $<$ 0.001 & $<$ 0.001 & 0.946 & 0.007\\
GLM & \num{1.84e-02} & \num{1e+04} & 184 & $<$ 0.001 & $<$ 0.001 & 0.941 & 0.004\\
GLM & \num{1.84e-02} & \num{5e+04} & 919 & $<$ 0.001 & $<$ 0.001 & 0.954 & 0.002\\
GLM & \num{1.84e-02} & \num{1e+05} & 1837 & $<$ 0.001 & $<$ 0.001 & 0.950 & 0.001\\
GLM & \num{1.84e-02} & \num{1e+06} & 18361 & $<$ 0.001 & $<$ 0.001 & 0.930 & 0.000\\
RF & \num{4.59e-03} & \num{5e+03} & 23 & $<$ 0.001 & $<$ 0.001 & 0.948 & 0.004\\
RF & \num{4.59e-03} & \num{1e+04} & 46 & $<$ 0.001 & $<$ 0.001 & 0.955 & 0.003\\
RF & \num{4.59e-03} & \num{5e+04} & 230 & $<$ 0.001 & $<$ 0.001 & 0.948 & 0.001\\
RF & \num{4.59e-03} & \num{1e+05} & 460 & $<$ 0.001 & $<$ 0.001 & 0.948 & 0.001\\
RF & \num{4.59e-03} & \num{1e+06} & 4591 & $<$ 0.001 & $<$ 0.001 & 0.938 & 0.000\\
RF & \num{9.18e-03} & \num{5e+03} & 46 & $<$ 0.001 & $<$ 0.001 & 0.950 & 0.005\\
RF & \num{9.18e-03} & \num{1e+04} & 92 & $<$ 0.001 & $<$ 0.001 & 0.954 & 0.003\\
RF & \num{9.18e-03} & \num{5e+04} & 460 & $<$ 0.001 & $<$ 0.001 & 0.952 & 0.002\\
RF & \num{9.18e-03} & \num{1e+05} & 919 & $<$ 0.001 & $<$ 0.001 & 0.948 & 0.001\\
RF & \num{9.18e-03} & \num{1e+06} & 9181 & $<$ 0.001 & $<$ 0.001 & 0.924 & 0.000\\
RF & \num{1.84e-02} & \num{5e+03} & 92 & $<$ 0.001 & $<$ 0.001 & 0.951 & 0.007\\
RF & \num{1.84e-02} & \num{1e+04} & 184 & $<$ 0.001 & $<$ 0.001 & 0.950 & 0.005\\
RF & \num{1.84e-02} & \num{5e+04} & 919 & $<$ 0.001 & $<$ 0.001 & 0.951 & 0.002\\
RF & \num{1.84e-02} & \num{1e+05} & 1837 & $<$ 0.001 & $<$ 0.001 & 0.950 & 0.002\\
RF & \num{1.84e-02} & \num{1e+06} & 18361 & $<$ 0.001 & $<$ 0.001 & 0.948 & 0.000\\
Ridge & \num{4.59e-03} & \num{5e+03} & 23 & $<$ 0.001 & $<$ 0.001 & 0.958 & 0.003\\
Ridge & \num{4.59e-03} & \num{1e+04} & 46 & $<$ 0.001 & $<$ 0.001 & 0.951 & 0.002\\
Ridge & \num{4.59e-03} & \num{5e+04} & 230 & $<$ 0.001 & $<$ 0.001 & 0.948 & 0.001\\
Ridge & \num{4.59e-03} & \num{1e+05} & 460 & $<$ 0.001 & $<$ 0.001 & 0.949 & 0.001\\
Ridge & \num{4.59e-03} & \num{1e+06} & 4591 & $<$ 0.001 & $<$ 0.001 & 0.937 & 0.000\\
Ridge & \num{9.18e-03} & \num{5e+03} & 46 & $<$ 0.001 & $<$ 0.001 & 0.960 & 0.005\\
Ridge & \num{9.18e-03} & \num{1e+04} & 92 & $<$ 0.001 & $<$ 0.001 & 0.953 & 0.003\\
Ridge & \num{9.18e-03} & \num{5e+04} & 460 & $<$ 0.001 & $<$ 0.001 & 0.949 & 0.001\\
Ridge & \num{9.18e-03} & \num{1e+05} & 919 & $<$ 0.001 & $<$ 0.001 & 0.948 & 0.001\\
Ridge & \num{9.18e-03} & \num{1e+06} & 9181 & $<$ 0.001 & $<$ 0.001 & 0.937 & 0.000\\
Ridge & \num{1.84e-02} & \num{5e+03} & 92 & $<$ 0.001 & $<$ 0.001 & 0.954 & 0.007\\
Ridge & \num{1.84e-02} & \num{1e+04} & 184 & $<$ 0.001 & $<$ 0.001 & 0.946 & 0.005\\
Ridge & \num{1.84e-02} & \num{5e+04} & 919 & $<$ 0.001 & $<$ 0.001 & 0.953 & 0.002\\
Ridge & \num{1.84e-02} & \num{1e+05} & 1837 & $<$ 0.001 & $<$ 0.001 & 0.953 & 0.001\\
Ridge & \num{1.84e-02} & \num{1e+06} & 18361 & $<$ 0.001 & $<$ 0.001 & 0.943 & 0.000\\
\bottomrule
\end{tabular}}
\end{table}

\begin{table}
\centering
\caption{Results for F1 90\% (percentile of predicted risk), measured with respect to mean evaluation-set performance. Bias and mean squared error (MSE) are truncated at 0.001; coverage and width are computed for intervals based on the empirical standard error (ESE). All algorithms (`GLM' = logistic regression, `RF' = random forests, `Ridge' = ridge regression), event rates, and training sample sizes are shown.\label{tab:F1_90_test}}
\centering
\resizebox{\ifdim\width>\linewidth\linewidth\else\width\fi}{!}{
\fontsize{9}{11}\selectfont
\begin{tabular}[t]{l>{\raggedright\arraybackslash}p{5em}lrll>{\raggedleft\arraybackslash}p{5em}>{\raggedleft\arraybackslash}p{5em}}
\toprule
Algorithm & Event rate & N & N event & Bias & MSE & Coverage (ESE) & Width (ESE)\\
\midrule
GLM & \num{4.59e-03} & \num{5e+03} & 23 & $<$ 0.001 & $<$ 0.001 & 0.950 & 0.039\\
GLM & \num{4.59e-03} & \num{1e+04} & 46 & -0.002 & $<$ 0.001 & 0.944 & 0.037\\
GLM & \num{4.59e-03} & \num{5e+04} & 230 & $<$ 0.001 & $<$ 0.001 & 0.951 & 0.019\\
GLM & \num{4.59e-03} & \num{1e+05} & 460 & $<$ 0.001 & $<$ 0.001 & 0.951 & 0.013\\
GLM & \num{4.59e-03} & \num{1e+06} & 4591 & $<$ 0.001 & $<$ 0.001 & 0.948 & 0.004\\
GLM & \num{9.18e-03} & \num{5e+03} & 46 & -0.002 & $<$ 0.001 & 0.951 & 0.065\\
GLM & \num{9.18e-03} & \num{1e+04} & 92 & -0.003 & $<$ 0.001 & 0.946 & 0.052\\
GLM & \num{9.18e-03} & \num{5e+04} & 460 & $<$ 0.001 & $<$ 0.001 & 0.949 & 0.025\\
GLM & \num{9.18e-03} & \num{1e+05} & 919 & $<$ 0.001 & $<$ 0.001 & 0.950 & 0.016\\
GLM & \num{9.18e-03} & \num{1e+06} & 9181 & $<$ 0.001 & $<$ 0.001 & 0.955 & 0.005\\
GLM & \num{1.84e-02} & \num{5e+03} & 92 & -0.004 & $<$ 0.001 & 0.948 & 0.089\\
GLM & \num{1.84e-02} & \num{1e+04} & 184 & -0.003 & $<$ 0.001 & 0.944 & 0.065\\
GLM & \num{1.84e-02} & \num{5e+04} & 919 & $<$ 0.001 & $<$ 0.001 & 0.946 & 0.027\\
GLM & \num{1.84e-02} & \num{1e+05} & 1837 & $<$ 0.001 & $<$ 0.001 & 0.950 & 0.019\\
GLM & \num{1.84e-02} & \num{1e+06} & 18361 & $<$ 0.001 & $<$ 0.001 & 0.957 & 0.006\\
RF & \num{4.59e-03} & \num{5e+03} & 23 & $<$ 0.001 & $<$ 0.001 & 0.952 & 0.052\\
RF & \num{4.59e-03} & \num{1e+04} & 46 & $<$ 0.001 & $<$ 0.001 & 0.942 & 0.037\\
RF & \num{4.59e-03} & \num{5e+04} & 230 & $<$ 0.001 & $<$ 0.001 & 0.953 & 0.016\\
RF & \num{4.59e-03} & \num{1e+05} & 460 & $<$ 0.001 & $<$ 0.001 & 0.949 & 0.012\\
RF & \num{4.59e-03} & \num{1e+06} & 4591 & $<$ 0.001 & $<$ 0.001 & 0.950 & 0.003\\
RF & \num{9.18e-03} & \num{5e+03} & 46 & $<$ 0.001 & $<$ 0.001 & 0.955 & 0.068\\
RF & \num{9.18e-03} & \num{1e+04} & 92 & $<$ 0.001 & $<$ 0.001 & 0.953 & 0.047\\
RF & \num{9.18e-03} & \num{5e+04} & 460 & $<$ 0.001 & $<$ 0.001 & 0.950 & 0.020\\
RF & \num{9.18e-03} & \num{1e+05} & 919 & $<$ 0.001 & $<$ 0.001 & 0.941 & 0.015\\
RF & \num{9.18e-03} & \num{1e+06} & 9181 & $<$ 0.001 & $<$ 0.001 & 0.945 & 0.005\\
RF & \num{1.84e-02} & \num{5e+03} & 92 & $<$ 0.001 & $<$ 0.001 & 0.955 & 0.080\\
RF & \num{1.84e-02} & \num{1e+04} & 184 & $<$ 0.001 & $<$ 0.001 & 0.948 & 0.055\\
RF & \num{1.84e-02} & \num{5e+04} & 919 & $<$ 0.001 & $<$ 0.001 & 0.947 & 0.024\\
RF & \num{1.84e-02} & \num{1e+05} & 1837 & $<$ 0.001 & $<$ 0.001 & 0.946 & 0.017\\
RF & \num{1.84e-02} & \num{1e+06} & 18361 & $<$ 0.001 & $<$ 0.001 & 0.949 & 0.006\\
Ridge & \num{4.59e-03} & \num{5e+03} & 23 & $<$ 0.001 & $<$ 0.001 & 0.948 & 0.054\\
Ridge & \num{4.59e-03} & \num{1e+04} & 46 & $<$ 0.001 & $<$ 0.001 & 0.952 & 0.039\\
Ridge & \num{4.59e-03} & \num{5e+04} & 230 & $<$ 0.001 & $<$ 0.001 & 0.945 & 0.018\\
Ridge & \num{4.59e-03} & \num{1e+05} & 460 & $<$ 0.001 & $<$ 0.001 & 0.948 & 0.012\\
Ridge & \num{4.59e-03} & \num{1e+06} & 4591 & $<$ 0.001 & $<$ 0.001 & 0.954 & 0.004\\
Ridge & \num{9.18e-03} & \num{5e+03} & 46 & $<$ 0.001 & $<$ 0.001 & 0.956 & 0.069\\
Ridge & \num{9.18e-03} & \num{1e+04} & 92 & $<$ 0.001 & $<$ 0.001 & 0.954 & 0.050\\
Ridge & \num{9.18e-03} & \num{5e+04} & 460 & $<$ 0.001 & $<$ 0.001 & 0.953 & 0.023\\
Ridge & \num{9.18e-03} & \num{1e+05} & 919 & $<$ 0.001 & $<$ 0.001 & 0.950 & 0.016\\
Ridge & \num{9.18e-03} & \num{1e+06} & 9181 & $<$ 0.001 & $<$ 0.001 & 0.952 & 0.005\\
Ridge & \num{1.84e-02} & \num{5e+03} & 92 & $<$ 0.001 & $<$ 0.001 & 0.946 & 0.082\\
Ridge & \num{1.84e-02} & \num{1e+04} & 184 & $<$ 0.001 & $<$ 0.001 & 0.950 & 0.059\\
Ridge & \num{1.84e-02} & \num{5e+04} & 919 & $<$ 0.001 & $<$ 0.001 & 0.948 & 0.026\\
Ridge & \num{1.84e-02} & \num{1e+05} & 1837 & $<$ 0.001 & $<$ 0.001 & 0.948 & 0.018\\
Ridge & \num{1.84e-02} & \num{1e+06} & 18361 & $<$ 0.001 & $<$ 0.001 & 0.951 & 0.006\\
\bottomrule
\end{tabular}}
\end{table}

\begin{table}
\centering
\caption{Results for F1 95\% (percentile of predicted risk), measured with respect to mean evaluation-set performance. Bias and mean squared error (MSE) are truncated at 0.001; coverage and width are computed for intervals based on the empirical standard error (ESE). All algorithms (`GLM' = logistic regression, `RF' = random forests, `Ridge' = ridge regression), event rates, and training sample sizes are shown.\label{tab:F1_95_test}}
\centering
\resizebox{\ifdim\width>\linewidth\linewidth\else\width\fi}{!}{
\fontsize{9}{11}\selectfont
\begin{tabular}[t]{l>{\raggedright\arraybackslash}p{5em}lrll>{\raggedleft\arraybackslash}p{5em}>{\raggedleft\arraybackslash}p{5em}}
\toprule
Algorithm & Event rate & N & N event & Bias & MSE & Coverage (ESE) & Width (ESE)\\
\midrule
GLM & \num{4.59e-03} & \num{5e+03} & 23 & $<$ 0.001 & $<$ 0.001 & 0.953 & 0.062\\
GLM & \num{4.59e-03} & \num{1e+04} & 46 & -0.002 & $<$ 0.001 & 0.952 & 0.059\\
GLM & \num{4.59e-03} & \num{5e+04} & 230 & -0.001 & $<$ 0.001 & 0.949 & 0.032\\
GLM & \num{4.59e-03} & \num{1e+05} & 460 & $<$ 0.001 & $<$ 0.001 & 0.948 & 0.022\\
GLM & \num{4.59e-03} & \num{1e+06} & 4591 & $<$ 0.001 & $<$ 0.001 & 0.943 & 0.007\\
GLM & \num{9.18e-03} & \num{5e+03} & 46 & -0.004 & $<$ 0.001 & 0.954 & 0.097\\
GLM & \num{9.18e-03} & \num{1e+04} & 92 & -0.005 & $<$ 0.001 & 0.945 & 0.082\\
GLM & \num{9.18e-03} & \num{5e+04} & 460 & -0.001 & $<$ 0.001 & 0.944 & 0.039\\
GLM & \num{9.18e-03} & \num{1e+05} & 919 & $<$ 0.001 & $<$ 0.001 & 0.950 & 0.026\\
GLM & \num{9.18e-03} & \num{1e+06} & 9181 & $<$ 0.001 & $<$ 0.001 & 0.950 & 0.008\\
GLM & \num{1.84e-02} & \num{5e+03} & 92 & -0.006 & $<$ 0.001 & 0.948 & 0.124\\
GLM & \num{1.84e-02} & \num{1e+04} & 184 & -0.005 & $<$ 0.001 & 0.939 & 0.095\\
GLM & \num{1.84e-02} & \num{5e+04} & 919 & -0.001 & $<$ 0.001 & 0.947 & 0.040\\
GLM & \num{1.84e-02} & \num{1e+05} & 1837 & $<$ 0.001 & $<$ 0.001 & 0.951 & 0.028\\
GLM & \num{1.84e-02} & \num{1e+06} & 18361 & $<$ 0.001 & $<$ 0.001 & 0.946 & 0.009\\
RF & \num{4.59e-03} & \num{5e+03} & 23 & $<$ 0.001 & $<$ 0.001 & 0.956 & 0.084\\
RF & \num{4.59e-03} & \num{1e+04} & 46 & $<$ 0.001 & $<$ 0.001 & 0.953 & 0.060\\
RF & \num{4.59e-03} & \num{5e+04} & 230 & $<$ 0.001 & $<$ 0.001 & 0.945 & 0.026\\
RF & \num{4.59e-03} & \num{1e+05} & 460 & $<$ 0.001 & $<$ 0.001 & 0.950 & 0.019\\
RF & \num{4.59e-03} & \num{1e+06} & 4591 & $<$ 0.001 & $<$ 0.001 & 0.942 & 0.006\\
RF & \num{9.18e-03} & \num{5e+03} & 46 & $<$ 0.001 & $<$ 0.001 & 0.952 & 0.107\\
RF & \num{9.18e-03} & \num{1e+04} & 92 & $<$ 0.001 & $<$ 0.001 & 0.948 & 0.074\\
RF & \num{9.18e-03} & \num{5e+04} & 460 & $<$ 0.001 & $<$ 0.001 & 0.951 & 0.031\\
RF & \num{9.18e-03} & \num{1e+05} & 919 & $<$ 0.001 & $<$ 0.001 & 0.951 & 0.022\\
RF & \num{9.18e-03} & \num{1e+06} & 9181 & $<$ 0.001 & $<$ 0.001 & 0.948 & 0.007\\
RF & \num{1.84e-02} & \num{5e+03} & 92 & -0.001 & $<$ 0.001 & 0.954 & 0.115\\
RF & \num{1.84e-02} & \num{1e+04} & 184 & $<$ 0.001 & $<$ 0.001 & 0.950 & 0.078\\
RF & \num{1.84e-02} & \num{5e+04} & 919 & $<$ 0.001 & $<$ 0.001 & 0.949 & 0.034\\
RF & \num{1.84e-02} & \num{1e+05} & 1837 & $<$ 0.001 & $<$ 0.001 & 0.945 & 0.024\\
RF & \num{1.84e-02} & \num{1e+06} & 18361 & $<$ 0.001 & $<$ 0.001 & 0.941 & 0.008\\
Ridge & \num{4.59e-03} & \num{5e+03} & 23 & $<$ 0.001 & $<$ 0.001 & 0.951 & 0.093\\
Ridge & \num{4.59e-03} & \num{1e+04} & 46 & $<$ 0.001 & $<$ 0.001 & 0.951 & 0.067\\
Ridge & \num{4.59e-03} & \num{5e+04} & 230 & $<$ 0.001 & $<$ 0.001 & 0.952 & 0.030\\
Ridge & \num{4.59e-03} & \num{1e+05} & 460 & $<$ 0.001 & $<$ 0.001 & 0.948 & 0.021\\
Ridge & \num{4.59e-03} & \num{1e+06} & 4591 & $<$ 0.001 & $<$ 0.001 & 0.953 & 0.007\\
Ridge & \num{9.18e-03} & \num{5e+03} & 46 & $<$ 0.001 & $<$ 0.001 & 0.954 & 0.112\\
Ridge & \num{9.18e-03} & \num{1e+04} & 92 & $<$ 0.001 & $<$ 0.001 & 0.951 & 0.080\\
Ridge & \num{9.18e-03} & \num{5e+04} & 460 & $<$ 0.001 & $<$ 0.001 & 0.948 & 0.036\\
Ridge & \num{9.18e-03} & \num{1e+05} & 919 & $<$ 0.001 & $<$ 0.001 & 0.947 & 0.025\\
Ridge & \num{9.18e-03} & \num{1e+06} & 9181 & $<$ 0.001 & $<$ 0.001 & 0.946 & 0.008\\
Ridge & \num{1.84e-02} & \num{5e+03} & 92 & -0.002 & $<$ 0.001 & 0.950 & 0.121\\
Ridge & \num{1.84e-02} & \num{1e+04} & 184 & $<$ 0.001 & $<$ 0.001 & 0.955 & 0.088\\
Ridge & \num{1.84e-02} & \num{5e+04} & 919 & $<$ 0.001 & $<$ 0.001 & 0.951 & 0.039\\
Ridge & \num{1.84e-02} & \num{1e+05} & 1837 & $<$ 0.001 & $<$ 0.001 & 0.948 & 0.026\\
Ridge & \num{1.84e-02} & \num{1e+06} & 18361 & $<$ 0.001 & $<$ 0.001 & 0.950 & 0.009\\
\bottomrule
\end{tabular}}
\end{table}

\begin{table}
\centering
\caption{Results for F1 99\% (percentile of predicted risk), measured with respect to mean evaluation-set performance. Bias and mean squared error (MSE) are truncated at 0.001; coverage and width are computed for intervals based on the empirical standard error (ESE). All algorithms (`GLM' = logistic regression, `RF' = random forests, `Ridge' = ridge regression), event rates, and training sample sizes are shown.\label{tab:F1_99_test}}
\centering
\resizebox{\ifdim\width>\linewidth\linewidth\else\width\fi}{!}{
\fontsize{9}{11}\selectfont
\begin{tabular}[t]{l>{\raggedright\arraybackslash}p{5em}lrll>{\raggedleft\arraybackslash}p{5em}>{\raggedleft\arraybackslash}p{5em}}
\toprule
Algorithm & Event rate & N & N event & Bias & MSE & Coverage (ESE) & Width (ESE)\\
\midrule
GLM & \num{4.59e-03} & \num{5e+03} & 23 & -0.002 & $<$ 0.001 & 0.965 & 0.111\\
GLM & \num{4.59e-03} & \num{1e+04} & 46 & -0.007 & $<$ 0.001 & 0.963 & 0.126\\
GLM & \num{4.59e-03} & \num{5e+04} & 230 & -0.004 & $<$ 0.001 & 0.947 & 0.082\\
GLM & \num{4.59e-03} & \num{1e+05} & 460 & -0.002 & $<$ 0.001 & 0.951 & 0.057\\
GLM & \num{4.59e-03} & \num{1e+06} & 4591 & $<$ 0.001 & $<$ 0.001 & 0.952 & 0.017\\
GLM & \num{9.18e-03} & \num{5e+03} & 46 & -0.008 & 0.002 & 0.964 & 0.167\\
GLM & \num{9.18e-03} & \num{1e+04} & 92 & -0.01 & 0.001 & 0.950 & 0.153\\
GLM & \num{9.18e-03} & \num{5e+04} & 460 & -0.004 & $<$ 0.001 & 0.937 & 0.079\\
GLM & \num{9.18e-03} & \num{1e+05} & 919 & -0.002 & $<$ 0.001 & 0.944 & 0.053\\
GLM & \num{9.18e-03} & \num{1e+06} & 9181 & $<$ 0.001 & $<$ 0.001 & 0.951 & 0.017\\
GLM & \num{1.84e-02} & \num{5e+03} & 92 & -0.012 & 0.002 & 0.950 & 0.177\\
GLM & \num{1.84e-02} & \num{1e+04} & 184 & -0.009 & 0.001 & 0.940 & 0.143\\
GLM & \num{1.84e-02} & \num{5e+04} & 919 & -0.004 & $<$ 0.001 & 0.943 & 0.063\\
GLM & \num{1.84e-02} & \num{1e+05} & 1837 & -0.002 & $<$ 0.001 & 0.950 & 0.043\\
GLM & \num{1.84e-02} & \num{1e+06} & 18361 & $<$ 0.001 & $<$ 0.001 & 0.956 & 0.013\\
RF & \num{4.59e-03} & \num{5e+03} & 23 & -0.003 & 0.002 & 0.965 & 0.209\\
RF & \num{4.59e-03} & \num{1e+04} & 46 & -0.002 & 0.001 & 0.965 & 0.152\\
RF & \num{4.59e-03} & \num{5e+04} & 230 & $<$ 0.001 & $<$ 0.001 & 0.959 & 0.064\\
RF & \num{4.59e-03} & \num{1e+05} & 460 & $<$ 0.001 & $<$ 0.001 & 0.949 & 0.045\\
RF & \num{4.59e-03} & \num{1e+06} & 4591 & $<$ 0.001 & $<$ 0.001 & 0.943 & 0.013\\
RF & \num{9.18e-03} & \num{5e+03} & 46 & -0.006 & 0.003 & 0.962 & 0.219\\
RF & \num{9.18e-03} & \num{1e+04} & 92 & -0.005 & 0.001 & 0.947 & 0.153\\
RF & \num{9.18e-03} & \num{5e+04} & 460 & $<$ 0.001 & $<$ 0.001 & 0.950 & 0.061\\
RF & \num{9.18e-03} & \num{1e+05} & 919 & $<$ 0.001 & $<$ 0.001 & 0.949 & 0.043\\
RF & \num{9.18e-03} & \num{1e+06} & 9181 & $<$ 0.001 & $<$ 0.001 & 0.946 & 0.013\\
RF & \num{1.84e-02} & \num{5e+03} & 92 & -0.006 & 0.002 & 0.953 & 0.182\\
RF & \num{1.84e-02} & \num{1e+04} & 184 & -0.005 & 0.001 & 0.950 & 0.127\\
RF & \num{1.84e-02} & \num{5e+04} & 919 & $<$ 0.001 & $<$ 0.001 & 0.952 & 0.052\\
RF & \num{1.84e-02} & \num{1e+05} & 1837 & -0.001 & $<$ 0.001 & 0.950 & 0.036\\
RF & \num{1.84e-02} & \num{1e+06} & 18361 & $<$ 0.001 & $<$ 0.001 & 0.949 & 0.011\\
Ridge & \num{4.59e-03} & \num{5e+03} & 23 & -0.008 & 0.004 & 0.947 & 0.247\\
Ridge & \num{4.59e-03} & \num{1e+04} & 46 & -0.005 & 0.002 & 0.947 & 0.177\\
Ridge & \num{4.59e-03} & \num{5e+04} & 230 & -0.002 & $<$ 0.001 & 0.952 & 0.079\\
Ridge & \num{4.59e-03} & \num{1e+05} & 460 & $<$ 0.001 & $<$ 0.001 & 0.943 & 0.056\\
Ridge & \num{4.59e-03} & \num{1e+06} & 4591 & $<$ 0.001 & $<$ 0.001 & 0.942 & 0.018\\
Ridge & \num{9.18e-03} & \num{5e+03} & 46 & -0.009 & 0.003 & 0.950 & 0.235\\
Ridge & \num{9.18e-03} & \num{1e+04} & 92 & -0.005 & 0.002 & 0.944 & 0.169\\
Ridge & \num{9.18e-03} & \num{5e+04} & 460 & -0.002 & $<$ 0.001 & 0.950 & 0.075\\
Ridge & \num{9.18e-03} & \num{1e+05} & 919 & -0.002 & $<$ 0.001 & 0.952 & 0.054\\
Ridge & \num{9.18e-03} & \num{1e+06} & 9181 & $<$ 0.001 & $<$ 0.001 & 0.955 & 0.016\\
Ridge & \num{1.84e-02} & \num{5e+03} & 92 & -0.011 & 0.003 & 0.946 & 0.196\\
Ridge & \num{1.84e-02} & \num{1e+04} & 184 & -0.006 & 0.001 & 0.953 & 0.140\\
Ridge & \num{1.84e-02} & \num{5e+04} & 919 & -0.002 & $<$ 0.001 & 0.954 & 0.061\\
Ridge & \num{1.84e-02} & \num{1e+05} & 1837 & $<$ 0.001 & $<$ 0.001 & 0.945 & 0.043\\
Ridge & \num{1.84e-02} & \num{1e+06} & 18361 & $<$ 0.001 & $<$ 0.001 & 0.948 & 0.014\\
\bottomrule
\end{tabular}}
\end{table}

\begin{table}
\centering
\caption{Results for F0.5 90\% (percentile of predicted risk), measured with respect to mean evaluation-set performance. Bias and mean squared error (MSE) are truncated at 0.001; coverage and width are computed for intervals based on the empirical standard error (ESE). All algorithms (`GLM' = logistic regression, `RF' = random forests, `Ridge' = ridge regression), event rates, and training sample sizes are shown.\label{tab:F0.5_90_test}}
\centering
\resizebox{\ifdim\width>\linewidth\linewidth\else\width\fi}{!}{
\fontsize{9}{11}\selectfont
\begin{tabular}[t]{l>{\raggedright\arraybackslash}p{5em}lrll>{\raggedleft\arraybackslash}p{5em}>{\raggedleft\arraybackslash}p{5em}}
\toprule
Algorithm & Event rate & N & N event & Bias & MSE & Coverage (ESE) & Width (ESE)\\
\midrule
GLM & \num{4.59e-03} & \num{5e+03} & 23 & 0.009 & $<$ 0.001 & 0.860 & 0.041\\
GLM & \num{4.59e-03} & \num{1e+04} & 46 & 0.003 & $<$ 0.001 & 0.894 & 0.020\\
GLM & \num{4.59e-03} & \num{5e+04} & 230 & $<$ 0.001 & $<$ 0.001 & 0.951 & 0.013\\
GLM & \num{4.59e-03} & \num{1e+05} & 460 & $<$ 0.001 & $<$ 0.001 & 0.952 & 0.009\\
GLM & \num{4.59e-03} & \num{1e+06} & 4591 & $<$ 0.001 & $<$ 0.001 & 0.950 & 0.003\\
GLM & \num{9.18e-03} & \num{5e+03} & 46 & 0.008 & $<$ 0.001 & 0.863 & 0.037\\
GLM & \num{9.18e-03} & \num{1e+04} & 92 & -0.001 & $<$ 0.001 & 0.950 & 0.034\\
GLM & \num{9.18e-03} & \num{5e+04} & 460 & $<$ 0.001 & $<$ 0.001 & 0.948 & 0.017\\
GLM & \num{9.18e-03} & \num{1e+05} & 919 & $<$ 0.001 & $<$ 0.001 & 0.951 & 0.011\\
GLM & \num{9.18e-03} & \num{1e+06} & 9181 & $<$ 0.001 & $<$ 0.001 & 0.954 & 0.004\\
GLM & \num{1.84e-02} & \num{5e+03} & 92 & $<$ 0.001 & $<$ 0.001 & 0.954 & 0.060\\
GLM & \num{1.84e-02} & \num{1e+04} & 184 & -0.002 & $<$ 0.001 & 0.946 & 0.048\\
GLM & \num{1.84e-02} & \num{5e+04} & 919 & $<$ 0.001 & $<$ 0.001 & 0.947 & 0.020\\
GLM & \num{1.84e-02} & \num{1e+05} & 1837 & $<$ 0.001 & $<$ 0.001 & 0.950 & 0.014\\
GLM & \num{1.84e-02} & \num{1e+06} & 18361 & $<$ 0.001 & $<$ 0.001 & 0.955 & 0.004\\
RF & \num{4.59e-03} & \num{5e+03} & 23 & 0.009 & $<$ 0.001 & 0.732 & 0.025\\
RF & \num{4.59e-03} & \num{1e+04} & 46 & 0.002 & $<$ 0.001 & 0.942 & 0.022\\
RF & \num{4.59e-03} & \num{5e+04} & 230 & $<$ 0.001 & $<$ 0.001 & 0.953 & 0.011\\
RF & \num{4.59e-03} & \num{1e+05} & 460 & $<$ 0.001 & $<$ 0.001 & 0.950 & 0.008\\
RF & \num{4.59e-03} & \num{1e+06} & 4591 & $<$ 0.001 & $<$ 0.001 & 0.952 & 0.002\\
RF & \num{9.18e-03} & \num{5e+03} & 46 & 0.003 & $<$ 0.001 & 0.941 & 0.041\\
RF & \num{9.18e-03} & \num{1e+04} & 92 & $<$ 0.001 & $<$ 0.001 & 0.952 & 0.032\\
RF & \num{9.18e-03} & \num{5e+04} & 460 & $<$ 0.001 & $<$ 0.001 & 0.949 & 0.014\\
RF & \num{9.18e-03} & \num{1e+05} & 919 & $<$ 0.001 & $<$ 0.001 & 0.942 & 0.010\\
RF & \num{9.18e-03} & \num{1e+06} & 9181 & $<$ 0.001 & $<$ 0.001 & 0.943 & 0.003\\
RF & \num{1.84e-02} & \num{5e+03} & 92 & $<$ 0.001 & $<$ 0.001 & 0.953 & 0.058\\
RF & \num{1.84e-02} & \num{1e+04} & 184 & $<$ 0.001 & $<$ 0.001 & 0.948 & 0.041\\
RF & \num{1.84e-02} & \num{5e+04} & 919 & $<$ 0.001 & $<$ 0.001 & 0.946 & 0.018\\
RF & \num{1.84e-02} & \num{1e+05} & 1837 & $<$ 0.001 & $<$ 0.001 & 0.950 & 0.013\\
RF & \num{1.84e-02} & \num{1e+06} & 18361 & $<$ 0.001 & $<$ 0.001 & 0.948 & 0.004\\
Ridge & \num{4.59e-03} & \num{5e+03} & 23 & 0.006 & $<$ 0.001 & 0.863 & 0.029\\
Ridge & \num{4.59e-03} & \num{1e+04} & 46 & $<$ 0.001 & $<$ 0.001 & 0.946 & 0.024\\
Ridge & \num{4.59e-03} & \num{5e+04} & 230 & $<$ 0.001 & $<$ 0.001 & 0.944 & 0.012\\
Ridge & \num{4.59e-03} & \num{1e+05} & 460 & $<$ 0.001 & $<$ 0.001 & 0.947 & 0.008\\
Ridge & \num{4.59e-03} & \num{1e+06} & 4591 & $<$ 0.001 & $<$ 0.001 & 0.957 & 0.003\\
Ridge & \num{9.18e-03} & \num{5e+03} & 46 & 0.002 & $<$ 0.001 & 0.942 & 0.045\\
Ridge & \num{9.18e-03} & \num{1e+04} & 92 & $<$ 0.001 & $<$ 0.001 & 0.954 & 0.034\\
Ridge & \num{9.18e-03} & \num{5e+04} & 460 & $<$ 0.001 & $<$ 0.001 & 0.954 & 0.016\\
Ridge & \num{9.18e-03} & \num{1e+05} & 919 & $<$ 0.001 & $<$ 0.001 & 0.950 & 0.011\\
Ridge & \num{9.18e-03} & \num{1e+06} & 9181 & $<$ 0.001 & $<$ 0.001 & 0.948 & 0.003\\
Ridge & \num{1.84e-02} & \num{5e+03} & 92 & $<$ 0.001 & $<$ 0.001 & 0.949 & 0.060\\
Ridge & \num{1.84e-02} & \num{1e+04} & 184 & $<$ 0.001 & $<$ 0.001 & 0.948 & 0.044\\
Ridge & \num{1.84e-02} & \num{5e+04} & 919 & $<$ 0.001 & $<$ 0.001 & 0.948 & 0.019\\
Ridge & \num{1.84e-02} & \num{1e+05} & 1837 & $<$ 0.001 & $<$ 0.001 & 0.947 & 0.013\\
Ridge & \num{1.84e-02} & \num{1e+06} & 18361 & $<$ 0.001 & $<$ 0.001 & 0.951 & 0.004\\
\bottomrule
\end{tabular}}
\end{table}

\begin{table}
\centering
\caption{Results for F0.5 95\% (percentile of predicted risk), measured with respect to mean evaluation-set performance. Bias and mean squared error (MSE) are truncated at 0.001; coverage and width are computed for intervals based on the empirical standard error (ESE). All algorithms (`GLM' = logistic regression, `RF' = random forests, `Ridge' = ridge regression), event rates, and training sample sizes are shown.\label{tab:F0.5_95_test}}
\centering
\resizebox{\ifdim\width>\linewidth\linewidth\else\width\fi}{!}{
\fontsize{9}{11}\selectfont
\begin{tabular}[t]{l>{\raggedright\arraybackslash}p{5em}lrll>{\raggedleft\arraybackslash}p{5em}>{\raggedleft\arraybackslash}p{5em}}
\toprule
Algorithm & Event rate & N & N event & Bias & MSE & Coverage (ESE) & Width (ESE)\\
\midrule
GLM & \num{4.59e-03} & \num{5e+03} & 23 & 0.022 & $<$ 0.001 & 0.843 & 0.084\\
GLM & \num{4.59e-03} & \num{1e+04} & 46 & 0.01 & $<$ 0.001 & 0.786 & 0.033\\
GLM & \num{4.59e-03} & \num{5e+04} & 230 & $<$ 0.001 & $<$ 0.001 & 0.950 & 0.022\\
GLM & \num{4.59e-03} & \num{1e+05} & 460 & $<$ 0.001 & $<$ 0.001 & 0.948 & 0.015\\
GLM & \num{4.59e-03} & \num{1e+06} & 4591 & $<$ 0.001 & $<$ 0.001 & 0.945 & 0.005\\
GLM & \num{9.18e-03} & \num{5e+03} & 46 & 0.022 & $<$ 0.001 & 0.730 & 0.061\\
GLM & \num{9.18e-03} & \num{1e+04} & 92 & $<$ 0.001 & $<$ 0.001 & 0.951 & 0.055\\
GLM & \num{9.18e-03} & \num{5e+04} & 460 & $<$ 0.001 & $<$ 0.001 & 0.944 & 0.029\\
GLM & \num{9.18e-03} & \num{1e+05} & 919 & $<$ 0.001 & $<$ 0.001 & 0.952 & 0.020\\
GLM & \num{9.18e-03} & \num{1e+06} & 9181 & $<$ 0.001 & $<$ 0.001 & 0.952 & 0.006\\
GLM & \num{1.84e-02} & \num{5e+03} & 92 & 0.005 & $<$ 0.001 & 0.948 & 0.088\\
GLM & \num{1.84e-02} & \num{1e+04} & 184 & -0.003 & $<$ 0.001 & 0.942 & 0.078\\
GLM & \num{1.84e-02} & \num{5e+04} & 919 & $<$ 0.001 & $<$ 0.001 & 0.947 & 0.033\\
GLM & \num{1.84e-02} & \num{1e+05} & 1837 & $<$ 0.001 & $<$ 0.001 & 0.950 & 0.024\\
GLM & \num{1.84e-02} & \num{1e+06} & 18361 & $<$ 0.001 & $<$ 0.001 & 0.947 & 0.007\\
RF & \num{4.59e-03} & \num{5e+03} & 23 & 0.025 & $<$ 0.001 & 0.420 & 0.043\\
RF & \num{4.59e-03} & \num{1e+04} & 46 & 0.007 & $<$ 0.001 & 0.878 & 0.035\\
RF & \num{4.59e-03} & \num{5e+04} & 230 & $<$ 0.001 & $<$ 0.001 & 0.947 & 0.018\\
RF & \num{4.59e-03} & \num{1e+05} & 460 & $<$ 0.001 & $<$ 0.001 & 0.951 & 0.013\\
RF & \num{4.59e-03} & \num{1e+06} & 4591 & $<$ 0.001 & $<$ 0.001 & 0.946 & 0.004\\
RF & \num{9.18e-03} & \num{5e+03} & 46 & 0.015 & $<$ 0.001 & 0.863 & 0.065\\
RF & \num{9.18e-03} & \num{1e+04} & 92 & 0.002 & $<$ 0.001 & 0.948 & 0.051\\
RF & \num{9.18e-03} & \num{5e+04} & 460 & $<$ 0.001 & $<$ 0.001 & 0.948 & 0.023\\
RF & \num{9.18e-03} & \num{1e+05} & 919 & $<$ 0.001 & $<$ 0.001 & 0.949 & 0.016\\
RF & \num{9.18e-03} & \num{1e+06} & 9181 & $<$ 0.001 & $<$ 0.001 & 0.946 & 0.005\\
RF & \num{1.84e-02} & \num{5e+03} & 92 & 0.005 & $<$ 0.001 & 0.946 & 0.089\\
RF & \num{1.84e-02} & \num{1e+04} & 184 & $<$ 0.001 & $<$ 0.001 & 0.952 & 0.064\\
RF & \num{1.84e-02} & \num{5e+04} & 919 & $<$ 0.001 & $<$ 0.001 & 0.950 & 0.028\\
RF & \num{1.84e-02} & \num{1e+05} & 1837 & $<$ 0.001 & $<$ 0.001 & 0.946 & 0.020\\
RF & \num{1.84e-02} & \num{1e+06} & 18361 & $<$ 0.001 & $<$ 0.001 & 0.945 & 0.006\\
Ridge & \num{4.59e-03} & \num{5e+03} & 23 & 0.018 & $<$ 0.001 & 0.747 & 0.051\\
Ridge & \num{4.59e-03} & \num{1e+04} & 46 & 0.003 & $<$ 0.001 & 0.940 & 0.041\\
Ridge & \num{4.59e-03} & \num{5e+04} & 230 & $<$ 0.001 & $<$ 0.001 & 0.953 & 0.021\\
Ridge & \num{4.59e-03} & \num{1e+05} & 460 & $<$ 0.001 & $<$ 0.001 & 0.949 & 0.015\\
Ridge & \num{4.59e-03} & \num{1e+06} & 4591 & $<$ 0.001 & $<$ 0.001 & 0.951 & 0.005\\
Ridge & \num{9.18e-03} & \num{5e+03} & 46 & 0.009 & $<$ 0.001 & 0.921 & 0.074\\
Ridge & \num{9.18e-03} & \num{1e+04} & 92 & $<$ 0.001 & $<$ 0.001 & 0.950 & 0.058\\
Ridge & \num{9.18e-03} & \num{5e+04} & 460 & $<$ 0.001 & $<$ 0.001 & 0.947 & 0.027\\
Ridge & \num{9.18e-03} & \num{1e+05} & 919 & $<$ 0.001 & $<$ 0.001 & 0.950 & 0.019\\
Ridge & \num{9.18e-03} & \num{1e+06} & 9181 & $<$ 0.001 & $<$ 0.001 & 0.945 & 0.006\\
Ridge & \num{1.84e-02} & \num{5e+03} & 92 & 0.002 & $<$ 0.001 & 0.950 & 0.097\\
Ridge & \num{1.84e-02} & \num{1e+04} & 184 & $<$ 0.001 & $<$ 0.001 & 0.956 & 0.073\\
Ridge & \num{1.84e-02} & \num{5e+04} & 919 & $<$ 0.001 & $<$ 0.001 & 0.954 & 0.032\\
Ridge & \num{1.84e-02} & \num{1e+05} & 1837 & $<$ 0.001 & $<$ 0.001 & 0.948 & 0.022\\
Ridge & \num{1.84e-02} & \num{1e+06} & 18361 & $<$ 0.001 & $<$ 0.001 & 0.951 & 0.007\\
\bottomrule
\end{tabular}}
\end{table}

\begin{table}
\centering
\caption{Results for F0.5 99\% (percentile of predicted risk), measured with respect to mean evaluation-set performance. Bias and mean squared error (MSE) are truncated at 0.001; coverage and width are computed for intervals based on the empirical standard error (ESE). All algorithms (`GLM' = logistic regression, `RF' = random forests, `Ridge' = ridge regression), event rates, and training sample sizes are shown.\label{tab:F0.5_99_test}}
\centering
\resizebox{\ifdim\width>\linewidth\linewidth\else\width\fi}{!}{
\fontsize{9}{11}\selectfont
\begin{tabular}[t]{l>{\raggedright\arraybackslash}p{5em}lrll>{\raggedleft\arraybackslash}p{5em}>{\raggedleft\arraybackslash}p{5em}}
\toprule
Algorithm & Event rate & N & N event & Bias & MSE & Coverage (ESE) & Width (ESE)\\
\midrule
GLM & \num{4.59e-03} & \num{5e+03} & 23 & 0.096 & 0.012 & 0.606 & 0.195\\
GLM & \num{4.59e-03} & \num{1e+04} & 46 & 0.067 & 0.005 & 0.431 & 0.116\\
GLM & \num{4.59e-03} & \num{5e+04} & 230 & -0.003 & $<$ 0.001 & 0.947 & 0.071\\
GLM & \num{4.59e-03} & \num{1e+05} & 460 & -0.001 & $<$ 0.001 & 0.953 & 0.050\\
GLM & \num{4.59e-03} & \num{1e+06} & 4591 & $<$ 0.001 & $<$ 0.001 & 0.947 & 0.015\\
GLM & \num{9.18e-03} & \num{5e+03} & 46 & 0.102 & 0.013 & 0.494 & 0.193\\
GLM & \num{9.18e-03} & \num{1e+04} & 92 & 0.03 & 0.002 & 0.854 & 0.133\\
GLM & \num{9.18e-03} & \num{5e+04} & 460 & -0.003 & $<$ 0.001 & 0.945 & 0.082\\
GLM & \num{9.18e-03} & \num{1e+05} & 919 & -0.001 & $<$ 0.001 & 0.948 & 0.055\\
GLM & \num{9.18e-03} & \num{1e+06} & 9181 & $<$ 0.001 & $<$ 0.001 & 0.952 & 0.018\\
GLM & \num{1.84e-02} & \num{5e+03} & 92 & 0.06 & 0.006 & 0.758 & 0.178\\
GLM & \num{1.84e-02} & \num{1e+04} & 184 & 0.002 & 0.002 & 0.954 & 0.161\\
GLM & \num{1.84e-02} & \num{5e+04} & 919 & -0.004 & $<$ 0.001 & 0.943 & 0.078\\
GLM & \num{1.84e-02} & \num{1e+05} & 1837 & -0.002 & $<$ 0.001 & 0.950 & 0.053\\
GLM & \num{1.84e-02} & \num{1e+06} & 18361 & $<$ 0.001 & $<$ 0.001 & 0.953 & 0.016\\
RF & \num{4.59e-03} & \num{5e+03} & 23 & 0.186 & 0.041 & 0.364 & 0.309\\
RF & \num{4.59e-03} & \num{1e+04} & 46 & 0.072 & 0.006 & 0.416 & 0.126\\
RF & \num{4.59e-03} & \num{5e+04} & 230 & 0.002 & $<$ 0.001 & 0.954 & 0.053\\
RF & \num{4.59e-03} & \num{1e+05} & 460 & $<$ 0.001 & $<$ 0.001 & 0.946 & 0.038\\
RF & \num{4.59e-03} & \num{1e+06} & 4591 & $<$ 0.001 & $<$ 0.001 & 0.942 & 0.011\\
RF & \num{9.18e-03} & \num{5e+03} & 46 & 0.126 & 0.019 & 0.430 & 0.229\\
RF & \num{9.18e-03} & \num{1e+04} & 92 & 0.041 & 0.003 & 0.777 & 0.132\\
RF & \num{9.18e-03} & \num{5e+04} & 460 & $<$ 0.001 & $<$ 0.001 & 0.951 & 0.062\\
RF & \num{9.18e-03} & \num{1e+05} & 919 & $<$ 0.001 & $<$ 0.001 & 0.949 & 0.043\\
RF & \num{9.18e-03} & \num{1e+06} & 9181 & $<$ 0.001 & $<$ 0.001 & 0.944 & 0.013\\
RF & \num{1.84e-02} & \num{5e+03} & 92 & 0.077 & 0.008 & 0.661 & 0.192\\
RF & \num{1.84e-02} & \num{1e+04} & 184 & 0.015 & 0.001 & 0.927 & 0.142\\
RF & \num{1.84e-02} & \num{5e+04} & 919 & $<$ 0.001 & $<$ 0.001 & 0.953 & 0.063\\
RF & \num{1.84e-02} & \num{1e+05} & 1837 & -0.001 & $<$ 0.001 & 0.951 & 0.044\\
RF & \num{1.84e-02} & \num{1e+06} & 18361 & $<$ 0.001 & $<$ 0.001 & 0.953 & 0.014\\
Ridge & \num{4.59e-03} & \num{5e+03} & 23 & 0.137 & 0.023 & 0.446 & 0.247\\
Ridge & \num{4.59e-03} & \num{1e+04} & 46 & 0.044 & 0.003 & 0.754 & 0.134\\
Ridge & \num{4.59e-03} & \num{5e+04} & 230 & $<$ 0.001 & $<$ 0.001 & 0.951 & 0.069\\
Ridge & \num{4.59e-03} & \num{1e+05} & 460 & $<$ 0.001 & $<$ 0.001 & 0.945 & 0.049\\
Ridge & \num{4.59e-03} & \num{1e+06} & 4591 & $<$ 0.001 & $<$ 0.001 & 0.945 & 0.016\\
Ridge & \num{9.18e-03} & \num{5e+03} & 46 & 0.09 & 0.011 & 0.618 & 0.210\\
Ridge & \num{9.18e-03} & \num{1e+04} & 92 & 0.021 & 0.002 & 0.916 & 0.159\\
Ridge & \num{9.18e-03} & \num{5e+04} & 460 & $<$ 0.001 & $<$ 0.001 & 0.951 & 0.078\\
Ridge & \num{9.18e-03} & \num{1e+05} & 919 & -0.001 & $<$ 0.001 & 0.952 & 0.056\\
Ridge & \num{9.18e-03} & \num{1e+06} & 9181 & $<$ 0.001 & $<$ 0.001 & 0.953 & 0.017\\
Ridge & \num{1.84e-02} & \num{5e+03} & 92 & 0.042 & 0.004 & 0.877 & 0.205\\
Ridge & \num{1.84e-02} & \num{1e+04} & 184 & 0.002 & 0.002 & 0.957 & 0.164\\
Ridge & \num{1.84e-02} & \num{5e+04} & 919 & -0.002 & $<$ 0.001 & 0.950 & 0.075\\
Ridge & \num{1.84e-02} & \num{1e+05} & 1837 & $<$ 0.001 & $<$ 0.001 & 0.946 & 0.052\\
Ridge & \num{1.84e-02} & \num{1e+06} & 18361 & $<$ 0.001 & $<$ 0.001 & 0.949 & 0.017\\
\bottomrule
\end{tabular}}
\end{table}

\begin{table}
\centering
\caption{Results for Brier score, measured with respect to mean evaluation-set performance. Bias and mean squared error (MSE) are truncated at 0.001; coverage and width are computed for intervals based on the empirical standard error (ESE). All algorithms (`GLM' = logistic regression, `RF' = random forests, `Ridge' = ridge regression), event rates, and training sample sizes are shown.\label{tab:brier_test}}
\centering
\resizebox{\ifdim\width>\linewidth\linewidth\else\width\fi}{!}{
\fontsize{9}{11}\selectfont
\begin{tabular}[t]{l>{\raggedright\arraybackslash}p{5em}lrll>{\raggedleft\arraybackslash}p{5em}>{\raggedleft\arraybackslash}p{5em}}
\toprule
Algorithm & Event rate & N & N event & Bias & MSE & Coverage (ESE) & Width (ESE)\\
\midrule
GLM & \num{4.59e-03} & \num{5e+03} & 23 & 0.002 & $<$ 0.001 & 0.918 & 0.021\\
GLM & \num{4.59e-03} & \num{1e+04} & 46 & $<$ 0.001 & $<$ 0.001 & 0.925 & 0.003\\
GLM & \num{4.59e-03} & \num{5e+04} & 230 & $<$ 0.001 & $<$ 0.001 & 0.955 & 0.001\\
GLM & \num{4.59e-03} & \num{1e+05} & 460 & $<$ 0.001 & $<$ 0.001 & 0.948 & 0.001\\
GLM & \num{4.59e-03} & \num{1e+06} & 4591 & $<$ 0.001 & $<$ 0.001 & 0.896 & 0.000\\
GLM & \num{9.18e-03} & \num{5e+03} & 46 & $<$ 0.001 & $<$ 0.001 & 0.926 & 0.006\\
GLM & \num{9.18e-03} & \num{1e+04} & 92 & $<$ 0.001 & $<$ 0.001 & 0.939 & 0.003\\
GLM & \num{9.18e-03} & \num{5e+04} & 460 & $<$ 0.001 & $<$ 0.001 & 0.949 & 0.001\\
GLM & \num{9.18e-03} & \num{1e+05} & 919 & $<$ 0.001 & $<$ 0.001 & 0.949 & 0.001\\
GLM & \num{9.18e-03} & \num{1e+06} & 9181 & $<$ 0.001 & $<$ 0.001 & 0.947 & 0.000\\
GLM & \num{1.84e-02} & \num{5e+03} & 92 & $<$ 0.001 & $<$ 0.001 & 0.941 & 0.007\\
GLM & \num{1.84e-02} & \num{1e+04} & 184 & $<$ 0.001 & $<$ 0.001 & 0.938 & 0.004\\
GLM & \num{1.84e-02} & \num{5e+04} & 919 & $<$ 0.001 & $<$ 0.001 & 0.951 & 0.002\\
GLM & \num{1.84e-02} & \num{1e+05} & 1837 & $<$ 0.001 & $<$ 0.001 & 0.950 & 0.001\\
GLM & \num{1.84e-02} & \num{1e+06} & 18361 & $<$ 0.001 & $<$ 0.001 & 0.929 & 0.000\\
RF & \num{4.59e-03} & \num{5e+03} & 23 & -0.004 & 0.004 & 0.021 & 0.004\\
RF & \num{4.59e-03} & \num{1e+04} & 46 & -0.006 & 0.006 & 0.000 & 0.003\\
RF & \num{4.59e-03} & \num{5e+04} & 230 & -0.005 & 0.005 & 0.000 & 0.001\\
RF & \num{4.59e-03} & \num{1e+05} & 460 & -0.006 & 0.006 & 0.000 & 0.001\\
RF & \num{4.59e-03} & \num{1e+06} & 4591 & $<$ 0.001 & $<$ 0.001 & 0.951 & 0.000\\
RF & \num{9.18e-03} & \num{5e+03} & 46 & $<$ 0.001 & $<$ 0.001 & 0.949 & 0.005\\
RF & \num{9.18e-03} & \num{1e+04} & 92 & $<$ 0.001 & $<$ 0.001 & 0.952 & 0.003\\
RF & \num{9.18e-03} & \num{5e+04} & 460 & $<$ 0.001 & $<$ 0.001 & 0.949 & 0.002\\
RF & \num{9.18e-03} & \num{1e+05} & 919 & $<$ 0.001 & $<$ 0.001 & 0.953 & 0.001\\
RF & \num{9.18e-03} & \num{1e+06} & 9181 & $<$ 0.001 & $<$ 0.001 & 0.948 & 0.000\\
RF & \num{1.84e-02} & \num{5e+03} & 92 & $<$ 0.001 & $<$ 0.001 & 0.954 & 0.007\\
RF & \num{1.84e-02} & \num{1e+04} & 184 & $<$ 0.001 & $<$ 0.001 & 0.950 & 0.005\\
RF & \num{1.84e-02} & \num{5e+04} & 919 & $<$ 0.001 & $<$ 0.001 & 0.949 & 0.002\\
RF & \num{1.84e-02} & \num{1e+05} & 1837 & $<$ 0.001 & $<$ 0.001 & 0.947 & 0.001\\
RF & \num{1.84e-02} & \num{1e+06} & 18361 & $<$ 0.001 & $<$ 0.001 & 0.943 & 0.000\\
Ridge & \num{4.59e-03} & \num{5e+03} & 23 & $<$ 0.001 & $<$ 0.001 & 0.952 & 0.004\\
Ridge & \num{4.59e-03} & \num{1e+04} & 46 & $<$ 0.001 & $<$ 0.001 & 0.951 & 0.003\\
Ridge & \num{4.59e-03} & \num{5e+04} & 230 & $<$ 0.001 & $<$ 0.001 & 0.944 & 0.001\\
Ridge & \num{4.59e-03} & \num{1e+05} & 460 & $<$ 0.001 & $<$ 0.001 & 0.950 & 0.001\\
Ridge & \num{4.59e-03} & \num{1e+06} & 4591 & $<$ 0.001 & $<$ 0.001 & 0.930 & 0.000\\
Ridge & \num{9.18e-03} & \num{5e+03} & 46 & $<$ 0.001 & $<$ 0.001 & 0.955 & 0.005\\
Ridge & \num{9.18e-03} & \num{1e+04} & 92 & $<$ 0.001 & $<$ 0.001 & 0.951 & 0.003\\
Ridge & \num{9.18e-03} & \num{5e+04} & 460 & $<$ 0.001 & $<$ 0.001 & 0.950 & 0.001\\
Ridge & \num{9.18e-03} & \num{1e+05} & 919 & $<$ 0.001 & $<$ 0.001 & 0.946 & 0.001\\
Ridge & \num{9.18e-03} & \num{1e+06} & 9181 & $<$ 0.001 & $<$ 0.001 & 0.947 & 0.000\\
Ridge & \num{1.84e-02} & \num{5e+03} & 92 & $<$ 0.001 & $<$ 0.001 & 0.955 & 0.006\\
Ridge & \num{1.84e-02} & \num{1e+04} & 184 & $<$ 0.001 & $<$ 0.001 & 0.946 & 0.004\\
Ridge & \num{1.84e-02} & \num{5e+04} & 919 & $<$ 0.001 & $<$ 0.001 & 0.952 & 0.002\\
Ridge & \num{1.84e-02} & \num{1e+05} & 1837 & $<$ 0.001 & $<$ 0.001 & 0.949 & 0.001\\
Ridge & \num{1.84e-02} & \num{1e+06} & 18361 & $<$ 0.001 & $<$ 0.001 & 0.931 & 0.000\\
\bottomrule
\end{tabular}}
\end{table}

\clearpage

\subsection{Determining fixed tuning parameters}\label{sec:fixing_overview}

To determine the values of fixed tuning parameters, we ran a small pilot simulation study. Here, we again varied $n \in \{5, 10, 50, 100, 1000\} \times 10^3$ and considered the three event rates defined in the main manuscript. In this pilot simulation, we bootstrap sampled outcomes along with covariates; we do not expect that this made a meaningful difference in the behavior of AUC. This investigation consisted of two parts: 1) nested cross-validation for tuning parameter selection (inner folds) and estimating AUC (outer folds); and 2) a small number of simulations, with relatively few replications, to assess if there was a large difference between the cross-validated AUC when the selected fixed tuning parameters were used and the cross-validated AUC obtained when we used 20-fold cross-validation to select the tuning parameter in the sampled training set. We referred to the second part of this fixed tuning parameter investigations as \textit{pilot simulations}. For ridge regression, we used internal tuning from the R function \texttt{cv.glmnet}. For random forests, we set the number of trees to be 250 for $n < 1 \times 10^5$, 100 for $n = 1\times 10^5$, and 50 for $n = 1 \times 10^6$ and tuned over grids of minimum node size given in Table~\ref{tab:rf_minnode_grid}. We assessed how variable the selected tuning parameters were across simulation replicates, and determined that the variability was small enough to justify fixing the tuning parameters as described above.

\begin{table}[h]
\centering
\caption[Minimum node size tuning grids]{Grid of minimum node sizes tuned over for each training set size, random forest.}
\label{tab:rf_minnode_grid}
\begin{tabular}{| c | c |} 
 \hline
 Training set size & Grid of minimum node sizes\\ 
 \hline\hline
 $5\times 10^3$ & $\{10,100,1000\}$ \\
 $1\times 10^4$ & $\{10,100,1000\}$\\
 $5\times 10^4$ & $\{100, 1000, 10000 \}$\\
 $1\times 10^5$ & $\{1000, 10000, 25000, 50000\}$\\
 $1\times 10^6$ & $\{10000, 25000, 50000\}$\\
 \hline
\end{tabular}
\end{table}

\ifnotarXiv


    \end{document}
\fi
\fi

{\small
\printbibliography
}

\end{document}